\documentclass[10pt,twocolumn,letterpaper]{article}

\usepackage{cvpr}
\usepackage{times}
\usepackage{epsfig}
\usepackage{graphicx}
\usepackage{amsmath}
\usepackage{amssymb}

\usepackage{subcaption}
\usepackage{bm}
\usepackage{cases}
\usepackage{siunitx}
\newcommand{\rulesep}{\color{black} \unskip\ \vrule\ }

\usepackage[breaklinks=true,bookmarks=false]{hyperref}

\cvprfinalcopy 

\def\cvprPaperID{****} 

\begin{document}

\title{Total3DUnderstanding: Joint Layout, Object Pose and Mesh Reconstruction for Indoor Scenes from a Single Image}

\author{Yinyu Nie\textsuperscript{1,2,3,}$^{\dagger}$, Xiaoguang Han\textsuperscript{2,3,}$^{\ast}$, Shihui Guo\textsuperscript{4}, Yujian Zheng\textsuperscript{2,3}, Jian Chang\textsuperscript{1}, Jian Jun Zhang\textsuperscript{1}\\
\textsuperscript{1}Bournemouth University\ \ \ \
\textsuperscript{2}The Chinese University of Hong Kong, Shenzhen\\
\textsuperscript{3}Shenzhen Research Institute of Big Data\ \ \ \ \
\textsuperscript{4}Xiamen University
}


\maketitle
\let\thefootnote\relax\footnotetext{$^{\dagger}$ Work done during visiting CUHKSZ and SRIBD.}
\let\thefootnote\relax\footnotetext{$^{\ast}$ Corresponding author: {\tt\small hanxiaoguang@cuhk.edu.cn}}
\begin{abstract}
	Semantic reconstruction of indoor scenes refers to both scene understanding and object reconstruction. Existing works either address one part of this problem or focus on independent objects. In this paper, we bridge the gap between understanding and reconstruction, and propose an end-to-end solution to jointly reconstruct room layout, object bounding boxes and meshes from a single image. Instead of separately resolving scene understanding and object reconstruction, our method builds upon a holistic scene context and proposes a coarse-to-fine hierarchy with three components: 1. room layout with camera pose; 2. 3D object bounding boxes; 3. object meshes. We argue that understanding the context of each component can assist the task of parsing the others, which enables joint understanding and reconstruction. The experiments on the SUN RGB-D and Pix3D datasets demonstrate that our method consistently outperforms existing methods in indoor layout estimation, 3D object detection and mesh reconstruction.
\end{abstract}

\section{Introduction}

Semantic reconstruction from an indoor image shows its unique importance in applications such as interior design and real estate. 
In recent years, this topic has received a rocketing interest from researchers in both computer vision and graphics communities.
However, the inherent ambiguity in depth perception, the clutter and complexity of real-world environments make it still challenging to
fully recover the scene context (both semantics and geometry) merely from a single image.

\begin{figure}[!ht]
	\centering
	\begin{subfigure}[t]{0.15\textwidth}
		\includegraphics[width=\textwidth]  
		{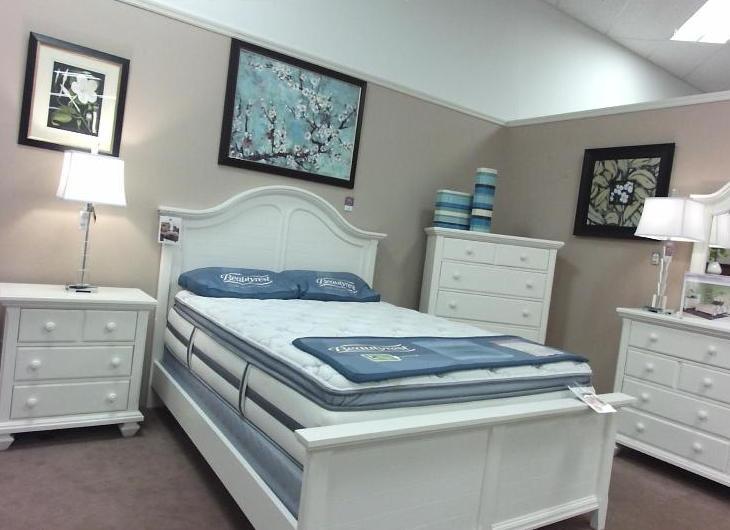}
		\includegraphics[width=\textwidth]
		{Figure/1/514.jpg}
	\end{subfigure}
	\begin{subfigure}[t]{0.15\textwidth}
		\includegraphics[width=\textwidth]  
		{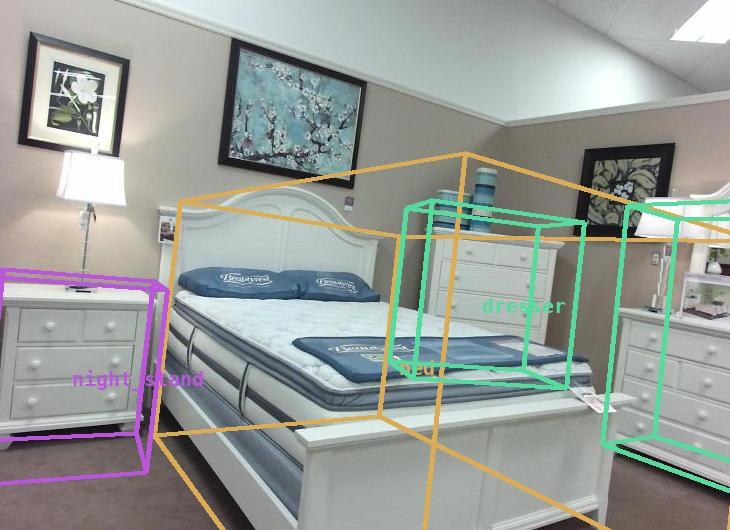}
		\includegraphics[width=\textwidth]
		{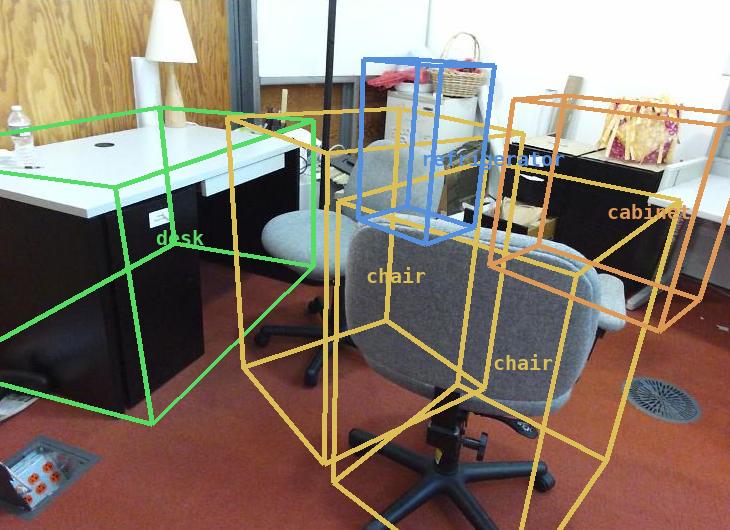}
	\end{subfigure}
	\begin{subfigure}[t]{0.15\textwidth}
		\includegraphics[width=\textwidth]  
		{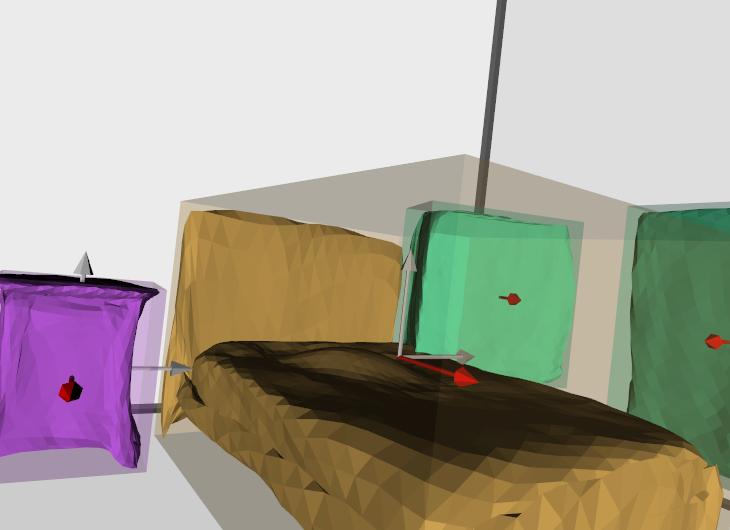}
		\includegraphics[width=\textwidth]
		{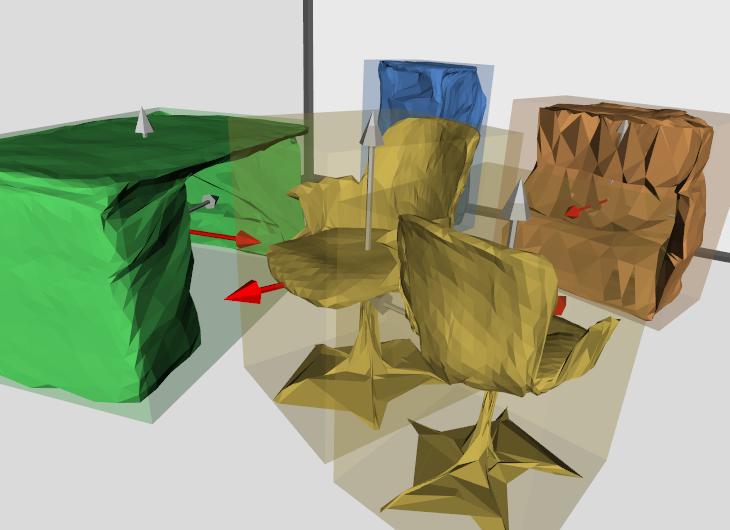}
	\end{subfigure}
	\caption{From a single image (left), we simultaneously predict the contextual knowledge including room layout, camera pose, and 3D object bounding boxes (middle) and reconstruct object meshes (right).}
	\label{fig:intro_figure}
\end{figure}

Previous works have attempted to address it via various approaches. \textbf{Scene understanding} methods \cite{schwing2013box,huang2018cooperative,choi2013understanding} obtain room layout and 3D bounding boxes of indoor objects without shape details. \textbf{Scene-level reconstruction} methods recover object shapes using contextual knowledge (room layout and object locations) for scene reconstruction, but most methods currently adopt depth or voxel representations \cite{shin20193d,li2019silhouette,tulsiani2018factoring,kulkarni20193d}. Voxel-grid presents better shape description than boxes, but its resolution is still limited, and the improvement of voxel quality exponentially increases the computational cost, which is more obvious in scene-level reconstruction. \textbf{Mesh-retrieval} methods \cite{izadinia2017im2cad,huang2018holistic,hueting2017seethrough} improve the shape quality in scene reconstruction using a 3D model retrieval module. As these approaches require iterations of rendering or model search, the mesh similarity and time efficiency depend on the size of the model repository and raise further concerns. \textbf{Object-wise mesh reconstruction} exhibits the advantages in both efficiency and accuracy \cite{wang2018pixel2mesh,groueix2018,Junyi,kato2018neural, gkioxari2019mesh}, where the target mesh is end-to-end predicted in its own object-centric coordinate system. For scene-level mesh reconstruction, predicting objects as isolated instances may not produce ideal results given the challenges of object alignment, occlusion relations and miscellaneous image background. Although Mesh R-CNN \cite{gkioxari2019mesh} is capable of predicting meshes for multiple objects from an image, its object-wise approach still ignores scene understanding and suffers from the artifacts of mesh generation on cubified voxels. So far, to the best of authors' knowledge, few works take into account both mesh reconstruction and scene context (room layout, camera pose and object locations) for total 3D scene understanding. 

To bridge the gap between scene understanding and object mesh reconstruction, we unify them together with joint learning, and simultaneously predict room layout, camera pose, 3D object bounding boxes and meshes (Figure~\ref{fig:intro_figure}). The insight is that object meshes in a scene manifest spatial occupancy that could help 3D object detection, and the 3D detection provides with object alignment that enables object-centric reconstruction at the instance-level. Unlike voxel grids, coordinates of reconstructed meshes are differentiable, thus enabling the joint training by comparing the output mesh with the scene point cloud (e.g. on SUN RGB-D \cite{song2015sun}). With the above settings, we observe that the performance on scene understanding and mesh reconstruction can make further progress and reach the state-of-the-art on the SUN RGB-D \cite{song2015sun} and Pix3D \cite{sun2018pix3d} datasets. In summary, we list our contributions as follows:
\begin{itemize}
	\item We provide a solution to automatically reconstruct room layout, object bounding boxes, and meshes from a single image. To our best knowledge, it is the first work of end-to-end learning for comprehensive 3D scene understanding with mesh reconstruction at the instance level. This integrative approach shows the complementary role of each component and reaches the state-of-the-art on each task.
	\item We propose a novel density-aware topology modifier in object mesh generation. It prunes mesh edges based on local density to approximate the target shape by progressively modifying mesh topology. Our method directly tackles the major bottleneck of \cite{Junyi}, which is in the requirement of a strict distance threshold to remove detached faces from the target shape. Compared with \cite{Junyi}, our method is robust to diverse shapes of indoor objects under complex backgrounds.
	\item Our method takes into account the attention mechanism and multilateral relations between objects. In 3D object detection, the object pose has an implicit and multilateral relation with surroundings, especially in indoor rooms (e.g., bed, nightstand, and lamp). Our strategy extracts the latent features for better deciding object locations and poses, and improves 3D detection.
\end{itemize}

\section{Related Work}
Single-view scene reconstruction presents a challenging task in computer vision and graphics since the first work \cite{roberts1963machine} in shape inference from a single photo. For indoor scene reconstruction, the difficulties increase with the complexity of clutter, occlusion and object diversity, etc. 

Early works only focus on room layout estimation \cite{hedau2009recovering, lee2009geometric, mallya2015learning, dasgupta2016delay, ren2016coarse} to represent rooms with a bounding box. With the advance of CNNs, more methods are developed to estimate object poses beyond the layout \cite{du2018learning, huang2018cooperative, chen2019holistic++}. Still, these methods are limited to the prediction of the 3D bounding box of each furniture. To recover object shapes, some methods \cite{izadinia2017im2cad,hueting2017seethrough,huang2018holistic} adopt shape retrieval approach to search for appearance-similar models from a dataset. However, its accuracy and time efficiency directly depend on the size and diversity of the dataset.

Scene reconstruction at the instance level remains problematic because of the large number of indoor objects with various categories. It leads to a high-dimensional latent space of object shapes subjected to diverse geometry and topology.
To first address single object reconstruction, some approaches represent shapes in the form of point cloud \cite{fan2017point,mandikal20183d,kurenkov2018deformnet,navaneet2019capnet}, patches \cite{groueix2018,wang2018adaptive} and primitives \cite{tian2019learning,tulsiani2017learning,paschalidou2019superquadrics,deprelle2019learning} which are adaptable to complex topology but require post-processing to obtain meshes. The structure of the voxel grid \cite{3D-R2N2,LiaoDG18,Wallace} is regular while suffering from the balance between resolution and efficiency, demanding the use of Octree to improve local details \cite{riegler2017octnet,tatarchenko2017octree,wang2018adaptive}. Some methods produce impressive mesh results using the form of signed distance fields \cite{park2019deepsdf} and implicit surfaces \cite{chen2019learning,michalkiewicz2019deep,xu2019disn,mescheder2019occupancy}. However, these methods are time-consuming and computationally intensive, making it impractical to reconstruct all objects in a scene. Another popular approach is to reconstruct meshes from a template \cite{wang2018pixel2mesh,groueix2018,kato2018neural}, but the topology of the reconstructed mesh is restricted. So far, the state-of-art approaches modify the mesh topology to approximate the ground-truth \cite{Junyi,tang2019skeleton}. However, existing methods estimate 3D shapes in the object-centric system, which cannot be applied to scene reconstruction directly.

The most relevant works to us are \cite{li2019silhouette,tulsiani2018factoring,kulkarni20193d,gkioxari2019mesh}, which take a single image as input and reconstruct multiple object shapes in a scene. However, the methods \cite{li2019silhouette,tulsiani2018factoring,kulkarni20193d} are designed for voxel reconstruction with limited resolution. Mesh R-CNN \cite{gkioxari2019mesh} produces object meshes, but still treats objects as isolated geometries without considering the scene context (room layout, object locations, etc.). Mesh R-CNN uses cubified voxels as an intermediate representation and suffers from the problem of limited resolution. Different from the above works, our method connects the object-centric reconstruction with 3D scene understanding, enabling joint learning of room layout, camera pose, object bounding boxes, and meshes from a single image.
\begin{figure*}
	\centering
	\begin{subfigure}[t]{0.55\textwidth}
		\includegraphics[width=\textwidth]  
		{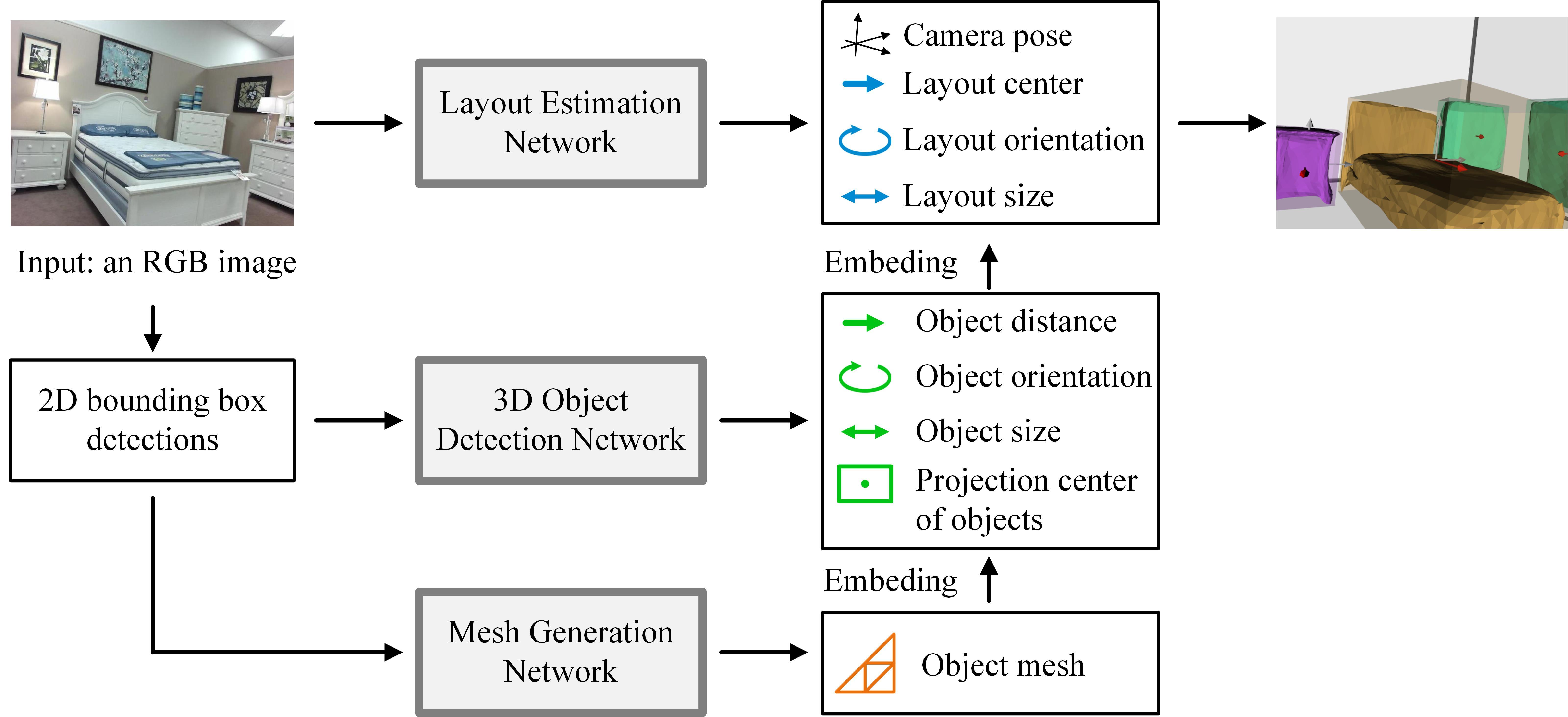}
		\caption{Architecture of the scene reconstruction network}
		\label{fig:arch}
	\end{subfigure}
	\rulesep
	\begin{subfigure}[t]{0.40\textwidth}
		\includegraphics[width=\textwidth]  
		{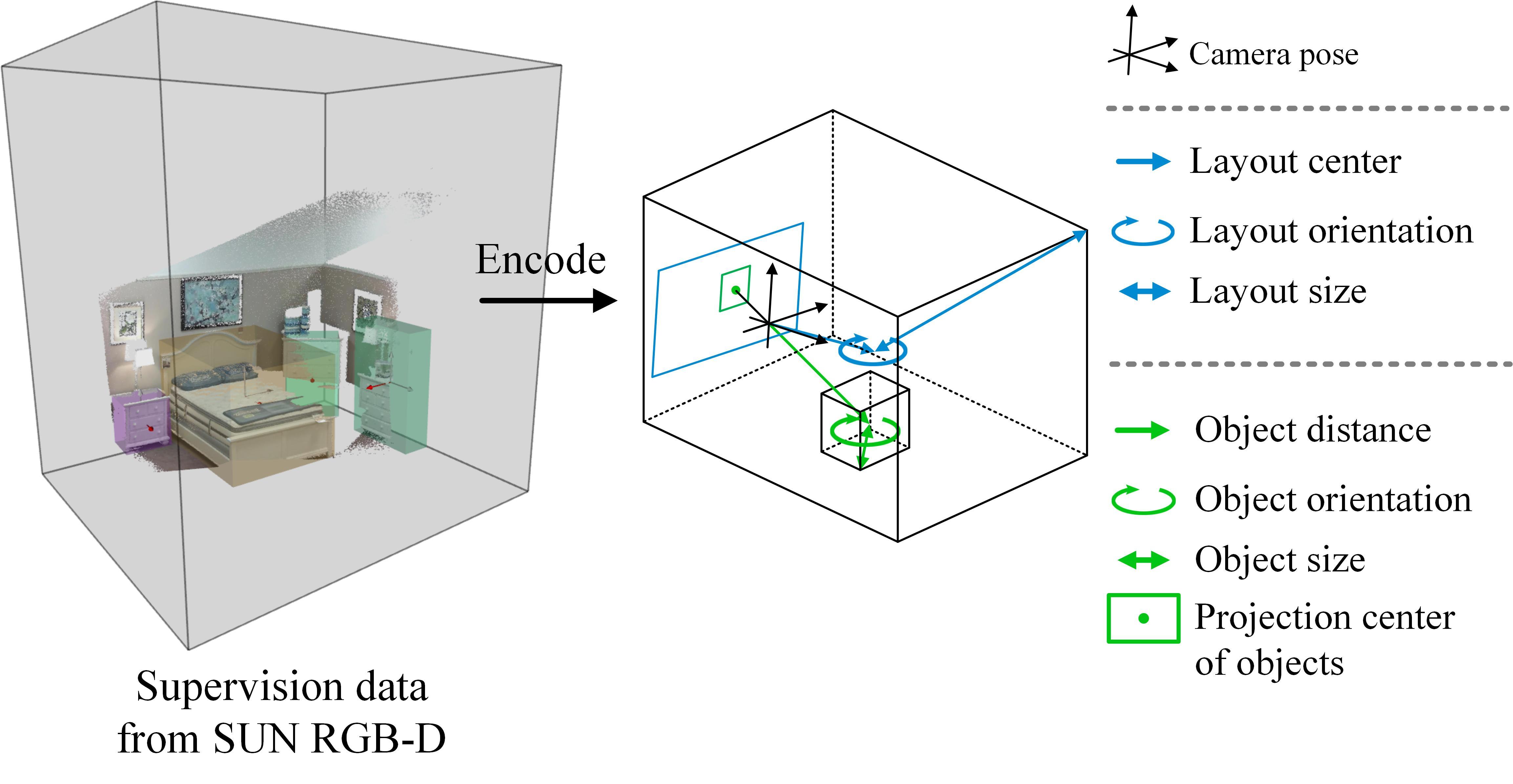}
		\caption{Parameterization of the learning targets}
		\label{fig:param}
	\end{subfigure}
	\caption{Overview of our approach. (a) The hierarchy of our method follows a `box-in-the-box' manner using three modules: the Layout Estimation Network (LEN), 3D Object Detection Network (ODN) and Mesh Generation Network (MGN). A full scene mesh is reconstructed by embedding them together with joint inference. (b) The parameterization of our learning targets in LEN and ODN \cite{huang2018cooperative}.}
	\label{fig:overview}
\end{figure*}

\section{Method}
We illustrate the overview of our method in Figure~\ref{fig:arch}. The network architecture follows a `box-in-the-box' manner and consists of three modules: 1. Layout Estimation Network (LEN); 2. 3D Object Detection Network (ODN); 3. Mesh Generation Network (MGN). 
From a single image, we first predict 2D object bounding boxes with Faster R-CNN \cite{ren2015faster}.
LEN takes the full image as input and produces the camera pose and the layout bounding box. 
Given the 2D detection of objects, ODN detects the 3D object bounding boxes in the camera system, while MGN generates the mesh geometry in their object-centric system.  
We reconstruct the full-scene mesh by embedding the outputs of all networks together with joint training and inference, where object meshes from MGN are scaled and placed into their bounding boxes (by ODN) and transformed into the world system with the camera pose (by LEN). We elaborate on the details of each network in this section.

\subsection{3D Object Detection and Layout Estimation}
\label{sec:3DD}
To make the bounding box of layout and objects learnable, we parameterize a box as the prior work \cite{huang2018cooperative} (Figure~\ref{fig:param}). We set up the world system located at the camera center with its vertical (y-) axis perpendicular to the floor, and its forward (x-) axis toward the camera, such that the camera pose $\mathbf{R}\left(\beta,\gamma\right)$ can be decided by the pitch and roll angles $\left(\beta,\gamma\right)$.
In the world system, a box can be determined by a 3D center $\bm{C} \in \mathbb{R}^{3}$, spatial size $\bm{s}\in \mathbb{R}^3$, orientation angle $\theta \in \left[-\pi, \pi\right)$. For indoor objects, the 3D center $\bm{C}$ is represented by its 2D projection $\bm{c} \in \mathbb{R}^{2}$ on the image plane with its distance $d \in \mathbb{R}$ to the camera center. Given the camera intrinsic matrix $\mathbf{K} \in \mathbb{R}^{3}$, $\bm{C}$ can be formulated by:
\begin{equation}
\label{eqn:01}
\begin{aligned}
\bm{C} &= \mathbf{R}^{-1}\left(\beta,\gamma\right)\cdot d \cdot \frac{\mathbf{K}^{-1}\left[\bm{c}, 1\right]^{T}}{\|\mathbf{K}^{-1}\left[\bm{c}, 1\right]^{T}\|_{2}}.
\end{aligned}
\end{equation}
The 2D projection center $\bm{c}$ can be further decoupled by $\bm{c}^{b} + \bm{\delta}$. $\bm{c}^{b}$ is the 2D bounding box center and $\bm{\delta} \in \mathbb{R}^{2}$ is the offset to be learned. From the 2D detection $\bm{I}$ to its 3D bounding box corners, the network can be represented as a function by $\mathbf{F}\left(\bm{I} | \bm{\delta},d,\beta,\gamma,\bm{s},\theta\right) \in \mathbb{R}^{3\times8}$. The ODN estimates the box property $\left(\bm{\delta},d,\bm{s},\theta\right)$ of each object, and the LEN decides the camera pose $\mathbf{R}\left(\beta,\gamma\right)$ with the layout box $\left(\bm{C},\bm{s}^{l},\theta^{l}\right)$.

\begin{figure}[!ht]
	\centering
	\includegraphics[width=\linewidth]{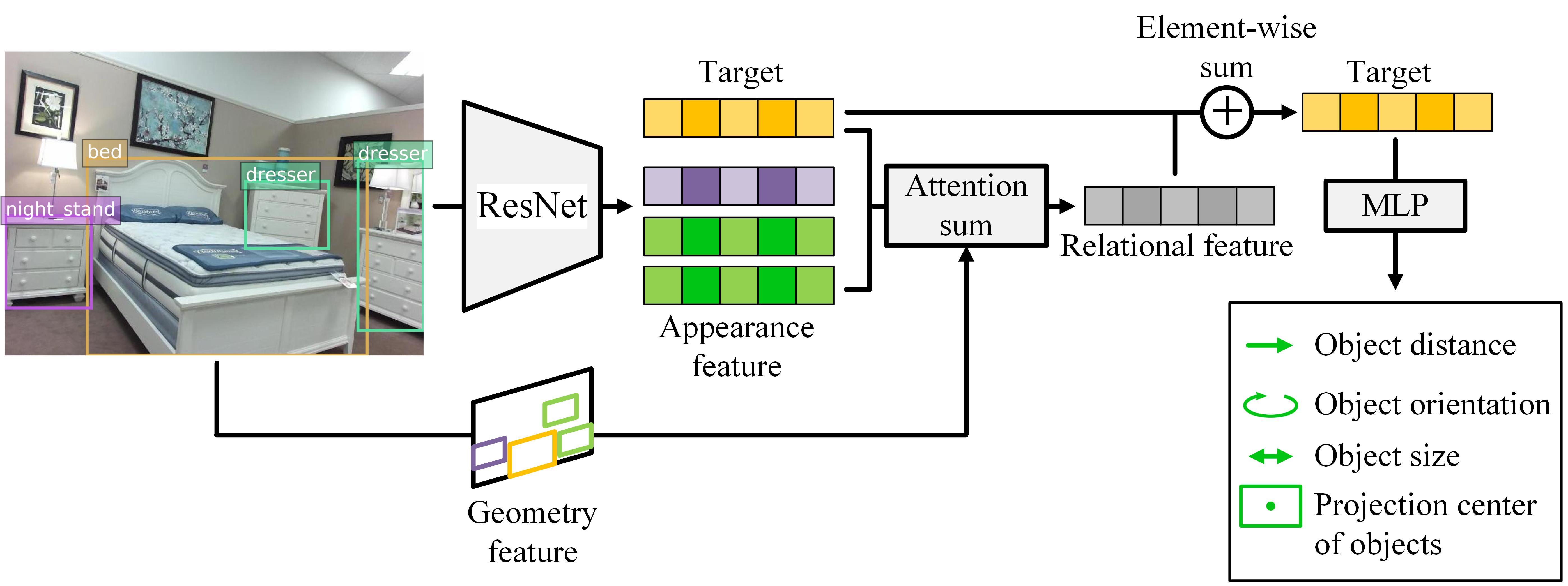}
	\caption{3D Object Detection Network (ODN)}
	\label{fig:ODN}
\end{figure}

\noindent\textbf{Object Detection Network (ODN).} In indoor environments, object poses generally follow a set of interior design principles, making it a latent pattern that can be learned. By parsing images, previous works either predict 3D boxes object-wisely \cite{huang2018cooperative,tulsiani2018factoring} or only consider pair-wise relations \cite{kulkarni20193d}. In our work, we assume each object has a $\textit{multi-lateral relation}$ between its surroundings, and take all in-room objects into account in predicting its bounding box. The network is illustrated in Figure~\ref{fig:ODN}. Our method is inspired by the consistent improvement of attention mechanism in 2D object detection \cite{hu2018relation}. For 3D detection, we first object-wisely extract the appearance feature with ResNet-34 \cite{he2016deep} from 2D detections, and encode the relative position and size between 2D object boxes into geometry feature with the method in \cite{hu2018relation,vaswani2017attention}. For each target object, we calculate its \textbf{relational feature} to the others with the object relation module \cite{hu2018relation}. It adopts a piece-wise feature summation weighted by the similarity in appearance and geometry from the target to the others, which we call `\textbf{attention sum}' in Figure~\ref{fig:ODN}. We then element-wisely add the relational feature to the target and regress each box parameter in $\left(\bm{\delta},d,\bm{s},\theta\right)$ with a two-layer MLP. For indoor reconstruction, the object relation module reflects the inherent significance in the physical world: objects generally have stronger relations with the others which are neighboring or appearance-similar. We demonstrate its effectiveness in improving 3D object detection in our ablation analysis.

\noindent\textbf{Layout Estimation Network (LEN).} The LEN predicts the camera pose $\mathbf{R}\left(\beta,\gamma\right)$ and its 3D box $\left(\bm{C},\bm{s}^{l},\theta^{l}\right)$ in the world system. In this part, we employ the same architecture as ODN but remove the relational feature. $\left(\beta,\gamma,\bm{C},\bm{s}^{l},\theta^{l}\right)$ are regressed with two fully-connected layers for each target after the ResNet. Similar to \cite{huang2018cooperative}, the 3D center $\bm{C}$ is predicted by learning an offset to the average layout center.

\subsection{Mesh Generation for Indoor Objects}
\label{sec:MGN}
Our Mesh Generation Network directly tackles the major issue with one recent work, Topology Modification Network (TMN) \cite{Junyi}: TMN approximates object shapes by deforming and modifying the mesh topology, where a predefined distance threshold is required to remove detached faces from the target shape. However, it is nontrivial to give a general threshold for different scales of object meshes (see Figure~\ref{fig:tmn_0.1}). One possible reason is that indoor objects have a large shape variance among different categories. Another one is that complex backgrounds and occlusions often cause the failure of estimating a precise distance value.

\begin{figure}[!ht]
	\centering
	\includegraphics[width=\linewidth]{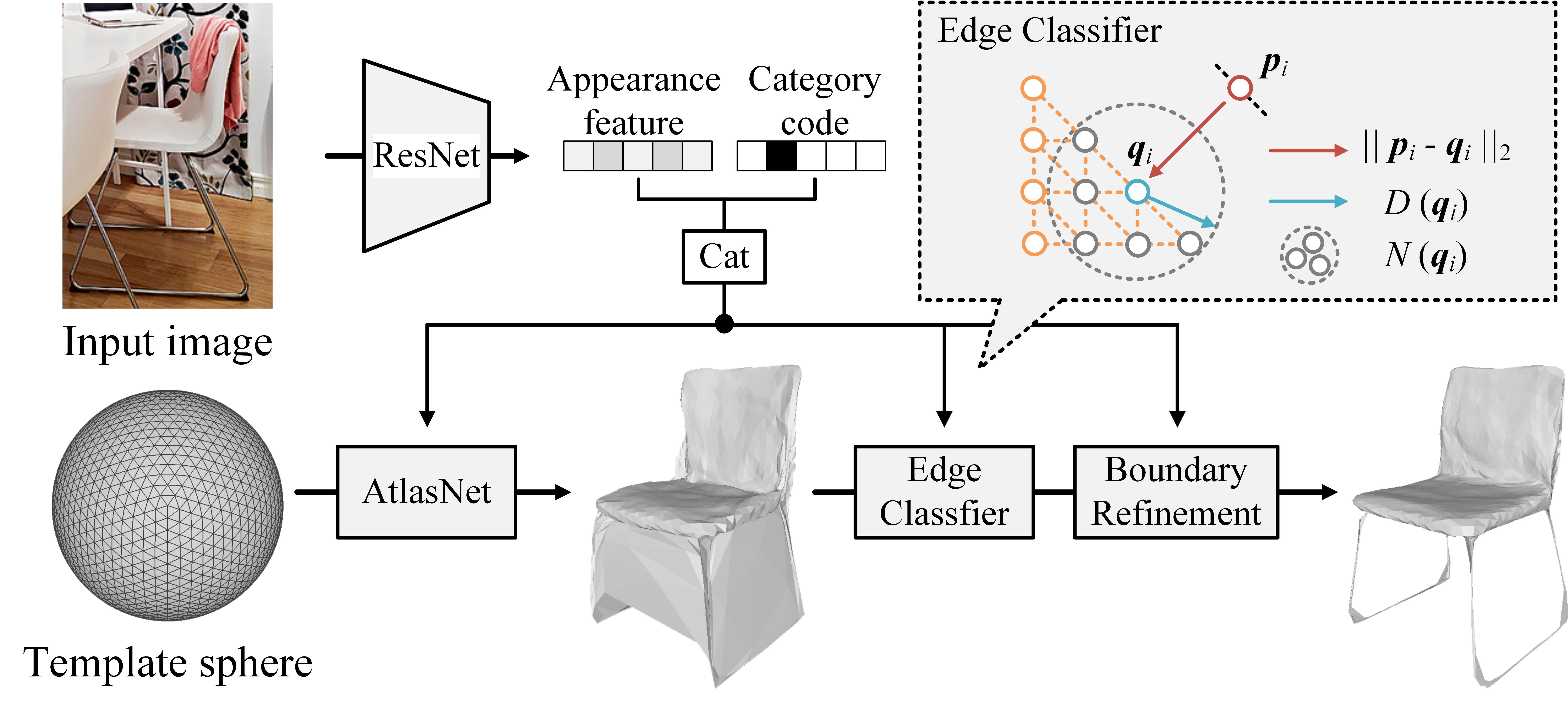}
	\caption{Mesh Generation Network (MGN). Our method takes as input a detected object which is vulnerable to occlusions, and outputs a plausible mesh.}
	\label{fig:MGN}
\end{figure}

\noindent\textbf{Density v.s. Distance.} Different from TMN where a strict distance threshold is used for topology modification, we argue that whether to reserve a face or not should be determined by its local geometry. In this part, we propose an adaptive manner that modifies meshes based on the local \textbf{density} of the ground-truth. We set $\bm{p}_{i} \in \mathbb{R}^{3}$ as a point on our reconstructed mesh, and $\bm{q}_{i} \in \mathbb{R}^{3}$ corresponds to its nearest neighbor on the ground-truth (see Figure~\ref{fig:MGN}). We design a binary classifier $f\left(*\right)$ to predict whether $\bm{p}_{i}$ is close to the ground-truth mesh in Equation~\ref{eqn:02}:
\begin{equation}
\label{eqn:02}
\begin{aligned}
f(\bm{p}_{i}) & =
\begin{cases}
\text{False}& \|\bm{p}_{i}-\bm{q}_{i}\|_{2} > D\left(\bm{q}_{i}\right)\\
\text{True}& \text{otherwise}
\end{cases}\\
D\left(\bm{q}_{i}\right) & = \underset{\bm{q}_{m},\bm{q}_{n} \in N\left(\bm{q}_{i}\right)}{\text{max}\ \text{min}} \|\bm{q}_{m}-\bm{q}_{n}\|_{2}, m \neq n
\end{aligned},
\end{equation}
where $N\left(\bm{q}_{i}\right)$ are the neighbors of $\bm{q}_{i}$ on the ground-truth mesh, and $D\left(\bm{q}_{i}\right)$ is defined as its local density. This classifier is designed by our insight that: in shape approximation, a point should be reserved if it belongs to the neighbors $N\left(*\right)$ of the ground-truth. We also observe that this classifier shows better robustness with different mesh scales than using a distance threshold (see Figure~\ref{fig:obj_compare}).

\noindent\textbf{Edges v.s. Faces.} Instead of removing faces, we choose to cut mesh edges for topology modification. We randomly sample points on mesh edges and use the classifier $f\left(*\right)$ to cut edges on which the average classification score is low. It is from the consideration that cutting false edges can reduce incorrect connections penalized by the edge loss \cite{wang2018pixel2mesh} and create compact mesh boundaries.

\noindent\textbf{Mesh Generation Network.} We illustrate our network architecture in Figure~\ref{fig:MGN}. It takes a 2D detection as input and uses ResNet-18 to produce image features. We encode the detected object category into a one-hot vector and concatenate it with the image feature. It is from our observation that the category code provides shape priors and helps to approximate the target shape faster. The augmented feature vector and a template sphere are fed into the decoder in AtlasNet \cite{groueix2018} to predict deformation displacement on the sphere and output a plausible shape with unchanged topology. The edge classifier has the same architecture with the shape decoder, where the last layer is replaced with a fully connected layer for classification. It shares the image feature, takes the deformed mesh as input and predicts the $f\left(*\right)$ to remove redundant meshes.
We then append our network with a boundary refinement module \cite{Junyi} to refine the smoothness of boundary edges and output the final mesh.

\subsection{Joint Learning for Total 3D Understanding}
\label{sec:joint}
In this section, we conclude the learning targets with the corresponding loss functions, and describe our joint loss for end-to-end training.

\noindent\textbf{Individual losses.} ODN predicts $\left(\bm{\delta},d,\bm{s},\theta\right)$ to recover the 3D object box in the camera system, and LEN produces $\left(\beta,\gamma,\bm{C},\bm{s}^{l},\theta^{l}\right)$ to represent the layout box, along with the camera pose to transform 3D objects into the world system. As directly regressing absolute angles or length with L2 loss is error-prone \cite{huang2018cooperative,qi2018frustum}. We keep inline with them by using the classification and regression loss $\mathcal{L}^{cls,reg} = \mathcal{L}^{cls} + \lambda_{r}\mathcal{L}^{reg}$ to optimize $\left(\theta, \theta^{l}, \beta, \gamma,d, \bm{s}, \bm{s}^{l}\right)$. We refer readers to \cite{huang2018cooperative} for details. As $\bm{C}$ and $\bm{\delta}$ are calculated by the offset from a pre-computed center, we predict them with L2 loss. For MGN, we adopt the Chamfer loss $\mathcal{L}_{c}$, edge loss $\mathcal{L}_{e}$, boundary loss $\mathcal{L}_{b}$ as \cite{groueix2018,wang2018pixel2mesh,Junyi} with our cross-entropy loss $\mathcal{L}_{ce}$ for modifying edges in mesh generation.

\noindent\textbf{Joint losses.} We define the joint loss between ODN, LEN and MGN based on two insights: 1. The camera pose estimation should improve 3D object detection, and vice versa; 2. object meshes in a scene present spatial occupancy that should benefit the 3D detection, and vice versa. For the first, we adopt the cooperative loss $\mathcal{L}_{co}$ from \cite{huang2018cooperative} to ensure the consistency between the predicted world coordinates of layout \& object boxes and the ground-truth. For the second, we require the reconstructed meshes close to their point cloud in the scene. It exhibits global constraints by aligning mesh coordinates with the ground-truth. We define the global loss as the partial Chamfer distance \cite{groueix2018}:
\begin{equation}
\label{eqn:03}
\begin{aligned}
\mathcal{L}_{g} = \frac{1}{N}\sum_{i=1}^{N}\frac{1}{|\mathbb{S}_{i}|}\sum_{\bm{q} \in \mathbb{S}_{i}} \min_{\bm{p}\in \mathbb{M}_{i}} \|\bm{p}-\bm{q}\|_{2}^{2}
\end{aligned},
\end{equation}

where $\bm{p}$ and $\bm{q}$ respectively indicate a point on a reconstructed mesh $\mathbb{M}_{i}$ and the ground-truth surface $\mathbb{S}_{i}$ of $i$-th object in the world system. $N$ is the number of objects and $|\mathbb{S}_{i}|$ denotes the point number on $\mathbb{S}_{i}$. Unlike single object meshes, real-scene point clouds are commonly coarse and partially covered (scanned with depth sensors), thus we do not use the Chamfer distance to define $\mathcal{L}_{g}$. All the loss functions in joint training can be concluded as:
\begin{equation}
\label{eqn:04}
\begin{aligned}
\mathcal{L} = & \sum_{x \in \{  \bm{\delta}, d, \bm{s}, \theta  \} }\lambda_{x}\mathcal{L}_{x} + \sum_{y \in \{ \beta,\gamma, \bm{C}, \bm{s}^{l}, \theta^{l} \}} \lambda_{y}\mathcal{L}_{y}\\
& + \sum_{z \in \{c, e, b,ce\}} \lambda_{z} \mathcal{L}_{z} + \lambda _{co}\mathcal{L}_{co} + \lambda _{g} \mathcal{L}_{g},
\end{aligned}
\end{equation}
where the first three terms represent the individual loss in ODN, LEN and MGN, and the last two are the joint terms. $\{\lambda_{*}\}$ are the weights used to balance their importance.

\section{Results and Evaluation}
\subsection{Experiment Setup}
\label{sec:exp}
\noindent \textbf{Datasets:} We use two datasets in experiments according to the types of ground-truths they provide.
1) \textbf{SUN RGB-D} dataset \cite{song2015sun} consists of 10,335 real indoor images with labeled 3D layout, object bounding boxes and coarse point cloud (depth map). 
We use the official train/test split and NYU-37 object labels \cite{silberman2012indoor} for evaluation on layout, camera pose estimation and 3D object detection. 
2) \textbf{Pix3D} dataset \cite{sun2018pix3d} contains 395 furniture models with 9 categories, which are aligned with 10,069 images. We use this for mesh reconstruction and keep the train/test split inline with \cite{gkioxari2019mesh}. The object label mapping from NYU-37 to Pix3D for scene reconstruction is listed in the supplementary file.

\noindent \textbf{Metrics:} Our results are measured on both scene understanding and mesh reconstruction metrics. We evaluate layout estimation with average 3D Intersection over Union (IoU). The camera pose is evaluated by the mean absolute error. Object detection is tested with the average precision (AP) on all object categories. We test the single-object mesh generation with the Chamfer distance as previous works \cite{gkioxari2019mesh,Junyi}, and evaluate the scene mesh with Equation \ref{eqn:03}.

\noindent \textbf{Implementation:} We train the 2D detector (Figure~\ref{fig:arch}) on the COCO dataset \cite{lin2014microsoft} first and fine-tune it on SUN RGB-D. Both ODN and LEN have the image encoder with ResNet-34 \cite{he2016deep}, and MGN is with ResNet-18. In LEN and ODN, we adopt a two-layer MLP to predict each target. In MGN, the template sphere has 2562 vertices with unit radius. We cut edges whose average classification score is lower than 0.2. Since SUN RGB-D does not provide instance meshes for 3D supervision, and Pix3D is only labeled with one object per image without layout information. We first train ODN, LEN on SUN-RGBD, and train MGN on Pix3D individually with the batch size of 32 and learning rate at 1e-3 (scaled by 0.5 for every 20 epochs, 100 epochs in total). We then combine Pix3D into SUN RGB-D to provide mesh supervision and jointly train all networks with the loss $\mathcal{L}$ in Equation~\ref{eqn:04}. Here we use one hierarchical batch (each batch contains one scene image with $N$ object images) and set the learning rate at 1e-4 (scaled by 0.5 for every 5 epochs, 20 epochs in total). We explain the full architecture, training strategies, time efficiency and parameter setting of our networks in the supplementary file.

\subsection{Qualitative Analysis and Comparison}
In this section, we evaluate the qualitative performance of our method on both object and scene levels.

\begin{figure}[!ht]
	\centering
	\begin{subfigure}[t]{0.075\textwidth}
		\includegraphics[width=\textwidth]  
		{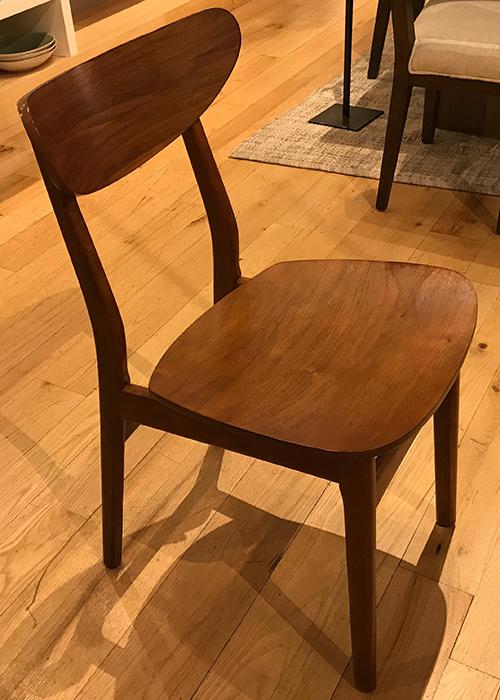}
		\includegraphics[width=\textwidth]
		{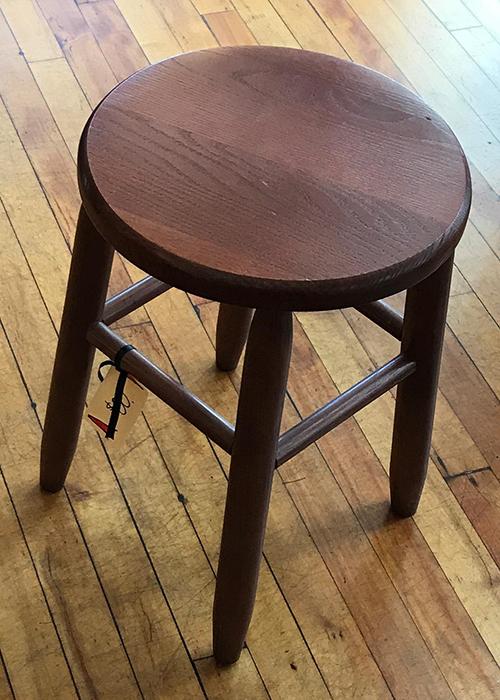}
		\includegraphics[width=\textwidth]  
		{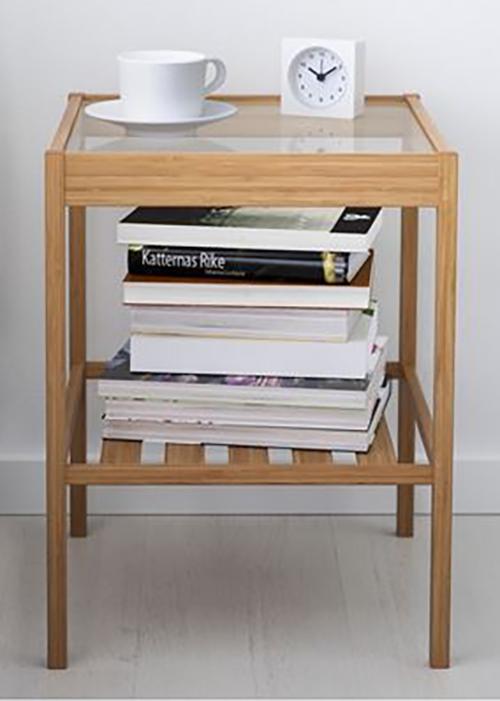}
		\includegraphics[width=\textwidth]
		{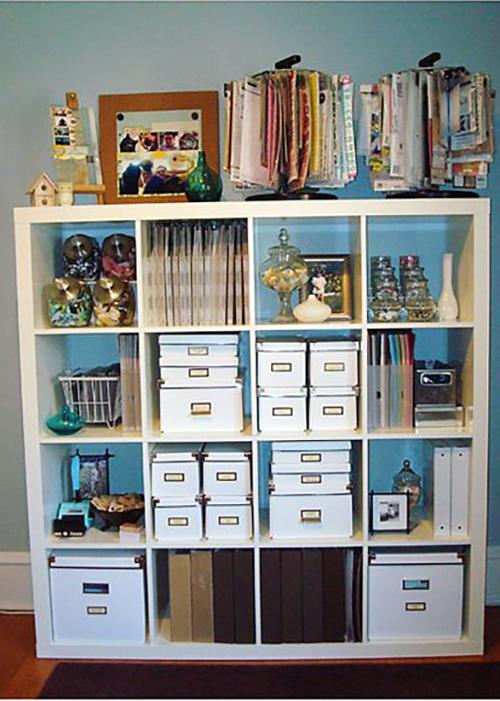}
		\includegraphics[width=\textwidth]  
		{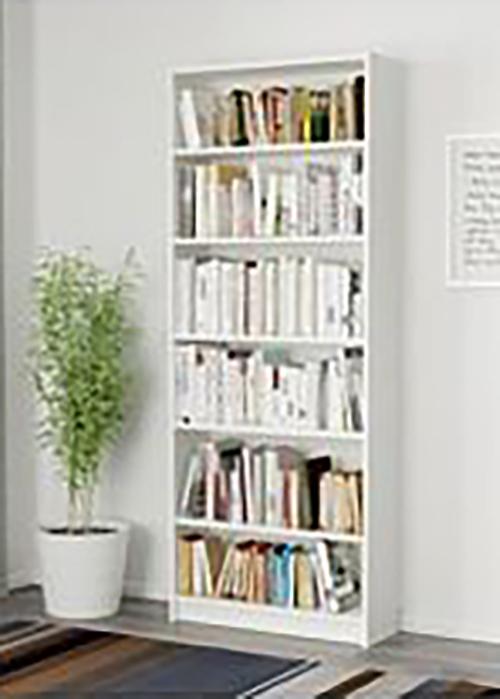}
		\includegraphics[width=\textwidth]
		{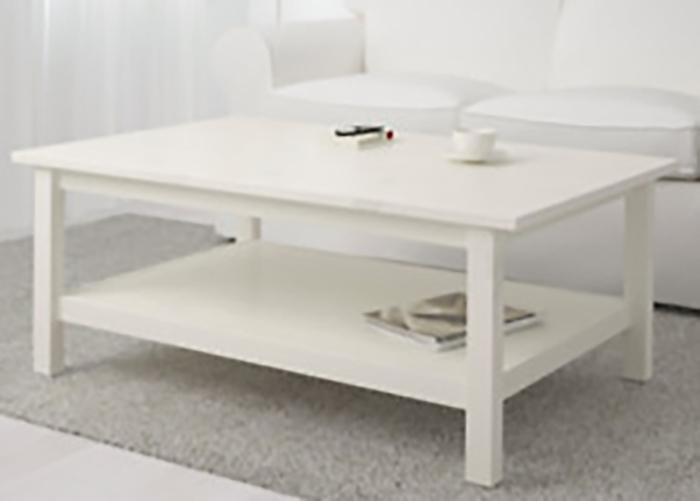}
		\caption{}
		\label{fig:obj_comp_input}
	\end{subfigure}
	\begin{subfigure}[t]{0.075\textwidth}
		\includegraphics[width=\textwidth]  
		{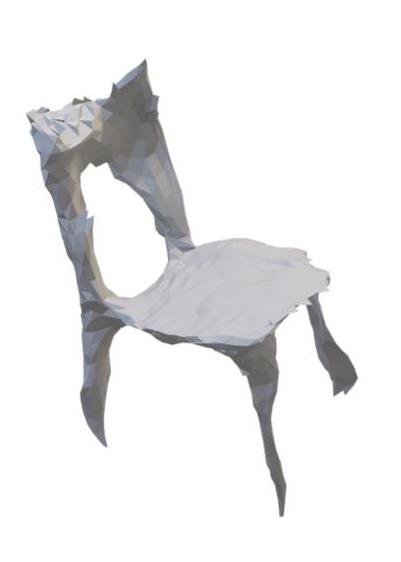}
		\includegraphics[width=\textwidth]
		{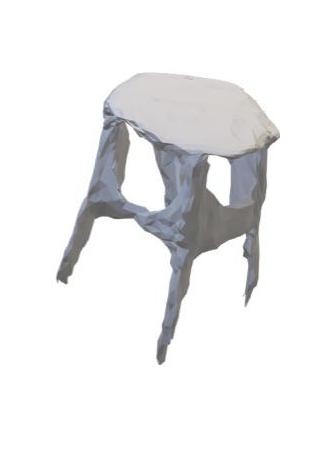}
		\includegraphics[width=\textwidth]  
		{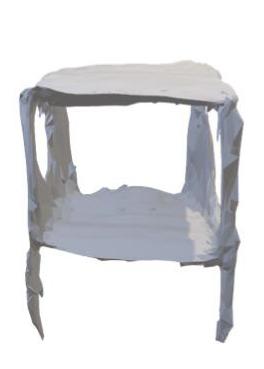}
		\includegraphics[width=\textwidth]
		{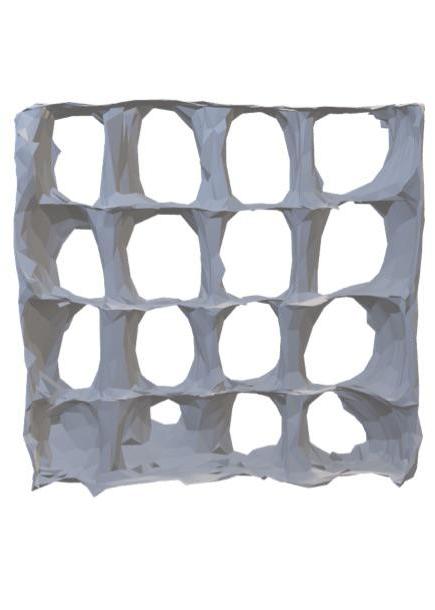}
		\includegraphics[width=\textwidth]  
		{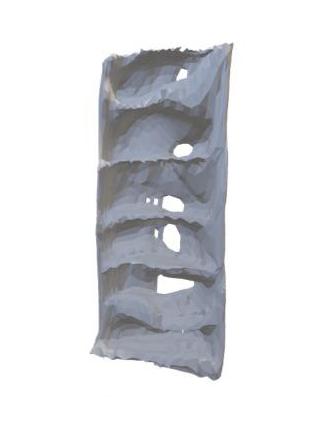}
		\includegraphics[width=\textwidth]
		{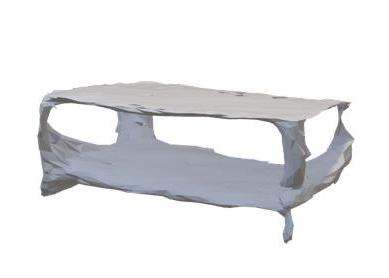}
		\caption{}
		\label{fig:obj_comp_meshrcnn}
	\end{subfigure}
	\begin{subfigure}[t]{0.075\textwidth}
		\includegraphics[width=\textwidth]  
		{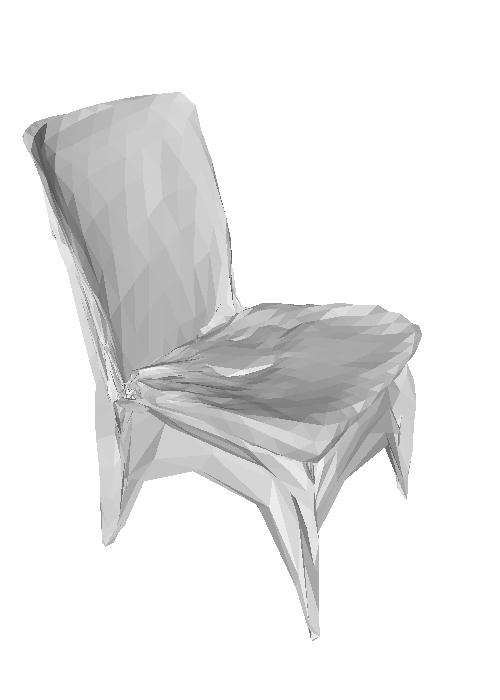}
		\includegraphics[width=\textwidth]
		{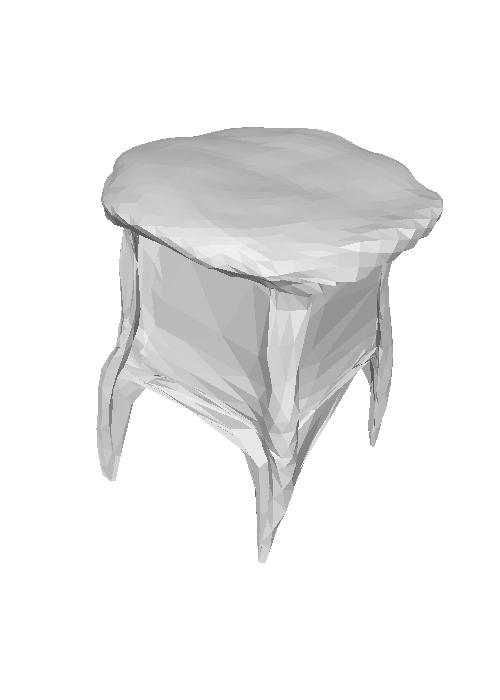}
		\includegraphics[width=\textwidth]  
		{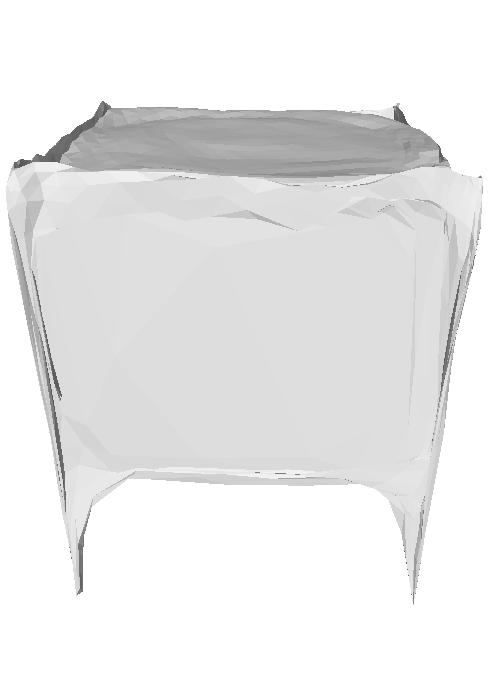}
		\includegraphics[width=\textwidth]
		{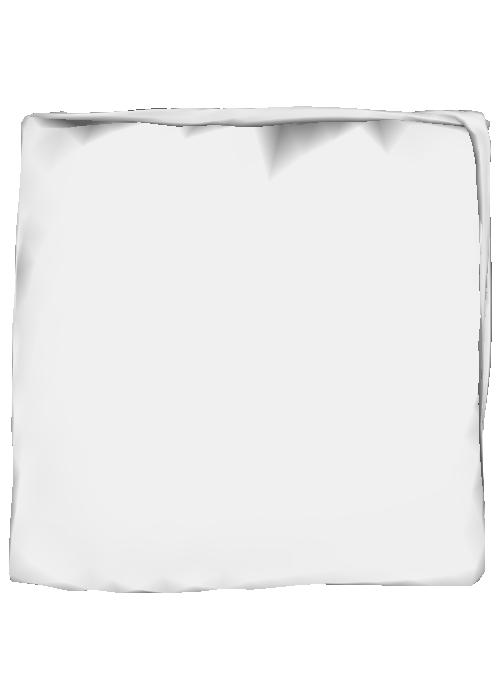}
		\includegraphics[width=\textwidth]  
		{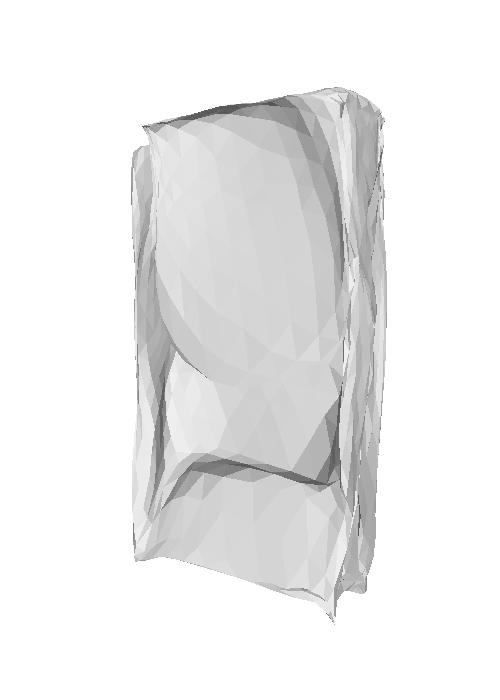}
		\includegraphics[width=\textwidth]
		{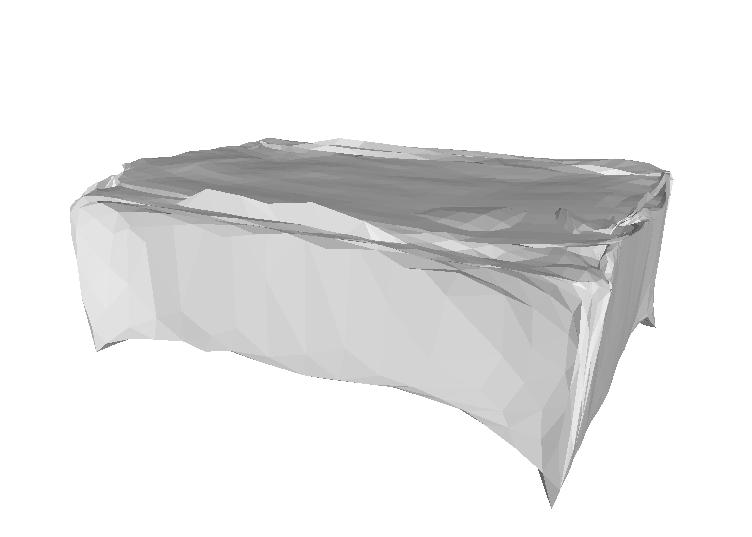}
		\caption{}
	\end{subfigure}
	\begin{subfigure}[t]{0.075\textwidth}
		\includegraphics[width=\textwidth]  
		{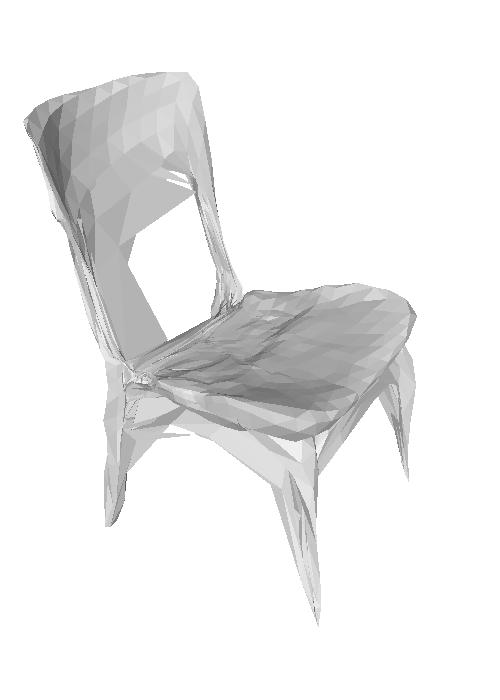}
		\includegraphics[width=\textwidth]
		{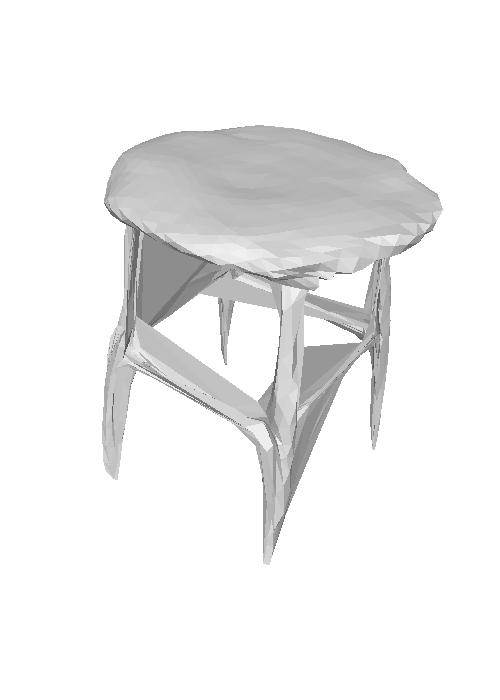}
		\includegraphics[width=\textwidth]  
		{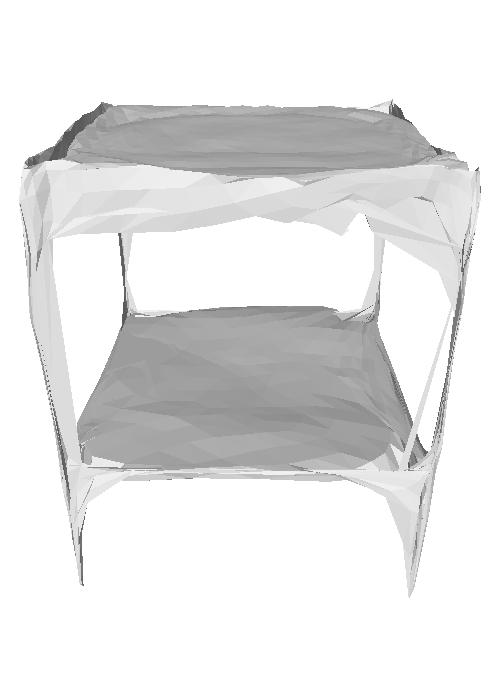}
		\includegraphics[width=\textwidth]
		{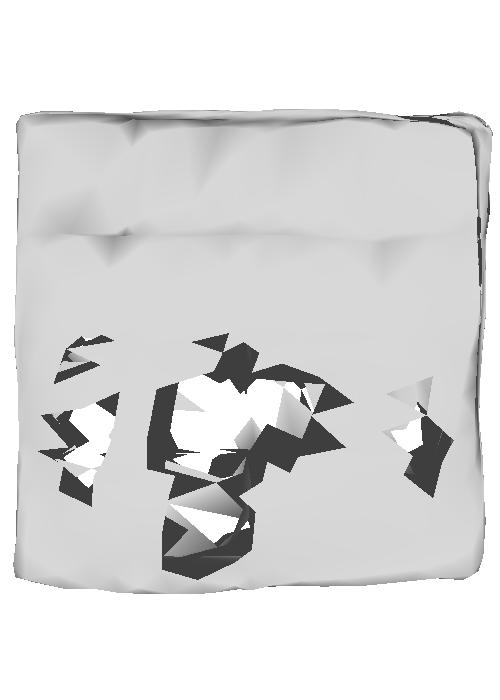}
		\includegraphics[width=\textwidth]  
		{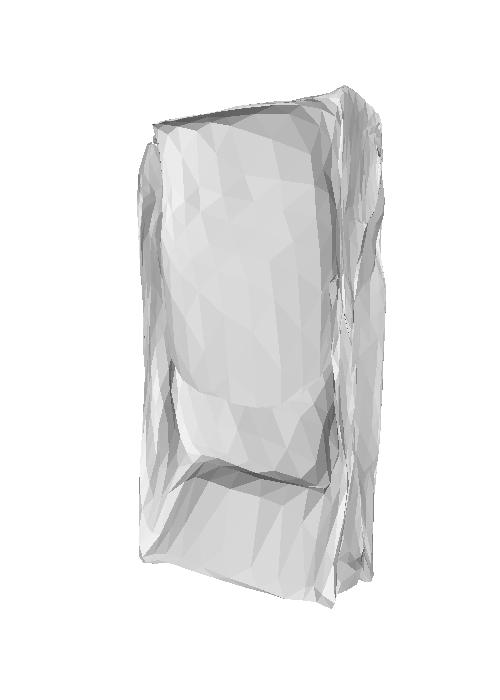}
		\includegraphics[width=\textwidth]
		{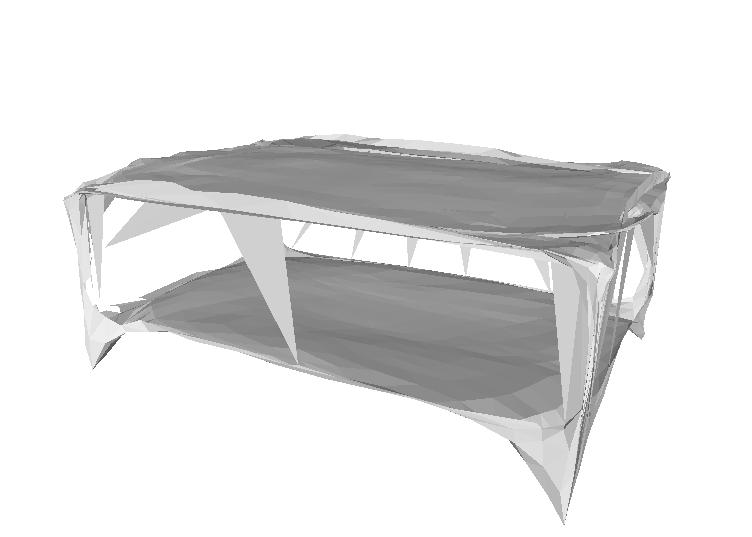}
		\caption{}
	\end{subfigure}
	\begin{subfigure}[t]{0.075\textwidth}
		\includegraphics[width=\textwidth]  
		{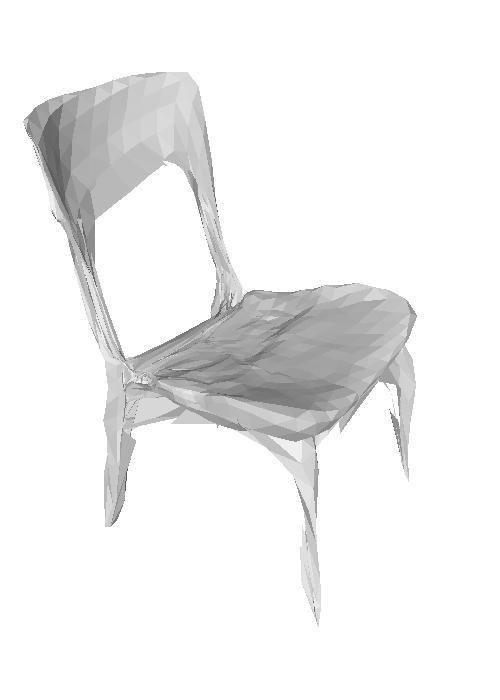}
		\includegraphics[width=\textwidth]
		{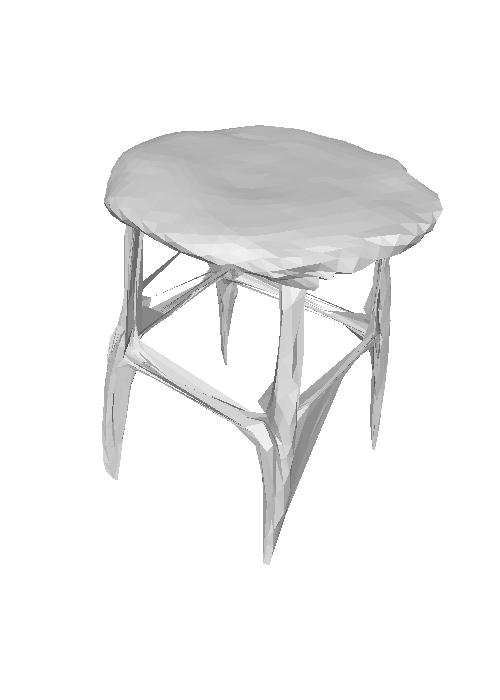}
		\includegraphics[width=\textwidth]  
		{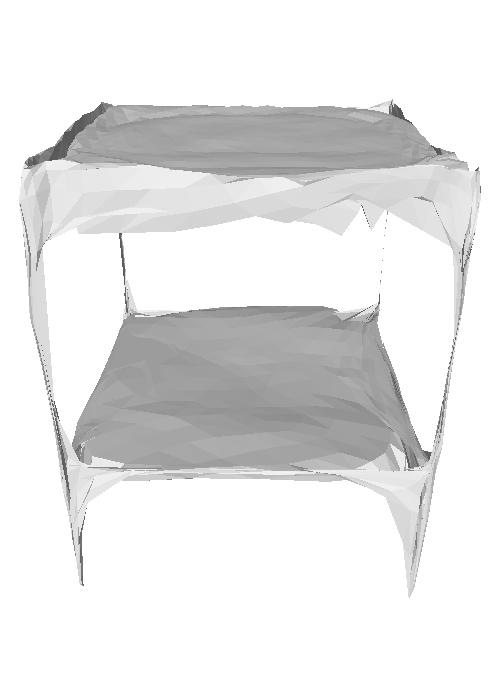}
		\includegraphics[width=\textwidth]
		{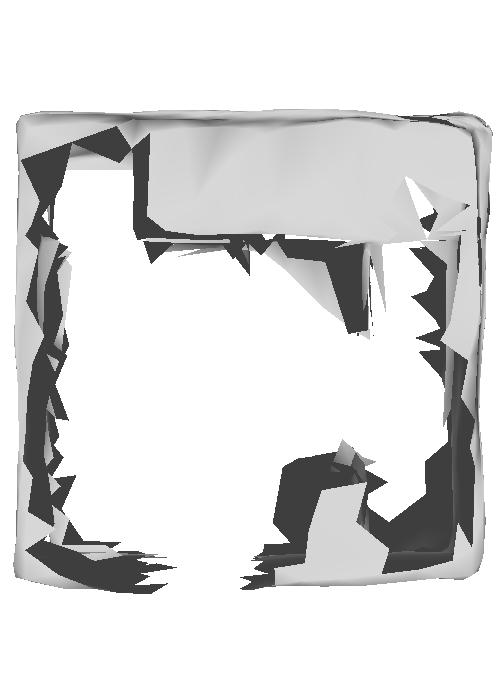}
		\includegraphics[width=\textwidth]  
		{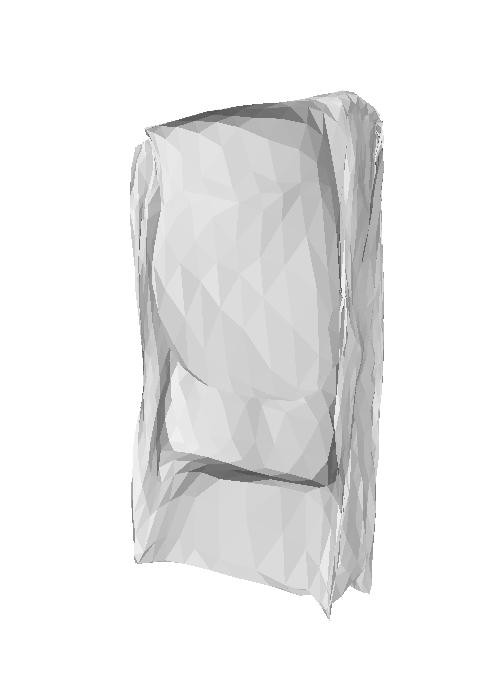}
		\includegraphics[width=\textwidth]
		{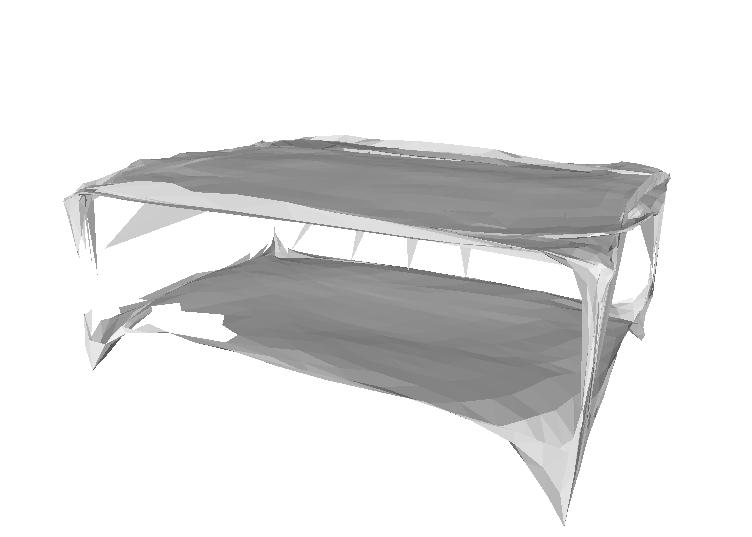}
		\caption{}
		\label{fig:tmn_0.1}
	\end{subfigure}
	\begin{subfigure}[t]{0.075\textwidth}
		\includegraphics[width=\textwidth]  
		{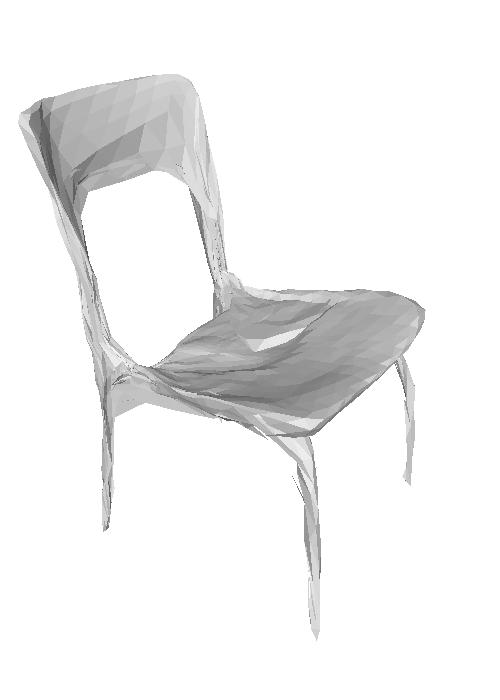}
		\includegraphics[width=\textwidth]
		{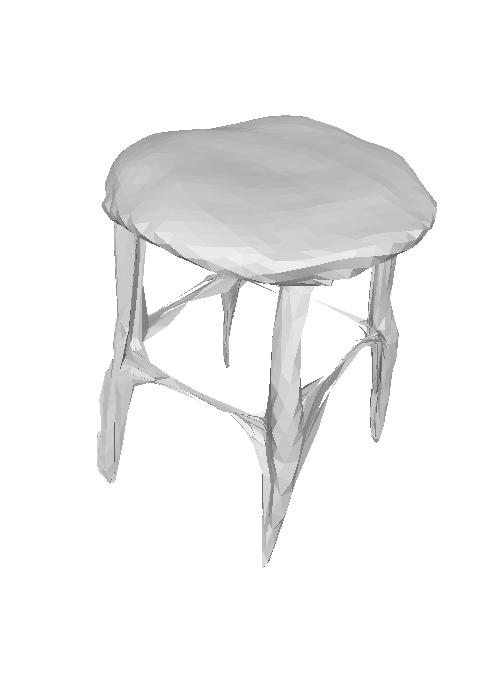}
		\includegraphics[width=\textwidth]  
		{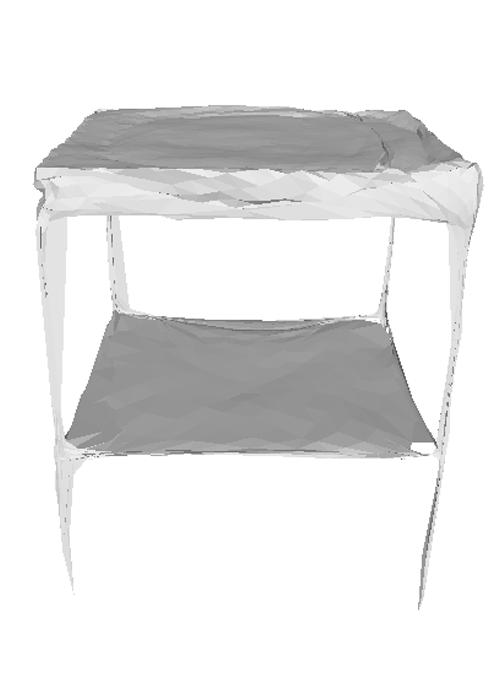}
		\includegraphics[width=\textwidth]
		{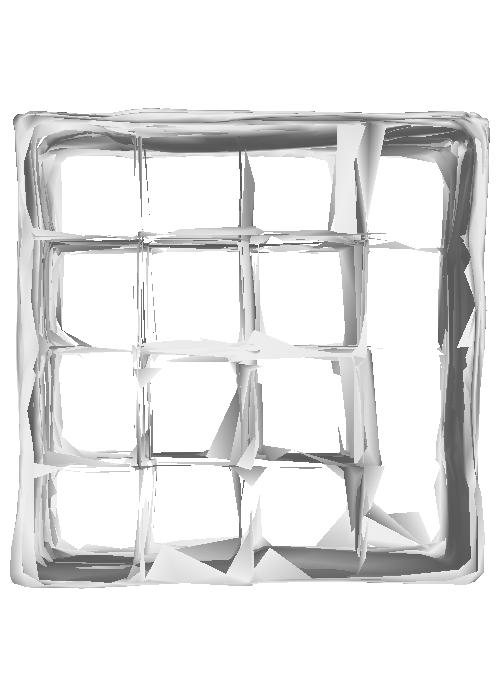}
		\includegraphics[width=\textwidth]  
		{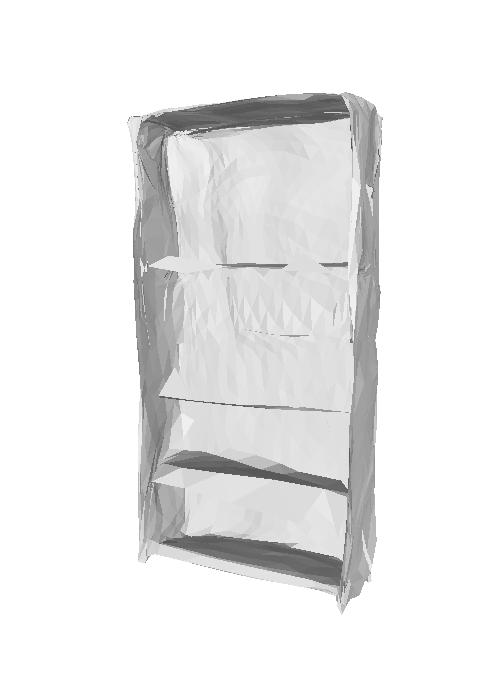}
		\includegraphics[width=\textwidth]
		{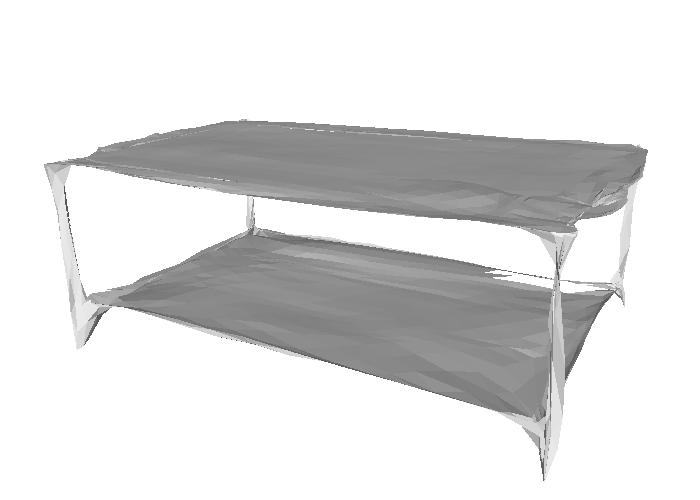}
		\caption{}
		\label{fig:obj_comp_ours}
	\end{subfigure}
	\caption{Mesh reconstruction for individual objects. From left to right: (a) Input images and results from (b) Mesh R-CNN \cite{gkioxari2019mesh}, (c) AtlasNet-Sphere \cite{groueix2018}, (d, e) TMN with $\tau=0.1$ and $\tau=0.05$ \cite{Junyi}, (f) Ours.}
	\label{fig:obj_compare}
\end{figure}

\noindent \textbf{Object Reconstruction:} We compare our MGN with the state-of-the-art mesh prediction methods \cite{gkioxari2019mesh,groueix2018,Junyi} on Pix3D. 
Because our method is designed to accomplish scene reconstruction in real scenes, we train all methods inputted with object images but without masks. For AtlasNet \cite{groueix2018} and Topology Modification Network (TMN) \cite{Junyi}, we also encode the object category into image features enabling a fair comparison. Both TMN and our method are trained following a `deformation+modification+refinement' process (see \cite{Junyi}). For Mesh R-CNN \cite{gkioxari2019mesh}, it involves an object recognition phase, and we directly compare with the results reported in their paper. The comparisons are illustrated in Figure~\ref{fig:obj_compare}, from which we observe that reconstruction from real images is challenging. Indoor furniture are often overlaid with miscellaneous objects (such as books on the shelf). From the results of Mesh R-CNN (Figure~\ref{fig:obj_comp_meshrcnn}), it generates meshes from low-resolution voxel grids ($24^{3}$ voxels) and thus results in noticeable artifacts on mesh surfaces. TMN improves from AtlasNet and refines shape topology. However, its distance threshold $\tau$ does not show consistent adaptability for all shapes in indoor environments (e.g. the stool and the bookcase in Figure~\ref{fig:tmn_0.1}). Our method relies on the edge classifier. It cuts edges depending on the local density, making the topology modification adaptive to different scales of shapes among various object categories (Figure~\ref{fig:obj_comp_ours}). The results also demonstrate that our method keeps better boundary smoothness and details.

\noindent \textbf{Scene Reconstruction:} As this is the first work, to our best knowledge, of combing scene understanding and mesh generation for full scene reconstruction, we illustrate our results on the testing set of SUN RGB-D in Figure~\ref{fig:scene_recon}
(see all samples in the supplementary file). Note that SUN RGB-D does not contain ground-truth object meshes for training. We present the results under different scene types and diverse complexities to test the robustness of our method. The first row in Figure~\ref{fig:scene_recon} shows the scenes with large repetitions and occlusions. We exhibit the cases with disordered object orientations in the second row. The third and the fourth rows present the results under various scene types, and the fifth row shows the performance in handling cluttered and `out-of-view' objects. All the results manifest that, with different complexities, our method maintains visually appealing object meshes with reasonable object placement.

\begin{figure*}[!ht]
	\centering
	\begin{subfigure}[t]{0.16\textwidth}
		\includegraphics[width=\textwidth]  
		{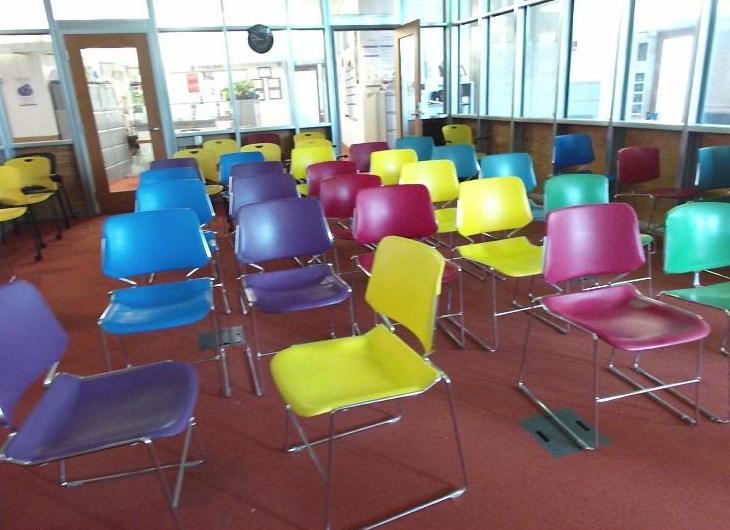}
		\includegraphics[width=\textwidth]
		{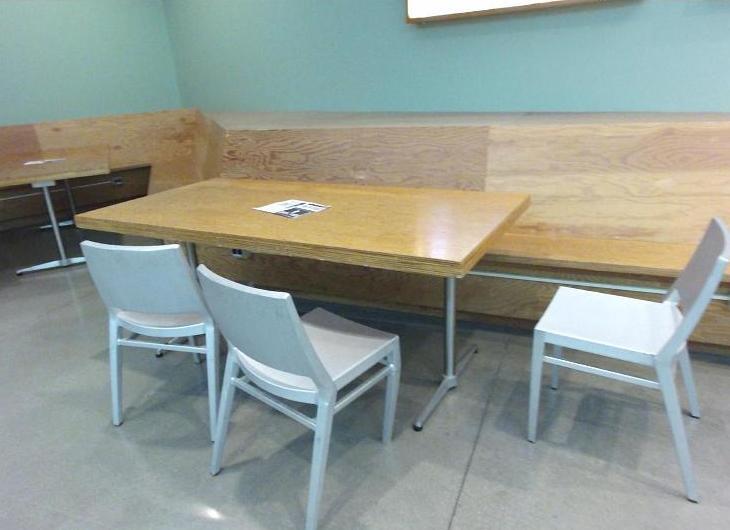}
		\includegraphics[width=\textwidth]
		{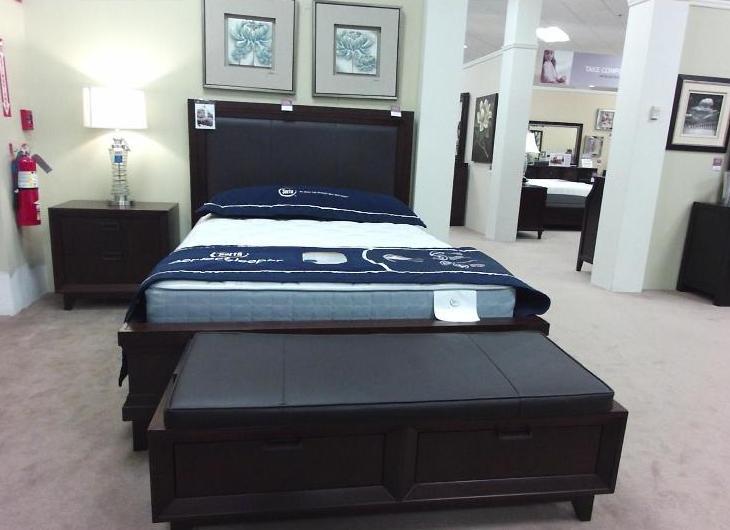}
		\includegraphics[width=\textwidth]
		{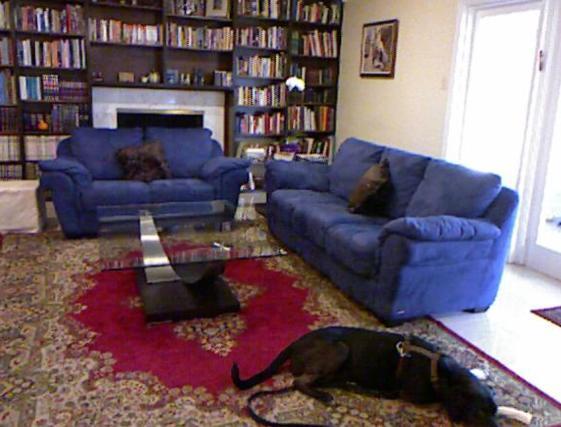}
		\includegraphics[width=\textwidth]
		{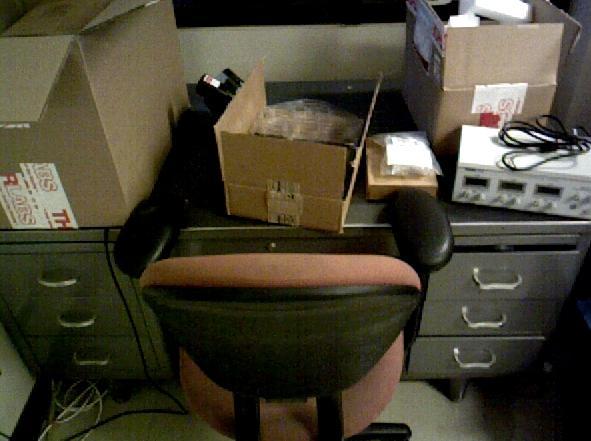}
	\end{subfigure}
	\begin{subfigure}[t]{0.16\textwidth}
		\includegraphics[width=\textwidth]
		{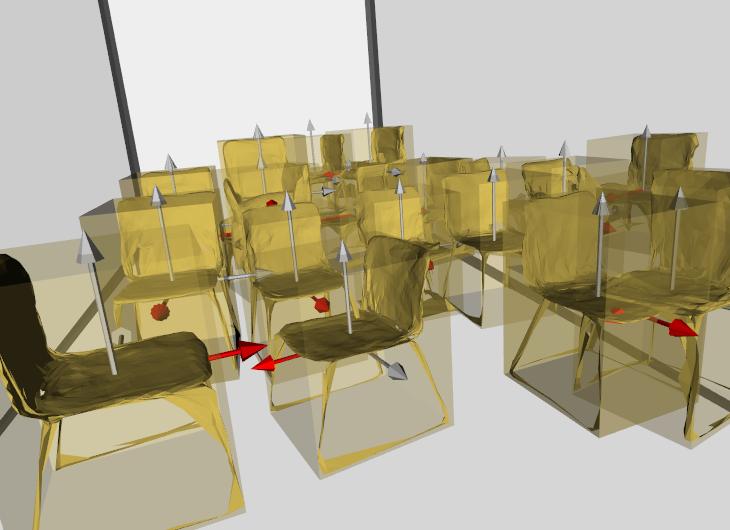}
		\includegraphics[width=\textwidth]
		{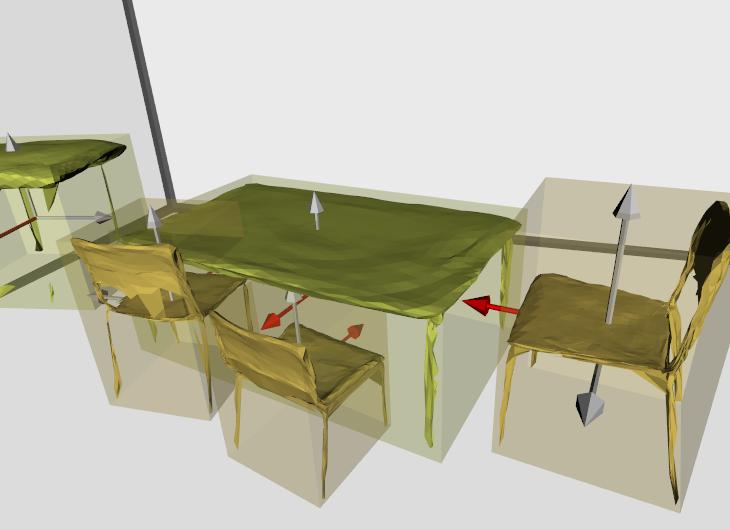}
		\includegraphics[width=\textwidth]
		{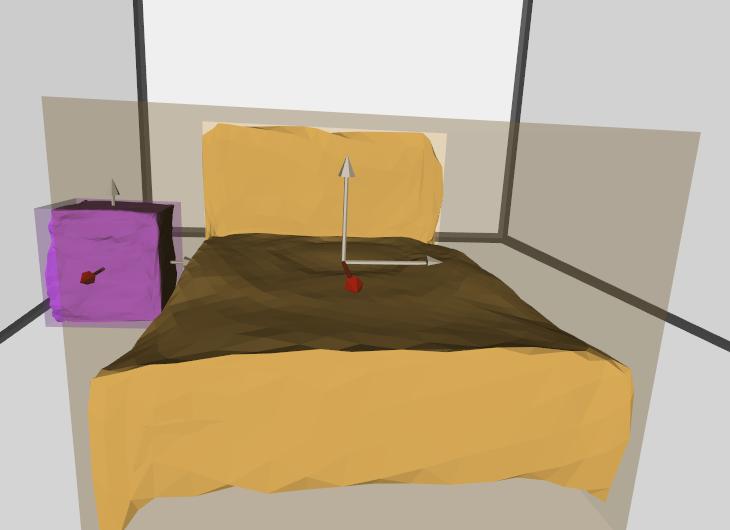}
		\includegraphics[width=\textwidth]
		{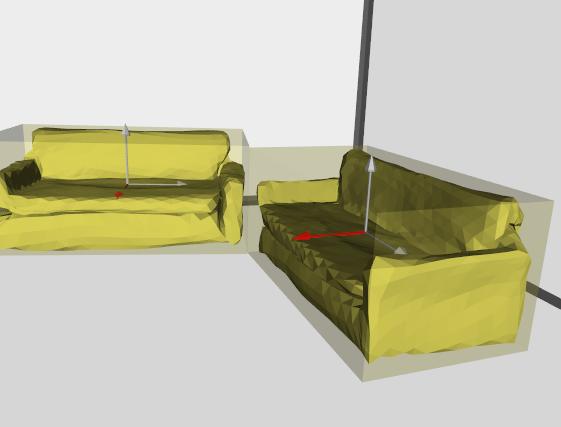}
		\includegraphics[width=\textwidth]
		{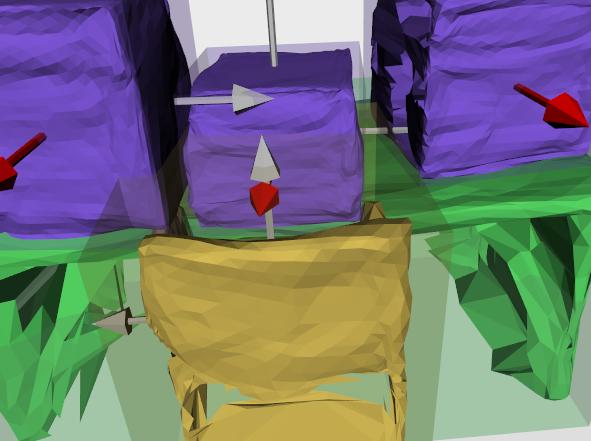}
	\end{subfigure}
	\hfill
	\begin{subfigure}[t]{0.16\textwidth}
		\includegraphics[width=\textwidth]  
		{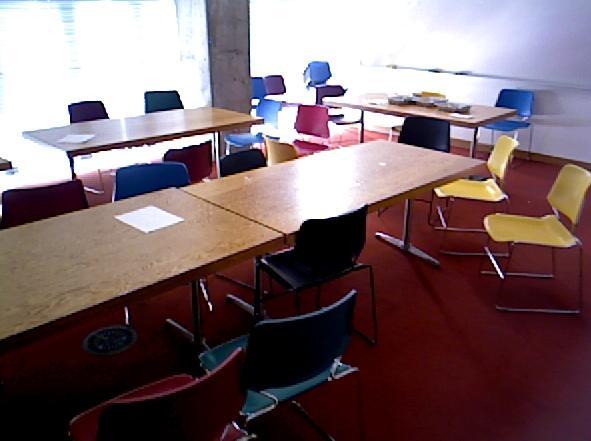}
		\includegraphics[width=\textwidth]
		{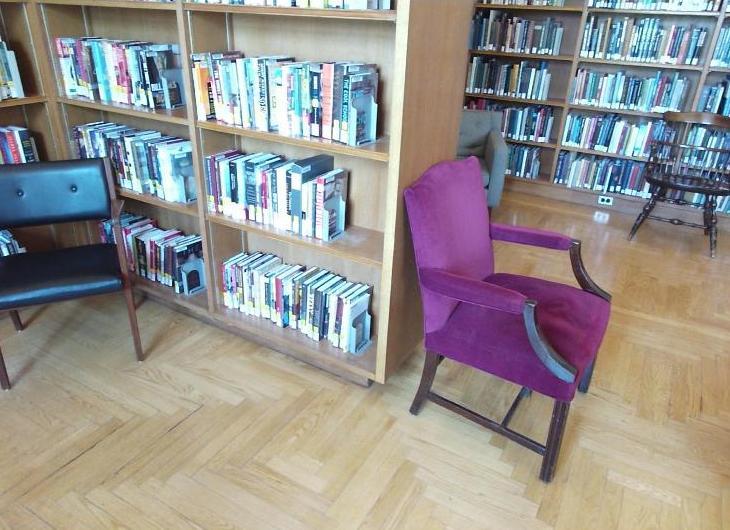}
		\includegraphics[width=\textwidth]
		{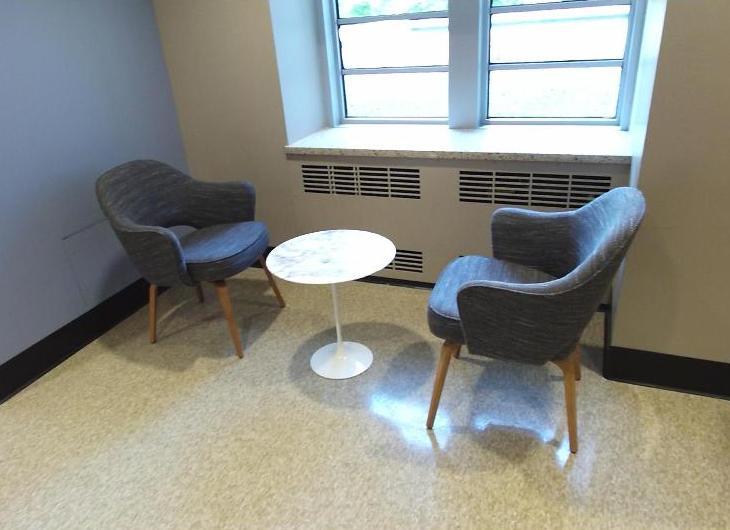}
		\includegraphics[width=\textwidth]
		{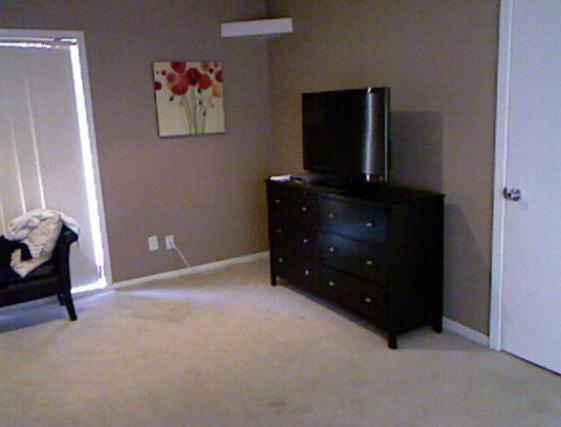}
		\includegraphics[width=\textwidth]
		{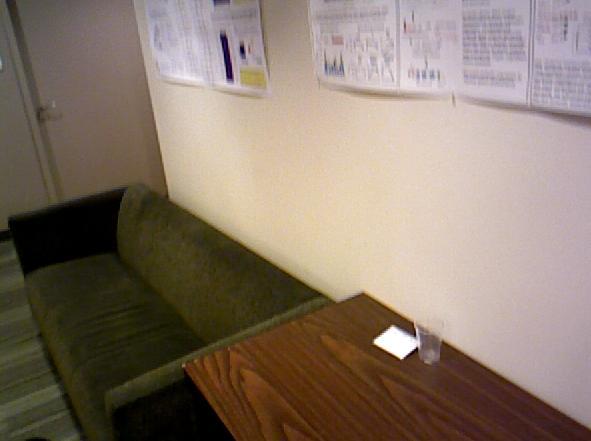}
	\end{subfigure}
	\begin{subfigure}[t]{0.16\textwidth}
		\includegraphics[width=\textwidth]  
		{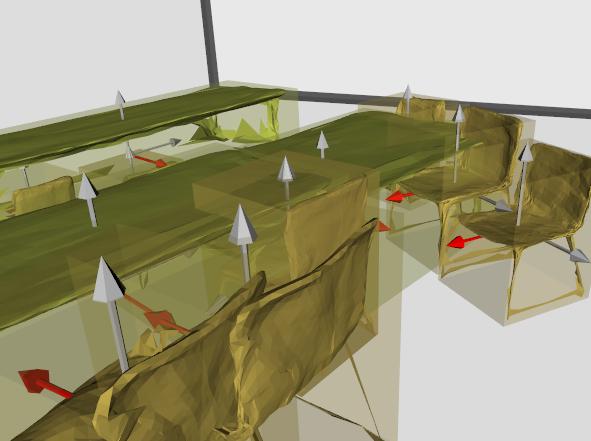}
		\includegraphics[width=\textwidth]
		{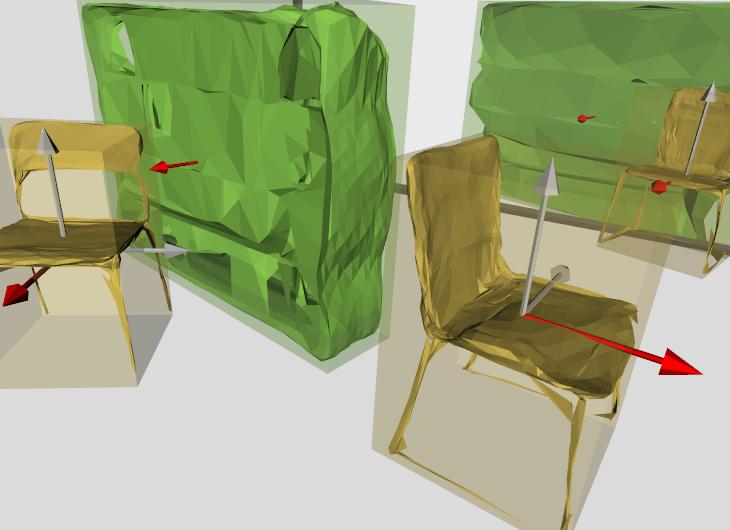}
		\includegraphics[width=\textwidth]
		{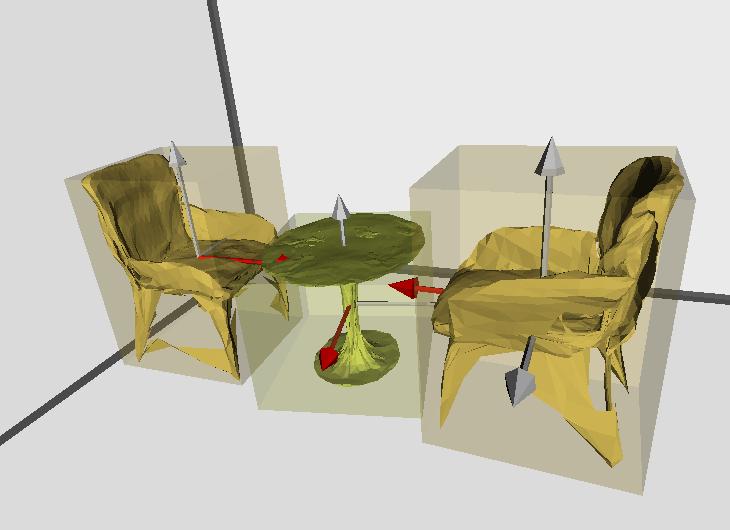}
		\includegraphics[width=\textwidth]
		{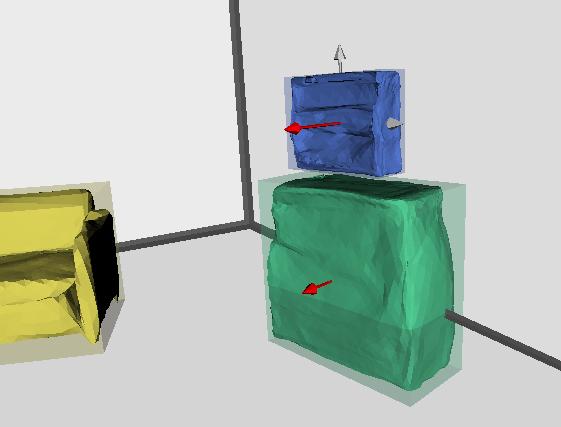}
		\includegraphics[width=\textwidth]
		{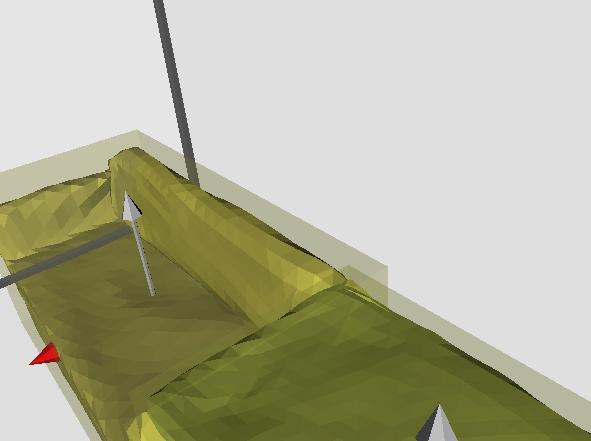}
	\end{subfigure}
	\hfill
	\begin{subfigure}[t]{0.16\textwidth}
		\includegraphics[width=\textwidth]  
		{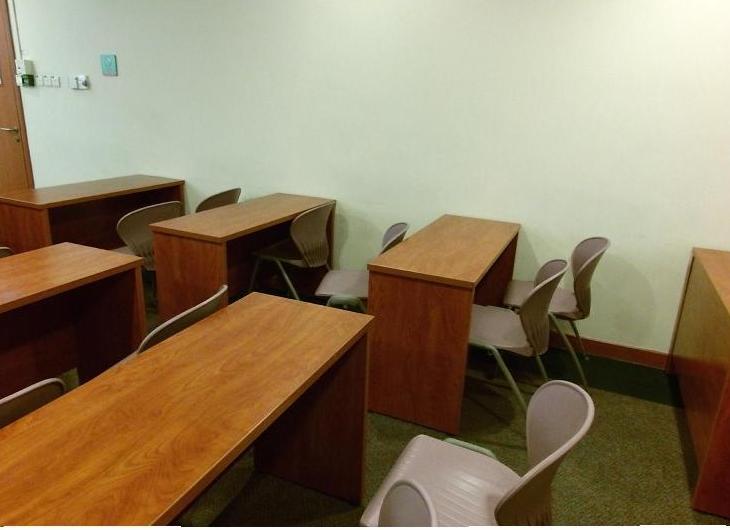}
		\includegraphics[width=\textwidth]
		{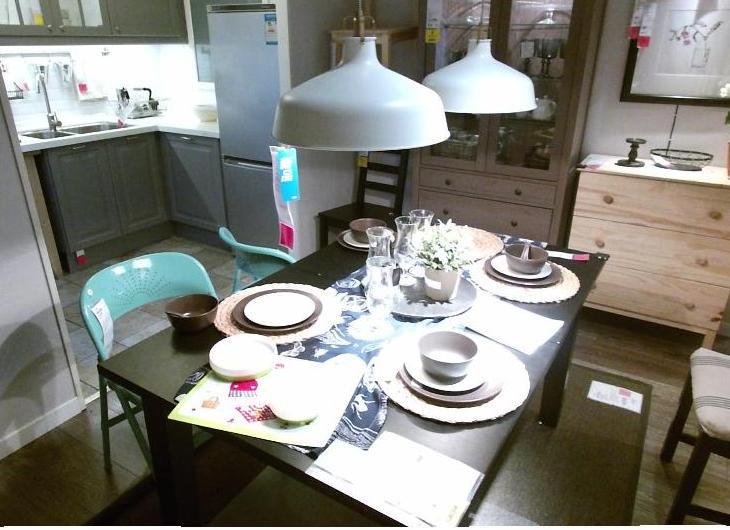}
		\includegraphics[width=\textwidth]
		{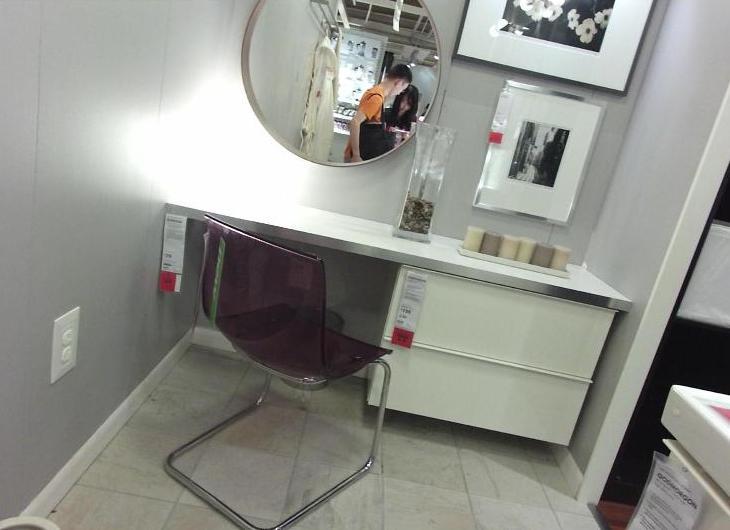}
		\includegraphics[width=\textwidth]
		{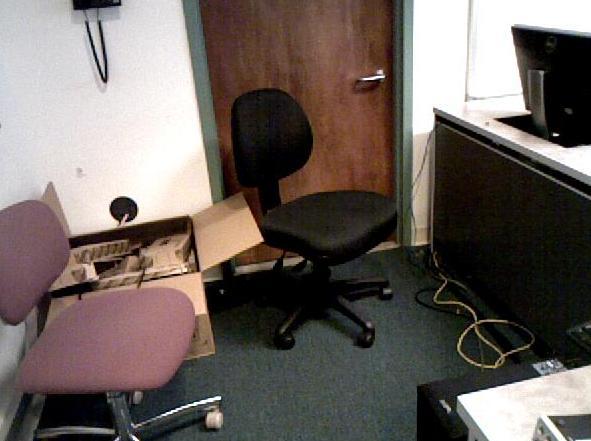}
		\includegraphics[width=\textwidth]
		{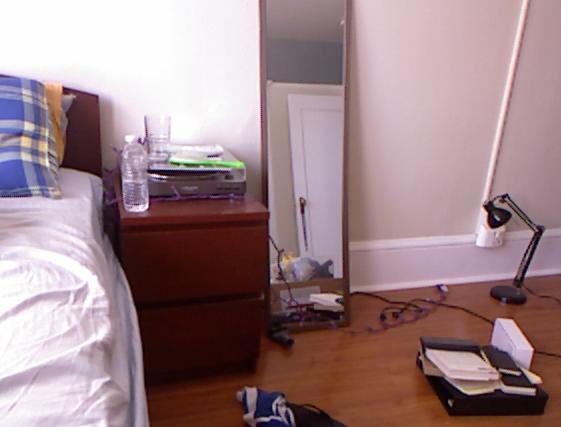}
	\end{subfigure}
	\begin{subfigure}[t]{0.16\textwidth}
		\includegraphics[width=\textwidth]  
		{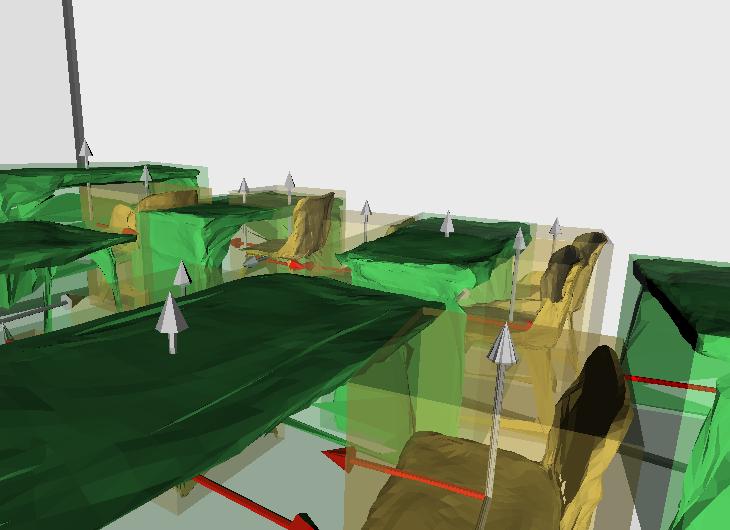}
		\includegraphics[width=\textwidth]
		{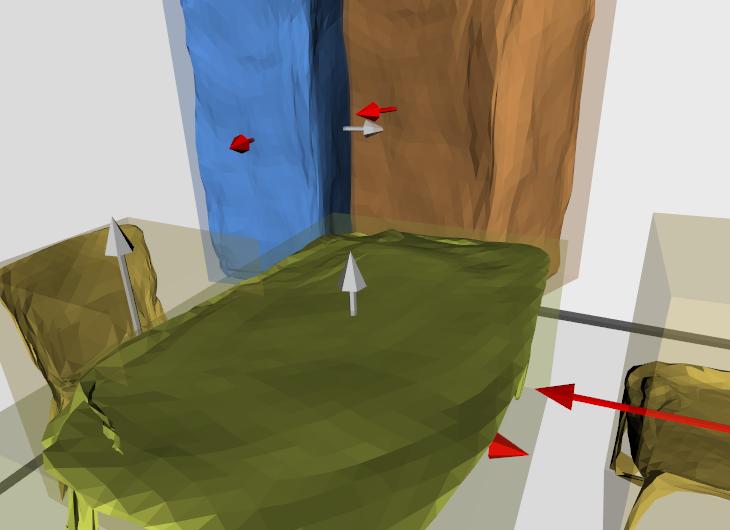}
		\includegraphics[width=\textwidth]
		{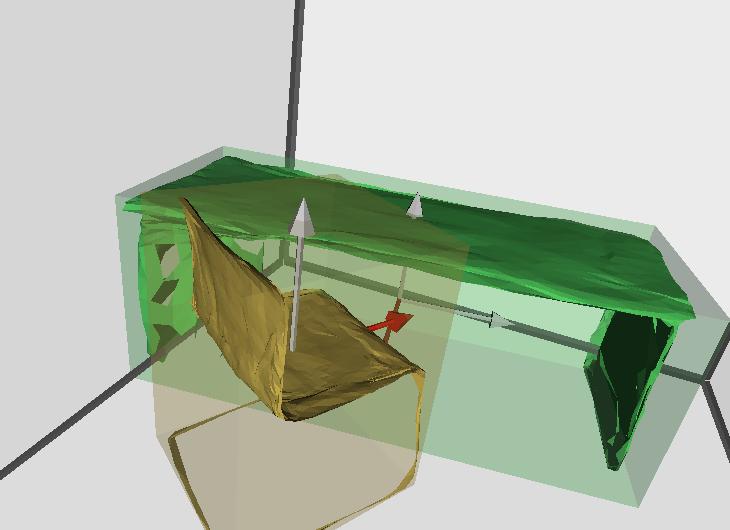}
		\includegraphics[width=\textwidth]
		{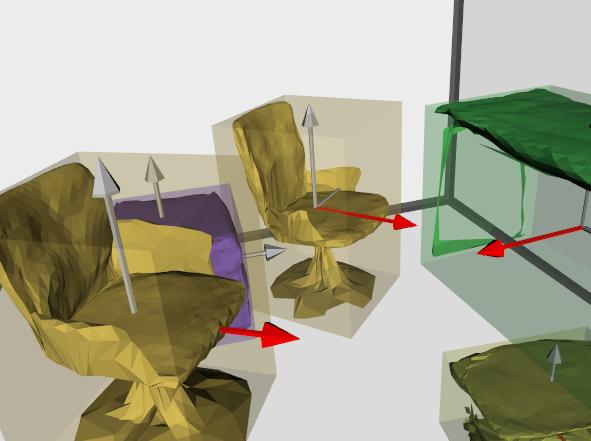}
		\includegraphics[width=\textwidth]
		{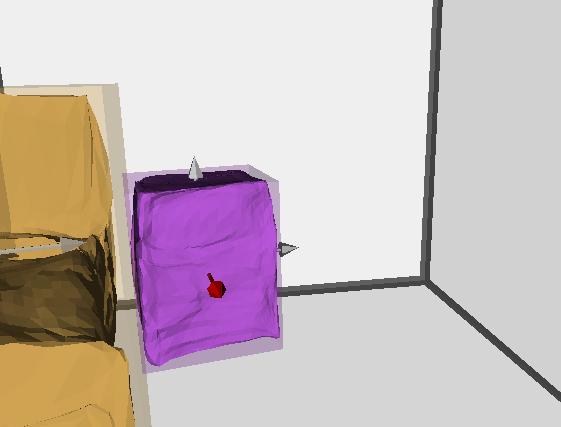}
	\end{subfigure}
	\caption{Scene reconstruction on SUN RGB-D. Given a single image, our method end-to-end reconstructs the room layout, camera pose with object bounding boxes, poses and meshes.}
	\label{fig:scene_recon}
\end{figure*}

\subsection{Quantitative Analysis and Comparison}
We compare the quantitative performance of our method with the state-of-the-arts on four aspects: 1. layout estimation; 2. camera pose prediction; 3. 3D object detection and 4. object and scene mesh reconstruction.  The object mesh reconstruction is tested on Pix3D, and the others are evaluated on SUN RGB-D. We also ablate our method by removing joint training: each subnetwork is trained individually, to investigate the complementary benefits of combining scene understanding and object reconstruction.

\noindent \textbf{Layout Estimation:} We compare our method with existing layout understanding works \cite{choi2013understanding,huang2018holistic,huang2018cooperative}. As shown in Table~\ref{compare:layout}, joint training with room layout, object bounding boxes and meshes helps to improve the layout estimation, providing a gain of 2 points than the state-of-the-arts.

\begin{table}
	\begin{center}
		\begin{tabular}{|l|c|c|c|}
			\hline
			Method & 3D Layout & Cam pitch & Cam roll \\
			\hline\hline
			3DGP \cite{choi2013understanding} & 19.2 & - & - \\
			Hedau \cite{hedau2009recovering} & - & 33.85 & 3.45\\
			HoPR \cite{huang2018holistic} & 54.9 & 7.60 & 3.12 \\
			CooP \cite{huang2018cooperative} & 56.9 & 3.28 & 2.19\\
			Ours (w/o. joint) & 57.6 & 3.68 & 2.59 \\
			Ours (joint) & \textbf{59.2} & \textbf{3.15} & \textbf{2.09} \\
			\hline
		\end{tabular}
	\end{center}
	\caption{Comparisons of 3D layout and camera pose estimation on SUN RGB-D. We report the average IoU to evaluate layout prediction (higher is better), and the mean absolute error of pitch and roll angles (in degree) to test camera pose (lower is better). Note that our camera axes are defined in a different order with \cite{huang2018cooperative} (see the supplementary file).}
	\label{compare:layout}
\end{table}

\begin{table*}[!h]
	\begin{center}
		\resizebox{2.1\columnwidth}{!}{
			\begin{tabular}{|l|c|c|c|c|c|c|c|c|c|c|c|}
				\hline
				Method & bed & chair & sofa & table & desk & dresser & nightstand & sink & cabinet & lamp & mAP\\
				\hline\hline
				3DGP \cite{choi2013understanding} & 5.62 & 2.31 & 3.24 & 1.23 & - & - & - & - & - & - & - \\
				HoPR \cite{huang2018holistic} & 58.29 & 13.56 & 28.37 & 12.12 & 4.79 & 13.71 & 8.80 & 2.18 & 0.48 & 2.41 & 14.47 \\
				CooP \cite{huang2018cooperative}\textsuperscript{*} & \textbf{63.58} & 17.12 & 41.22 & 26.21 & 9.55 & 4.28 & 6.34 & 5.34 & 2.63 & 1.75 & 17.80\\
				CooP \cite{huang2018cooperative}\textsuperscript{**} & 57.71 & 15.21 & 36.67 & 31.16 & 19.90 & 15.98 & 11.36 & 15.95 & 10.47 & 3.28 & 21.77\\
				Ours (w/o. joint) & 59.03 & 15.98 & 43.95 & 35.28 & 23.65 & 19.20 & 6.87 & 14.40 & 11.39 & 3.46 & 23.32 \\
				Ours (joint) & 60.65 & \textbf{17.55} & \textbf{44.90} & \textbf{36.48} & \textbf{27.93} & \textbf{21.19} & \textbf{17.01} & \textbf{18.50} & \textbf{14.51}& \textbf{5.04}& \textbf{26.38}\\
				\hline
		\end{tabular}}
	\end{center}
	\caption{Comparisons of 3D object detection. We compare the average precision of detected objects on SUN RGB-D (higher is better). \cite{huang2018cooperative}\textsuperscript{*} shows the results from their paper, which are trained with fewer object categories. CooP \cite{huang2018cooperative}\textsuperscript{**} presents the model trained on the NYU-37 object labels for a fair comparison.}
	\label{compare:3ddetection}
\end{table*}
\begin{table*}[!h]
	\begin{center}
		\resizebox{2.1\columnwidth}{!}{
			\begin{tabular}{|l|c c c | c c c | c c c|}
				\hline
				& \multicolumn{3}{c|}{Translation (meters)} & \multicolumn{3}{c|}{Rotation (degrees)} & \multicolumn{3}{c|}{Scale} \\
				Method & Median & Mean & (Err$\leq$0.5m)\% & Median & Mean & (Err$\leq$\ang{30})\% & Median & Mean & (Err$\leq$0.2)\% \\
				& \multicolumn{2}{c}{(lower is better)} & (higher is better) & \multicolumn{2}{c}{(lower is better)} & (higher is better) & \multicolumn{2}{c}{(lower is better)} & (higher is better)\\
				\hline\hline
				Tulsiani \textit{et al.}\cite{tulsiani2018factoring} & 0.49 & 0.62 & 51.0 & 14.6 & \textbf{42.6} & 63.8 & 0.37 & 0.40 & 18.9\\
				Ours (w/o. joint) & 0.52 & 0.65 & 49.2 & 15.3 & 45.1 & 64.1 & 0.28 & 0.29 & 42.1 \\
				Ours (joint) & \textbf{0.48} & \textbf{0.61} & \textbf{51.8} & \textbf{14.4} & 43.7 & \textbf{66.5} & \textbf{0.22} & \textbf{0.26} & \textbf{43.7} \\
				\hline
		\end{tabular}}
	\end{center}
	\caption{Comparisons of object pose prediction. The difference values of translation, rotation and scale between the predicted and the ground-truth bounding boxes on NYU v2 are reported, where the median and mean of the differences are listed in the first two columns (lower is better). The third column presents the correct rate within a threshold (higher is better).}
	\label{compare:object_pose}
\end{table*}
\begin{table*}[!h]
	\begin{center}
		\begin{tabular}{|l|c c c c c c c c c|c|}
			\hline
			Category & bed & bookcase & chair & desk & sofa & table & tool & wardrobe & misc & mean\\
			\hline\hline
			AtlasNet \cite{groueix2018} & 9.03 & 6.91 & 8.37 & 8.59 & 6.24 & 19.46 & 6.95 & 4.78 & 40.05 & 12.26\\
			TMN \cite{Junyi} & 7.78 & \textbf{5.93} & 6.86 & 7.08 & 4.25 & 17.42 & 4.13 & 4.09 & \textbf{23.68} & 9.03\\
			Ours (w/o. edge) & 8.19 & 6.81 & 6.26 & 5.97 & 4.12 & 15.09 & 3.93 & 4.01 & 25.19 & 8.84\\
			Ours (w/o. dens) & 8.16 & 6.70 & 6.38 & \textbf{5.12} & 4.07 & 16.16 & 3.63 & 4.32 & 24.22 & 8.75\\
			Ours & \textbf{5.99} & 6.56 & \textbf{5.32} & 5.93 & \textbf{3.36} & \textbf{14.19} & \textbf{3.12} & \textbf{3.83} & 26.93 & \textbf{8.36}\\
			\hline
		\end{tabular}
	\end{center}
	\caption{Comparisons of object reconstruction on Pix3D. The Chamfer distance is used in evaluation. 10K points are sampled from the predicted mesh after being aligned with the ground-truth using ICP. The values are in units of $10^{-3}$ (lower is better).}
	\label{compare:object_recon}
\end{table*}

\noindent \textbf{Camera Pose Estimation:} Camera pose is defined by $\mathbf{R}\left(\beta,\gamma\right)$, hence we evaluate the pitch $\beta$ and roll $\gamma$ with the mean absolute error with the ground-truth. The results are show in Table~\ref{compare:layout}, where we observe that joint learning also benefits the camera pose estimation.

\noindent \textbf{3D Object Detection:} We investigate the object detection with the benchmark consistent with \cite{huang2018cooperative}, where the mean average precision (mAP) is employed using 3D bounding box IoU. A detection is considered true positive if its IoU with the ground-truth is larger than 0.15. We compare our method with existing 3D detection works \cite{choi2013understanding,huang2018holistic,huang2018cooperative} on the shared object categories in Table~\ref{compare:3ddetection}. The full table on all object categories is listed in the supplementary file. The comparisons show that our method significantly improves over the state-of-the-art methods, and consistently advances the ablated version. The reason could be two-fold. One is that the global loss $\mathcal{L}_{g}$ in joint learning involves geometry constraint which ensures the physical rationality, and the other is that multi-lateral relational features in ODN benefit the 3D detection in predicting spatial occupancy.

We also compare our work with \cite{tulsiani2018factoring} to evaluate object pose prediction. We keep consistent with them by training on the NYU v2 dataset \cite{silberman2012indoor} with their six object categories and ground-truth 2D boxes. The results are reported in Table~\ref{compare:object_pose}. Object poses are tested with errors in object translation, rotation and scale. We refer readers to \cite{tulsiani2018factoring} for the definition of the metrics. The results further demonstrate that our method not only obtains reasonable spatial occupancy (mAP), but also retrieves faithful object poses.

\noindent\textbf{Mesh Reconstruction:} We evaluate mesh reconstruction on both the object and scene levels. For object reconstruction, we compare our MGN with the state-of-the-arts \cite{groueix2018,Junyi} in Table~\ref{compare:object_recon}. We ablate our topology modification method with two versions: 1. removing faces instead of edges (w/o. edge); 2. using distance threshold \cite{Junyi} instead of our local density (w/o. dens) for topology modification. The results show that each module improves the mean accuracy, and combining them advances our method to the state-of-the-art. A possible reason is that using local density keeps small-scale topology, and cutting edges is more robust in avoiding incorrect mesh modification than removing faces. Mesh reconstruction of scenes is evaluated with $\mathcal{L}_{g}$ in Equation~\ref{eqn:03}, where the loss is calculated with the average distance from the point cloud of each object to its nearest neighbor on the reconstructed mesh. Different from single object reconstruction, scene meshes are evaluated considering object alignment in the world system. In our test, $\mathcal{L}_{g}$ decreases from 1.89e-2 to 1.43e-2 with our joint learning.

\subsection{Ablation Analysis and Discussion}
To better understand the effect of each design on the final result, we ablate our method with five configurations:\\
\noindent \textbf{$\bm{C_{0}}$:} without relational features (in ODN) and joint training (Baseline).\\
\noindent \textbf{$\bm{C_{1}}$:} Baseline + relational features.\\
\noindent \textbf{$\bm{C_{2}}$:} Baseline + (only) cooperative loss $\mathcal{L}_{co}$ in joint training.\\
\noindent \textbf{$\bm{C_{3}}$:} Baseline + (only) global loss $\mathcal{L}_{g}$ in joint training.\\
\noindent \textbf{$\bm{C_{4}}$:} Baseline + joint training ($\mathcal{L}_{g}$ + $\mathcal{L}_{co}$).\\
\noindent \textbf{Full:} Baseline + relational features + joint training.

We test the layout estimation, 3D detection and scene mesh reconstruction with 3D IoU, mAP and $\mathcal{L}_{g}$. The results are reported in Table~\ref{compare:ablative}, from which we observe that:

\noindent$\bm{C_{0}}$ v.s.$\bm{C_{4}}$ and $\bm{C_{1}}$ v.s. \textbf{Full}: Joint training consistently improves layout estimation, object detection and scene mesh reconstruction no matter using relational features or not.\\
$\bm{C_{0}}$ v.s.$\bm{C_{1}}$ and $\bm{C_{4}}$ v.s. \textbf{Full}: Relational features help to improve 3D object detection, which indirectly reduces the loss in scene mesh reconstruction.\\
$\bm{C_{0}}$ v.s.$\bm{C_{2}}$ and $\bm{C_{0}}$ v.s. $\bm{C_{3}}$: In joint loss, both $\mathcal{L}_{co}$ and $\mathcal{L}_{g}$ in joint training benefit the final outputs, and combing them further advances the accuracy.

We also observe that the global loss $\mathcal{L}_{g}$ shows the most effect on object detection and scene reconstruction, and the cooperative loss $\mathcal{L}_{co}$ provides more benefits than others on layout estimation. Besides, scene mesh loss decreases with the increasing of object detection performance. It is inline with the intuition that object alignment significantly affects mesh reconstruction. Fine-tuning MGN on SUN RGB-D can not improve single object reconstruction on Pix3D. It reflects that object reconstruction depends on clean mesh for supervision. All the facts above explain that the targets for full scene reconstruction actually are intertwined together, which makes joint reconstruction a feasible solution toward total scene understanding.

\begin{table}[!h]
	\begin{center}
		\resizebox{\columnwidth}{!}{
			\begin{tabular}{|l|c|c|c|}
				\hline
				Version & Layout (IoU) & 3D Objects (mAP) & Scene mesh $(\mathcal{L}_{g})$ \\
				& (higher is better) & (higher is better) & (lower is better) \\
				\hline\hline
				$\bm{C_{0}}$ & 57.63 & 20.19 & 2.10 \\
				$\bm{C_{1}}$ & 57.63 & 23.32 & 1.89 \\
				$\bm{C_{2}}$ & 58.21 & 21.77 & 1.73 \\
				$\bm{C_{3}}$ & 57.92 & 24.59 & 1.64 \\
				$\bm{C_{4}}$ & 58.87 & 25.62 & 1.52 \\
				\textbf{Full} & \textbf{59.25} & \textbf{26.38} & \textbf{1.43}\\
				\hline
		\end{tabular}}
	\end{center}
	\caption{Ablation analysis in layout estimation, 3d object detection and scene mesh reconstruction on SUN RGB-D. The $\mathcal{L}_{g}$ values are in units of $10^{-2}$.}
	\label{compare:ablative}
\end{table}
\section{Conclusion}
We develop an end-to-end indoor scene reconstruction approach from a single image. It embeds scene understanding and mesh reconstruction for joint training, and automatically generates the room layout, camera pose, object bounding boxes and meshes to fully recover the room and object geometry. Extensive experiments show that our joint learning approach significantly improves the performance on each subtask and advances the state-of-the-arts. It indicates that each individual scene parsing process has an implicit impact on the others, revealing the necessity of training them integratively toward total 3D reconstruction. One limitation of our method is the requirement for dense point cloud for learning object meshes, which is labor-consuming to obtain in real scenes. To tackle this problem, a self or weakly supervised scene reconstruction method would be a desirable solution in the future work.

\noindent\textbf{Acknowledgment} This work was partially supported by grants No.2018YFB1800800, No.2018B030338001, NSFC-61902334, NSFC-61629101, NSFC-61702433, NSFC-61661146002, No.ZDSYS201707251409055, No. 2017ZT07X152, VISTA AR project (funded by the Interreg France (Channel) England, ERDF), Innovate UK Smart Grants (39012), the Fundamental Research Funds for the Central Universities, the China Scholarship Council and Bournemouth University.

{\small
\bibliographystyle{ieee_fullname}
\bibliography{egpaper_final}

\begin{thebibliography}{10}\itemsep=-1pt

\bibitem{chen2019holistic++}
Yixin Chen, Siyuan Huang, Tao Yuan, Siyuan Qi, Yixin Zhu, and Song-Chun Zhu.
\newblock Holistic++ scene understanding: Single-view 3d holistic scene parsing
  and human pose estimation with human-object interaction and physical
  commonsense.
\newblock {\em arXiv preprint arXiv:1909.01507}, 2019.

\bibitem{chen2019learning}
Zhiqin Chen and Hao Zhang.
\newblock Learning implicit fields for generative shape modeling.
\newblock In {\em Proceedings of the IEEE Conference on Computer Vision and
  Pattern Recognition}, pages 5939--5948, 2019.

\bibitem{choi2013understanding}
Wongun Choi, Yu-Wei Chao, Caroline Pantofaru, and Silvio Savarese.
\newblock Understanding indoor scenes using 3d geometric phrases.
\newblock In {\em Proceedings of the IEEE Conference on Computer Vision and
  Pattern Recognition}, pages 33--40, 2013.

\bibitem{3D-R2N2}
Christopher~Bongsoo Choy, Danfei Xu, JunYoung Gwak, Kevin Chen, and Silvio
  Savarese.
\newblock 3d-r2n2: {A} unified approach for single and multi-view 3d object
  reconstruction.
\newblock In {\em Proceedings of the European Conference on Computer Vision
  (ECCV)}, pages 628--644, 2016.

\bibitem{dasgupta2016delay}
Saumitro Dasgupta, Kuan Fang, Kevin Chen, and Silvio Savarese.
\newblock Delay: Robust spatial layout estimation for cluttered indoor scenes.
\newblock In {\em Proceedings of the IEEE conference on computer vision and
  pattern recognition}, pages 616--624, 2016.

\bibitem{deprelle2019learning}
Theo Deprelle, Thibault Groueix, Matthew Fisher, Vladimir~G Kim, Bryan~C
  Russell, and Mathieu Aubry.
\newblock Learning elementary structures for 3d shape generation and matching.
\newblock {\em arXiv preprint arXiv:1908.04725}, 2019.

\bibitem{du2018learning}
Yilun Du, Zhijian Liu, Hector Basevi, Ales Leonardis, Bill Freeman, Josh
  Tenenbaum, and Jiajun Wu.
\newblock Learning to exploit stability for 3d scene parsing.
\newblock In {\em Advances in Neural Information Processing Systems}, pages
  1726--1736, 2018.

\bibitem{fan2017point}
Haoqiang Fan, Hao Su, and Leonidas~J Guibas.
\newblock A point set generation network for 3d object reconstruction from a
  single image.
\newblock In {\em Proceedings of the IEEE conference on computer vision and
  pattern recognition}, pages 605--613, 2017.

\bibitem{gkioxari2019mesh}
Georgia Gkioxari, Jitendra Malik, and Justin Johnson.
\newblock Mesh r-cnn.
\newblock {\em arXiv preprint arXiv:1906.02739}, 2019.

\bibitem{groueix2018}
Thibault Groueix, Matthew Fisher, Vladimir~G. Kim, Bryan Russell, and Mathieu
  Aubry.
\newblock {AtlasNet: A Papier-M\^ach\'e Approach to Learning 3D Surface
  Generation}.
\newblock In {\em Proceedings of IEEE Conference on Computer Vision and Pattern
  Recognition (CVPR)}, 2018.

\bibitem{he2016deep}
Kaiming He, Xiangyu Zhang, Shaoqing Ren, and Jian Sun.
\newblock Deep residual learning for image recognition.
\newblock In {\em Proceedings of the IEEE conference on computer vision and
  pattern recognition}, pages 770--778, 2016.

\bibitem{hedau2009recovering}
Varsha Hedau, Derek Hoiem, and David Forsyth.
\newblock Recovering the spatial layout of cluttered rooms.
\newblock In {\em 2009 IEEE 12th international conference on computer vision},
  pages 1849--1856. IEEE, 2009.

\bibitem{hu2018relation}
Han Hu, Jiayuan Gu, Zheng Zhang, Jifeng Dai, and Yichen Wei.
\newblock Relation networks for object detection.
\newblock In {\em Proceedings of the IEEE Conference on Computer Vision and
  Pattern Recognition}, pages 3588--3597, 2018.

\bibitem{huang2018cooperative}
Siyuan Huang, Siyuan Qi, Yinxue Xiao, Yixin Zhu, Ying~Nian Wu, and Song-Chun
  Zhu.
\newblock Cooperative holistic scene understanding: Unifying 3d object, layout,
  and camera pose estimation.
\newblock In {\em Advances in Neural Information Processing Systems}, pages
  207--218, 2018.

\bibitem{huang2018holistic}
Siyuan Huang, Siyuan Qi, Yixin Zhu, Yinxue Xiao, Yuanlu Xu, and Song-Chun Zhu.
\newblock Holistic 3d scene parsing and reconstruction from a single rgb image.
\newblock In {\em Proceedings of the European Conference on Computer Vision
  (ECCV)}, pages 187--203, 2018.

\bibitem{hueting2017seethrough}
Moos Hueting, Pradyumna Reddy, Vladimir Kim, Ersin Yumer, Nathan Carr, and
  Niloy Mitra.
\newblock Seethrough: finding chairs in heavily occluded indoor scene images.
\newblock {\em arXiv preprint arXiv:1710.10473}, 2017.

\bibitem{izadinia2017im2cad}
Hamid Izadinia, Qi Shan, and Steven~M Seitz.
\newblock Im2cad.
\newblock In {\em Proceedings of the IEEE Conference on Computer Vision and
  Pattern Recognition}, pages 5134--5143, 2017.

\bibitem{kato2018neural}
Hiroharu Kato, Yoshitaka Ushiku, and Tatsuya Harada.
\newblock Neural 3d mesh renderer.
\newblock In {\em Proceedings of the IEEE Conference on Computer Vision and
  Pattern Recognition}, pages 3907--3916, 2018.

\bibitem{kulkarni20193d}
Nilesh Kulkarni, Ishan Misra, Shubham Tulsiani, and Abhinav Gupta.
\newblock 3d-relnet: Joint object and relational network for 3d prediction.
\newblock {\em International Conference on Computer Vision (ICCV)}.

\bibitem{kurenkov2018deformnet}
Andrey Kurenkov, Jingwei Ji, Animesh Garg, Viraj Mehta, JunYoung Gwak,
  Christopher Choy, and Silvio Savarese.
\newblock Deformnet: Free-form deformation network for 3d shape reconstruction
  from a single image.
\newblock In {\em 2018 IEEE Winter Conference on Applications of Computer
  Vision (WACV)}, pages 858--866. IEEE, 2018.

\bibitem{lee2009geometric}
David~C Lee, Martial Hebert, and Takeo Kanade.
\newblock Geometric reasoning for single image structure recovery.
\newblock In {\em 2009 IEEE Conference on Computer Vision and Pattern
  Recognition}, pages 2136--2143. IEEE, 2009.

\bibitem{li2019silhouette}
Lin Li, Salman Khan, and Nick Barnes.
\newblock Silhouette-assisted 3d object instance reconstruction from a
  cluttered scene.
\newblock In {\em Proceedings of the IEEE International Conference on Computer
  Vision Workshops}, pages 0--0, 2019.

\bibitem{LiaoDG18}
Yiyi Liao, Simon Donne, and Andreas Geiger.
\newblock Deep marching cubes: Learning explicit surface representations.
\newblock In {\em Proceedings of the IEEE Conference on Computer Vision and
  Pattern Recognition}, pages 2916--2925, 2018.

\bibitem{lin2014microsoft}
Tsung-Yi Lin, Michael Maire, Serge Belongie, James Hays, Pietro Perona, Deva
  Ramanan, Piotr Doll{\'a}r, and C~Lawrence Zitnick.
\newblock Microsoft coco: Common objects in context.
\newblock In {\em European conference on computer vision}, pages 740--755.
  Springer, 2014.

\bibitem{mallya2015learning}
Arun Mallya and Svetlana Lazebnik.
\newblock Learning informative edge maps for indoor scene layout prediction.
\newblock In {\em Proceedings of the IEEE international conference on computer
  vision}, pages 936--944, 2015.

\bibitem{mandikal20183d}
Priyanka Mandikal, Navaneet KL, and R Venkatesh~Babu.
\newblock 3d-psrnet: Part segmented 3d point cloud reconstruction from a single
  image.
\newblock In {\em Proceedings of the European Conference on Computer Vision
  (ECCV)}, pages 0--0, 2018.

\bibitem{mescheder2019occupancy}
Lars Mescheder, Michael Oechsle, Michael Niemeyer, Sebastian Nowozin, and
  Andreas Geiger.
\newblock Occupancy networks: Learning 3d reconstruction in function space.
\newblock In {\em Proceedings of the IEEE Conference on Computer Vision and
  Pattern Recognition}, pages 4460--4470, 2019.

\bibitem{michalkiewicz2019deep}
Mateusz Michalkiewicz, Jhony~K Pontes, Dominic Jack, Mahsa Baktashmotlagh, and
  Anders Eriksson.
\newblock Deep level sets: Implicit surface representations for 3d shape
  inference.
\newblock {\em arXiv preprint arXiv:1901.06802}, 2019.

\bibitem{navaneet2019capnet}
KL Navaneet, Priyanka Mandikal, Mayank Agarwal, and R~Venkatesh Babu.
\newblock Capnet: Continuous approximation projection for 3d point cloud
  reconstruction using 2d supervision.
\newblock In {\em Proceedings of the AAAI Conference on Artificial
  Intelligence}, volume~33, pages 8819--8826, 2019.

\bibitem{Junyi}
Junyi Pan, Xiaoguang Han, Weikai Chen, Jiapeng Tang, and Kui Jia.
\newblock Deep mesh reconstruction from single rgb images via topology
  modification networks.
\newblock In {\em Proceedings of the IEEE International Conference on Computer
  Vision}, pages 9964--9973, 2019.

\bibitem{park2019deepsdf}
Jeong~Joon Park, Peter Florence, Julian Straub, Richard Newcombe, and Steven
  Lovegrove.
\newblock Deepsdf: Learning continuous signed distance functions for shape
  representation.
\newblock {\em arXiv preprint arXiv:1901.05103}, 2019.

\bibitem{paschalidou2019superquadrics}
Despoina Paschalidou, Ali~Osman Ulusoy, and Andreas Geiger.
\newblock Superquadrics revisited: Learning 3d shape parsing beyond cuboids.
\newblock In {\em Proceedings of the IEEE Conference on Computer Vision and
  Pattern Recognition}, pages 10344--10353, 2019.

\bibitem{qi2018frustum}
Charles~R Qi, Wei Liu, Chenxia Wu, Hao Su, and Leonidas~J Guibas.
\newblock Frustum pointnets for 3d object detection from rgb-d data.
\newblock In {\em Proceedings of the IEEE Conference on Computer Vision and
  Pattern Recognition}, pages 918--927, 2018.

\bibitem{ren2015faster}
Shaoqing Ren, Kaiming He, Ross Girshick, and Jian Sun.
\newblock Faster r-cnn: Towards real-time object detection with region proposal
  networks.
\newblock In {\em Advances in neural information processing systems}, pages
  91--99, 2015.

\bibitem{ren2016coarse}
Yuzhuo Ren, Shangwen Li, Chen Chen, and C-C~Jay Kuo.
\newblock A coarse-to-fine indoor layout estimation (cfile) method.
\newblock In {\em Asian Conference on Computer Vision}, pages 36--51. Springer,
  2016.

\bibitem{riegler2017octnet}
Gernot Riegler, Ali Osman~Ulusoy, and Andreas Geiger.
\newblock Octnet: Learning deep 3d representations at high resolutions.
\newblock In {\em Proceedings of the IEEE Conference on Computer Vision and
  Pattern Recognition}, pages 3577--3586, 2017.

\bibitem{roberts1963machine}
Lawrence~G Roberts.
\newblock {\em Machine perception of three-dimensional solids}.
\newblock PhD thesis, Massachusetts Institute of Technology, 1963.

\bibitem{schwing2013box}
Alexander~G Schwing, Sanja Fidler, Marc Pollefeys, and Raquel Urtasun.
\newblock Box in the box: Joint 3d layout and object reasoning from single
  images.
\newblock In {\em Proceedings of the IEEE International Conference on Computer
  Vision}, pages 353--360, 2013.

\bibitem{shin20193d}
Daeyun Shin, Zhile Ren, Erik~B Sudderth, and Charless~C Fowlkes.
\newblock 3d scene reconstruction with multi-layer depth and epipolar
  transformers.
\newblock In {\em Proceedings of the IEEE International Conference on Computer
  Vision (ICCV)}, 2019.

\bibitem{silberman2012indoor}
Nathan Silberman, Derek Hoiem, Pushmeet Kohli, and Rob Fergus.
\newblock Indoor segmentation and support inference from rgbd images.
\newblock In {\em European Conference on Computer Vision}, pages 746--760.
  Springer, 2012.

\bibitem{song2015sun}
Shuran Song, Samuel~P Lichtenberg, and Jianxiong Xiao.
\newblock Sun rgb-d: A rgb-d scene understanding benchmark suite.
\newblock In {\em Proceedings of the IEEE conference on computer vision and
  pattern recognition}, pages 567--576, 2015.

\bibitem{sun2018pix3d}
Xingyuan Sun, Jiajun Wu, Xiuming Zhang, Zhoutong Zhang, Chengkai Zhang, Tianfan
  Xue, Joshua~B Tenenbaum, and William~T Freeman.
\newblock Pix3d: Dataset and methods for single-image 3d shape modeling.
\newblock In {\em Proceedings of the IEEE Conference on Computer Vision and
  Pattern Recognition}, pages 2974--2983, 2018.

\bibitem{tang2019skeleton}
Jiapeng Tang, Xiaoguang Han, Junyi Pan, Kui Jia, and Xin Tong.
\newblock A skeleton-bridged deep learning approach for generating meshes of
  complex topologies from single rgb images.
\newblock In {\em Proceedings of the IEEE Conference on Computer Vision and
  Pattern Recognition}, pages 4541--4550, 2019.

\bibitem{tatarchenko2017octree}
Maxim Tatarchenko, Alexey Dosovitskiy, and Thomas Brox.
\newblock Octree generating networks: Efficient convolutional architectures for
  high-resolution 3d outputs.
\newblock In {\em Proceedings of the IEEE International Conference on Computer
  Vision}, pages 2088--2096, 2017.

\bibitem{tian2019learning}
Yonglong Tian, Andrew Luo, Xingyuan Sun, Kevin Ellis, William~T Freeman,
  Joshua~B Tenenbaum, and Jiajun Wu.
\newblock Learning to infer and execute 3d shape programs.
\newblock {\em arXiv preprint arXiv:1901.02875}, 2019.

\bibitem{tulsiani2018factoring}
Shubham Tulsiani, Saurabh Gupta, David~F Fouhey, Alexei~A Efros, and Jitendra
  Malik.
\newblock Factoring shape, pose, and layout from the 2d image of a 3d scene.
\newblock In {\em Proceedings of the IEEE Conference on Computer Vision and
  Pattern Recognition}, pages 302--310, 2018.

\bibitem{tulsiani2017learning}
Shubham Tulsiani, Hao Su, Leonidas~J Guibas, Alexei~A Efros, and Jitendra
  Malik.
\newblock Learning shape abstractions by assembling volumetric primitives.
\newblock In {\em Proceedings of the IEEE Conference on Computer Vision and
  Pattern Recognition}, pages 2635--2643, 2017.

\bibitem{vaswani2017attention}
Ashish Vaswani, Noam Shazeer, Niki Parmar, Jakob Uszkoreit, Llion Jones,
  Aidan~N Gomez, {\L}ukasz Kaiser, and Illia Polosukhin.
\newblock Attention is all you need.
\newblock In {\em Advances in neural information processing systems}, pages
  5998--6008, 2017.

\bibitem{Wallace}
Bram Wallace and Bharath Hariharan.
\newblock Few-shot generalization for single-image 3d reconstruction via
  priors.
\newblock In {\em Proceedings of the IEEE International Conference on Computer
  Vision}, pages 3818--3827, 2019.

\bibitem{wang2018pixel2mesh}
Nanyang Wang, Yinda Zhang, Zhuwen Li, Yanwei Fu, Wei Liu, and Yu-Gang Jiang.
\newblock Pixel2mesh: Generating 3d mesh models from single rgb images.
\newblock In {\em Proceedings of the European Conference on Computer Vision
  (ECCV)}, pages 52--67, 2018.

\bibitem{wang2018adaptive}
Peng-Shuai Wang, Chun-Yu Sun, Yang Liu, and Xin Tong.
\newblock Adaptive o-cnn: a patch-based deep representation of 3d shapes.
\newblock In {\em SIGGRAPH Asia 2018 Technical Papers}, page 217. ACM, 2018.

\bibitem{xu2019disn}
Qiangeng Xu, Weiyue Wang, Duygu Ceylan, Radomir Mech, and Ulrich Neumann.
\newblock Disn: Deep implicit surface network for high-quality single-view 3d
  reconstruction.
\newblock {\em arXiv preprint arXiv:1905.10711}, 2019.

\end{thebibliography}
}

\appendix
\noindent{\textbf{\large{The supplementary material contains:}}}
\begin{itemize}
	\item Camera and world system configuration.
	\item Network architecture, parameter setting and training strategies.
	\item 3D detection results on SUN RGB-D.
	\item Object class mapping from NYU-37 to Pix3D.
	\item More qualitative comparisons on Pix3D.
	\item More reconstruction samples on SUN RGB-D.
\end{itemize}
\section{Camera and World System Setting}
\begin{figure}[!ht]
	\centering
	\includegraphics[width=0.5\linewidth]{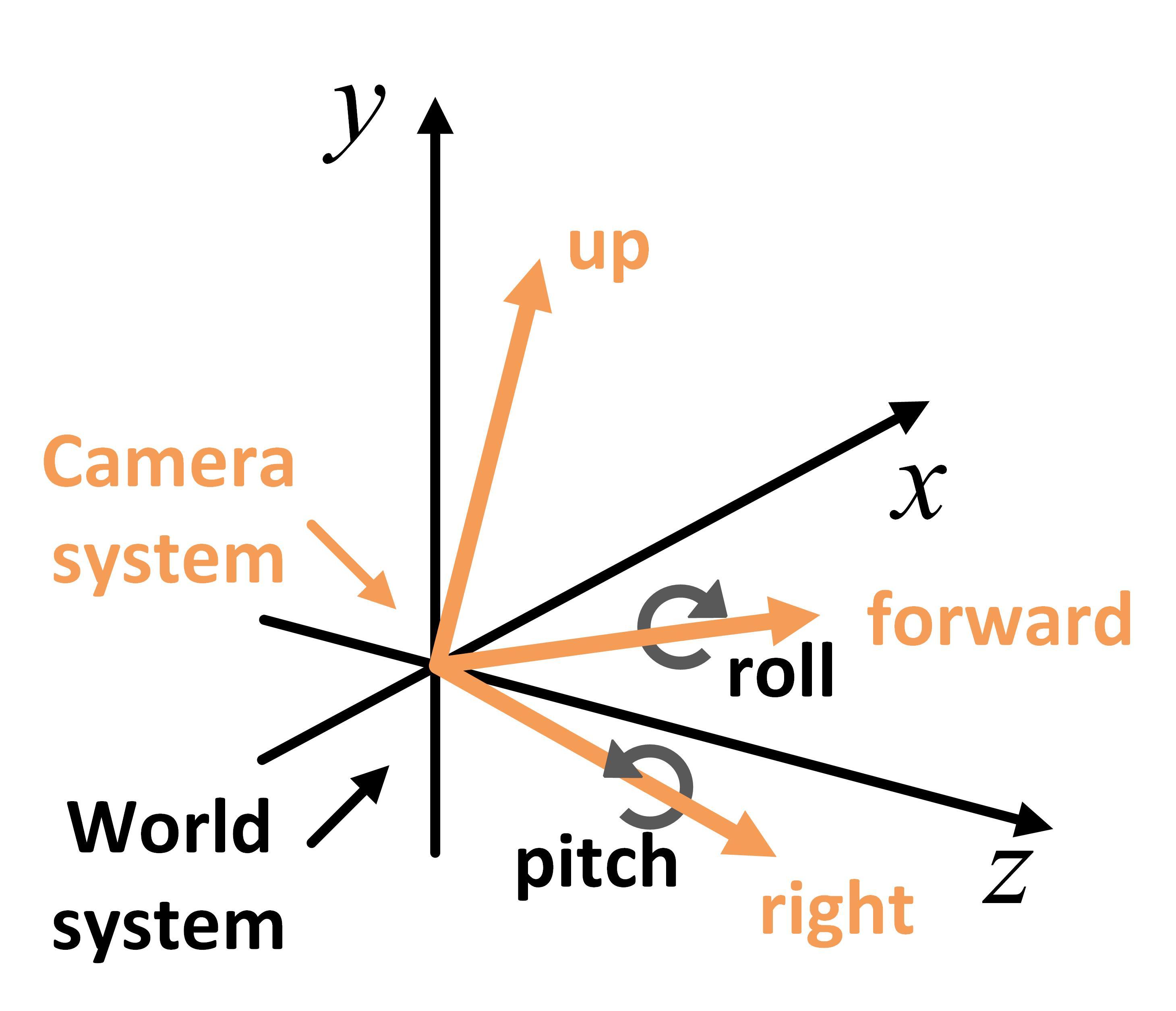}
	\caption{Camera and world systems}
	\label{fig:camera}
\end{figure}

We build the world and the camera systems in this paper as Figure~\ref{fig:camera} shows. The two systems share the same center. The y-axis indicates the vertical direction perpendicular to the floor. We rotate the world system around its y-axis to align the x-axis toward the forward direction of the camera, such that the camera's yaw angle can be removed. Then the camera pose relative to the world system can be expressed by the angles of pitch $\beta$ and roll $\gamma$:

\begin{equation}
	\label{eqn:01}
	\begin{aligned}
		\mathbf{R}\left(\beta,\gamma\right) &= 
		\begin{bmatrix}
			\cos\left(\beta\right) & -\cos\left(\gamma\right)\sin\left(\beta\right) & \sin\left(\beta\right)\sin\left(\gamma\right) \\
			\sin\left(\beta\right) & \cos\left(\beta\right)\cos\left(\gamma\right) & -\cos\left(\beta\right)\sin\left(\gamma\right) \\
			0 & \sin\left(\gamma\right) & \cos\left(\gamma\right)
		\end{bmatrix}.
	\end{aligned}
	\nonumber
\end{equation}

\section{Network Architecture}
\noindent\textbf{Architecture.} We present the architecture of our Object Detection Network (ODN), Layout Estimation Network (LEN) and Mesh Generation Network (MGN) in Table~\ref{ODN}-\ref{MGN}. 

\noindent\textbf{Training strategy.} Our training involves two phases. We first train the three networks individually. ODN and LEN are trained on SUN RGB-D \cite{song2015sun}, while MGN is trained on Pix3D \cite{sun2018pix3d} with their specific loss ( $\sum\lambda_{x}\mathcal{L}_{x}$, $\sum\lambda_{y}\mathcal{L}_{y}$ and $\sum\lambda_{z} \mathcal{L}_{z}$ respectively) (see Line 455, Page 5). All of them are with the batch size of 32 and learning rate at 1e-3 (scaled by 0.5 for every 20 epochs, 100 epochs in total). The MGN is trained with a progressive manner following \cite{Junyi}.  Afterwards, we fine-tune them with the joint losses $\lambda _{co}\mathcal{L}_{co}$ and $\lambda _{g}\mathcal{L}_{g}$ (see Equation 4) together on SUN RGB-D. Specifically, in the joint training, we randomly blend a few Pix3D samples into each batch of SUN RGB-D data to supervise the mesh generation network (i.e. to optimize the mesh loss $\sum\lambda_{z}\mathcal{L}_{z}$). We do so to regularize the mesh generation network because not like Pix3D, SUN RGB-D provides only a partial point-cloud scan of objects, which is not sufficient to supervise full mesh generation. For joint training, we input the full network with a hierarchical batch, where the scene image (from SUN RGB-D) is inputted to LEN, and the object images (from SUN RGB-D and Pix3D) are fed into ODN and MGN for object detection and mesh prediction. We set the hierarchical batch size at 1, and the learning rate at 1e-4 (scaled by 0.5 for every 5 epochs, 20 epochs in total). All the training tasks are implemented on 6x Nvidia 2080Ti GPUs. During testing, our network requires 1.2 seconds on average to predict a scene mesh on a single GPU. 

\noindent\textbf{Parameters.} We set the threshold in our MGN at 0.2. Edges with the classification score below it are removed. In joint training (Section~3.3), we let $\lambda_{r}=10$, $\lambda_{x}=1, \forall x \in \{  \bm{\delta}, d, \bm{s}, \theta  \}$, $\lambda_{y}=1, \forall y \in \{ \beta,\gamma, \bm{C}, \bm{s}^{l}, \theta^{l} \}$, $\lambda_{c}=100$, $\lambda_{e}=10$, $\lambda_{b}=50$, $\lambda_{ce}=0.01$, $\lambda_{co}=10$, $\lambda_{g}=100$.

\begin{table}[!ht]
	\begin{center}
		\resizebox{\columnwidth}{!}{
			\begin{tabular}{|l|c|c|c|}
				\hline
				Index & Inputs & Operation & Output shape \\
				\hline\hline
				(1) & Input & Object images in a scene & Nx3x256x256\\
				(2) & Input & Geometry features \cite{hu2018relation,vaswani2017attention} & N x N x 64\\
				(3) & (1) & ResNet-34 \cite{he2016deep} & Nx2048 \\
				(4) & (2), (3) & Relation Module \cite{hu2018relation} & Nx2048\\
				(5) & (3), (4) & Element-wise sum & Nx2048 \\
				(6) & (5) & FC(128-d)+ReLU+Dropout+FC & $\bm{\delta}$\\
				(7) & (5) & FC(128-d)+ReLU+Dropout+FC & $d$ \\
				(8) & (5) & FC(128-d)+ReLU+Dropout+FC & $\theta$ \\
				(9) & (5) & FC(128-d)+ReLU+Dropout+FC & $\bm{s}$ \\
				\hline
		\end{tabular}}
	\end{center}
	\caption{Architecture of Object Detection Network. It takes all object detections in a scene as input and outputs their projection offset $\bm{\delta}$, distance $d$, orientation $\theta$ and size $\bm{s}$. N is the number of objects in a scene.}
	\label{ODN}
\end{table}

\begin{table}[!ht]
	\begin{center}
		\resizebox{\columnwidth}{!}{
			\begin{tabular}{|l|c|c|c|}
				\hline
				Index & Inputs & Operation & Output shape \\
				\hline\hline
				(1) & Input & Scene image & 3x256x256\\
				(2) & (1) & ResNet-34 \cite{he2016deep}& 2048 \\
				(3) & (2) & FC(1024-d)+ReLU+Dropout+FC & $\beta$\\
				(4) & (2) & FC(1024-d)+ReLU+Dropout+FC & $\gamma$ \\
				(5) & (2) & FC+ReLU+Dropout & 2048 \\
				(6) & (5) & FC(1024-d)+ReLU+Dropout+FC & $\bm{C}$ \\
				(7) & (5) & FC(1024-d)+ReLU+Dropout+FC & $\bm{s}^{l}$ \\
				(8) & (5) & FC(1024-d)+ReLU+Dropout+FC & $\theta^{l}$ \\
				\hline
		\end{tabular}}
	\end{center}
	\caption{Architecture of Layout Estimation Network. LEN takes the full scene image as input and produces the camera pitch $\beta$ and roll $\gamma$ angles, the 3D layout center $\bm{C}$, size $\bm{s}$ and orientation $\theta$ in the world system. }
	\label{LEN}
\end{table}

\begin{table}[!ht]
	\begin{center}
		\resizebox{\columnwidth}{!}{
			\begin{tabular}{|l|c|c|c|}
				\hline
				Index & Inputs & Operation & Output shape\\
				\hline\hline
				(1) & Input & Object image & 3x256x256\\
				(2) & Input & Object class code & $d_{c}$ \\
				(3) & Input & Template Sphere & 3x2562\\
				(4) & (1) & ResNet-18 \cite{he2016deep} & 1024 \\
				(5) & (2),(4) & Concatenate & 1024+$d_{c}$ \\
				(6) & (5) & Repeat & (1024+$d_{c}$)x2562 \\
				(7) & (3),(6) & Concatenate & (1024+$d_{c}$+3)x2562 \\
				(8) & (7) & AtlasNet decoder \cite{groueix2018} & 3x2562 \\
				(9) & (3),(8) & Element-wise sum & 3x2562 \\
				(10) & (9) & Sample points & 3xN\textsubscript{e} \\
				(11) & (5) & Repeat & (1024+$d_{c}$)xN\textsubscript{e} \\
				(12) & (10),(11) & Concatenate & (1024+$d_{c}$+3)xN\textsubscript{e} \\
				(13) & (12) & Edge classifier & 1xN\textsubscript{e} \\
				(14) & (13) & Threshold & 1xN\textsubscript{e} (Mesh topology) \\
				(15) & (6),(9) & Concatenate & (1024+$d_{c}$+3)x2562 \\
				(16) & (15) & AtlasNet decoder \cite{groueix2018} & 3x2562 \\
				(17) & (9),(16) & Element-wise sum & 3x2562 (Mesh points) \\
				\hline
		\end{tabular}}
	\end{center}
	\caption{Architecture of Mesh Generation Network. Note that $d_{c}$ denotes the number of object categories, and N\textsubscript{e} represents the number of points sampled on edges. The edge classifier has the same architecture with AtlasNet decoder, where the last layer is replaced with a fully connected layer for classification.}
	\label{MGN}
\end{table}

\section{3D Detection on SUN RGB-D}
We report the full results of 3D object detection on SUN RGB-D in Table~\ref{compare:more3ddetection}.

\begin{table}[!ht]
	\begin{center}
		\resizebox{\columnwidth}{!}{
			\begin{tabular}{|c|c|c|c|c|c|c|}
				\hline
				cabinet & bed & chair & sofa & table & door & window\\
				\hline\hline
				8 & 1 & 3 & 5 & 6 & 8 & 9 \\
				\hline
				\hline
				bookshelf & picture & counter & blinds & desk & shelves & curtain\\
				\hline\hline
				2 & 9 & 9 & 9 & 4 & 2 & 9 \\
				\hline
				\hline
				dresser & pillow & mirror & floor mat & clothes & books & fridge\\
				\hline\hline
				8 & 9 & 9 & 9 & 9 & 9 & 8\\
				\hline
				\hline
				tv & paper & towel & shower curtain & box & whiteboard & person\\
				\hline\hline
				8 & 9 & 9 & 9 & 8 & 8 & 9 \\
				\hline\hline
				nightstand & toilet & sink & lamp & bathtub & bag & wall\\
				\hline\hline
				8 & 9 & 9 & 9 & 9 & 8 & -\\
				\hline\hline
				floor & ceiling & - & - & - & -  & - \\
				\hline\hline
				- & - & - & - & - & - & - \\
				\hline
		\end{tabular}}
	\end{center}
	\caption{Object class mapping from NYU-37 to Pix3D }
	\label{label_map}
\end{table}

\begin{table*}[!h]
	\begin{center}
		\resizebox{2.1\columnwidth}{!}{
			\begin{tabular}{|l|c|c|c|c|c|c|c|c|c|c|}
				\hline
				Method & cabinet & bed & chair & sofa & table & door & window & bookshelf & picture & counter\\
				\hline\hline
				CooP \cite{huang2018cooperative}\textsuperscript{**} & 10.47 & 57.71 & 15.21 & 36.67 & 31.16 & 0.14 & 0.00 & 3.81 & 0.00 & 27.67\\
				Ours (w/o. joint) & 11.39 & 59.03 & 15.98 & 43.95 & 35.28 & 0.36 & 0.16 & \textbf{5.26} & 0.24 & \textbf{33.51}\\
				Ours (joint) & \textbf{14.51} & \textbf{60.65} & \textbf{17.55} & \textbf{44.90} & \textbf{36.48} & \textbf{0.69} & \textbf{0.62} & 4.93 & \textbf{0.37} & 32.08\\
				\hline
				\hline
				Method & blinds & desk & shelves & curtain & dresser & pillow & mirror & floor mat & clothes & books\\
				\hline\hline
				CooP \cite{huang2018cooperative}\textsuperscript{**} & \textbf{2.27} & 19.90 & 2.96 & 1.35 & 15.98 & 2.53 & \textbf{0.47} & - & 0.00 & \textbf{3.19}\\
				Ours (w/o. joint) & 0.00 & 23.65 & \textbf{4.96} & 2.68 & 19.20 & 2.99 & 0.19 & - & 0.00 & 1.30\\
				Ours (joint) & 0.00 & \textbf{27.93} & 3.70 & \textbf{3.04} & \textbf{21.19} & \textbf{4.46} & 0.29 & - & 0.00 & 2.02\\
				\hline
				\hline
				Method & fridge & tv & paper & towel & shower curtain & box & whiteboard & person & nightstand & toilet\\
				\hline\hline
				CooP \cite{huang2018cooperative}\textsuperscript{**} & 21.50 & 5.20 & 0.20 & 2.14 & 20.00 & \textbf{2.59} & 0.16 & 20.96 & 11.36 & 42.53\\
				Ours (w/o. joint) & 20.68 & 4.44 & 0.41 & \textbf{2.20} & 20.00 & 2.25 & 0.43 & 23.36 & 6.87 & \textbf{48.37}\\
				Ours (joint) & \textbf{24.42} & \textbf{5.60} & \textbf{0.97} & 2.07 & 20.00 & 2.46 & \textbf{0.61} & \textbf{31.29} & \textbf{17.01} & 44.24 \\
				\hline
				\hline
				Method & sink & lamp & bathtub & bag & wall & floor & ceiling &  &  & \\
				\hline\hline
				CooP \cite{huang2018cooperative}\textsuperscript{**} & 15.95 & 3.28 & 24.71 & 1.53 & - & - & - &  &  & \\
				Ours (w/o. joint) & 14.40 & 3.46 & \textbf{27.85} & 2.27 & - & - & - &  &  & \\
				Ours (joint) & \textbf{18.50} & \textbf{5.04} & 21.15 & \textbf{2.47} & - & - & - &  &  & \\
				\hline
		\end{tabular}}
	\end{center}
	\caption{Comparison of 3D object detection. We compare the average precision (AP) of detected objects on SUN RGB-D (higher is better). CooP \cite{huang2018cooperative}\textsuperscript{**} presents the model trained on the NYU-37 object labels for a fair comparison.}
	\label{compare:more3ddetection}
\end{table*}

\section{Object Class Mapping}
Pix3D has nine object categories for mesh reconstruction, which contains: 1. bed, 2. bookcase, 3. chair, 4. desk, 5. sofa, 6. table, 7. tool, 8. wardrobe, 9. miscellaneous. In 3D object detection, we obtain object bounding boxes with NYU-37 labels in SUN RGB-D. As our MGN is pretrained on Pix3D, and the object class code is required as an input for mesh deformation, we manually map the NYU-37 labels to Pix3D labels based on topology similarity for scene reconstruction (see Table~\ref{label_map}).

\section{More Comparisons of Object Mesh Reconstruction on Pix3D}
\begin{figure*}[!h]
	\centering
	\begin{subfigure}[t]{0.15\textwidth}
		\includegraphics[width=\textwidth]  
		{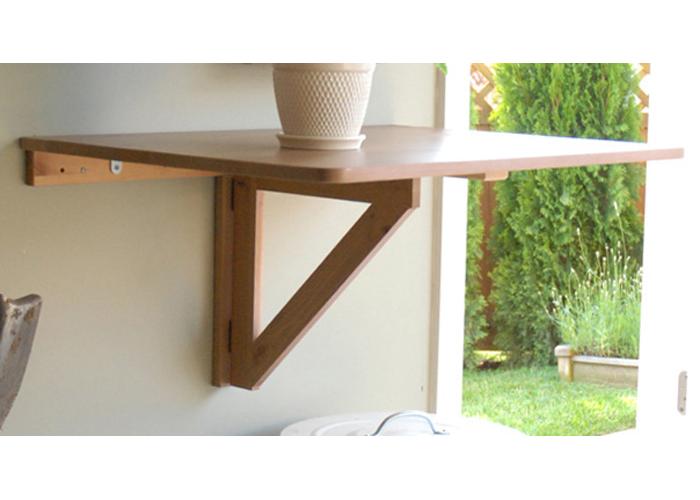}
		\includegraphics[width=\textwidth]  
		{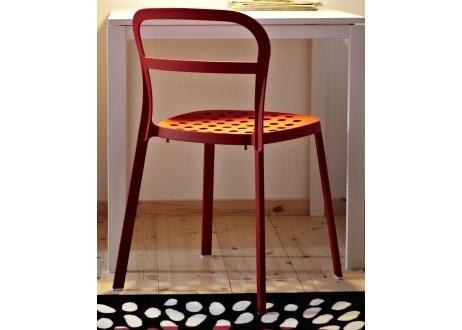}
		\includegraphics[width=\textwidth]  
		{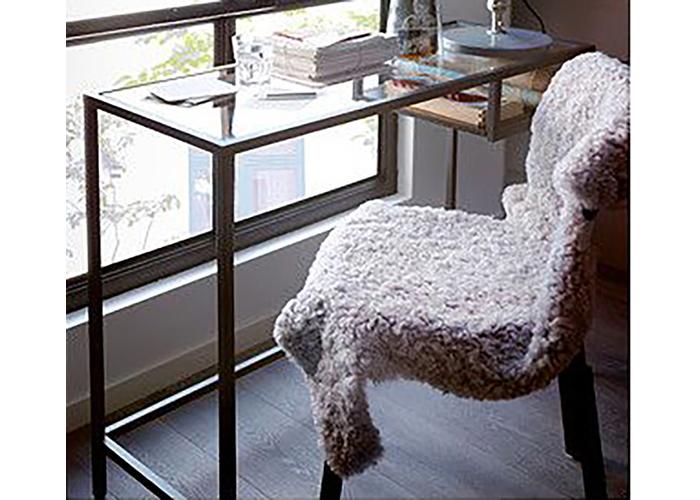}
		\includegraphics[width=\textwidth]  
		{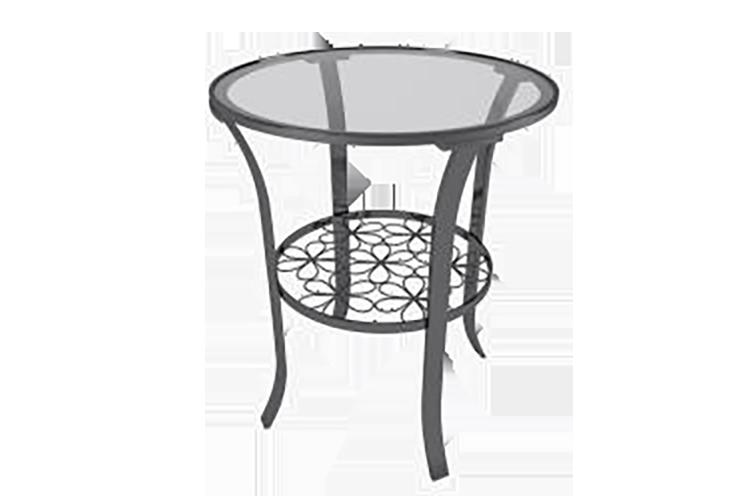}
		\includegraphics[width=\textwidth]  
		{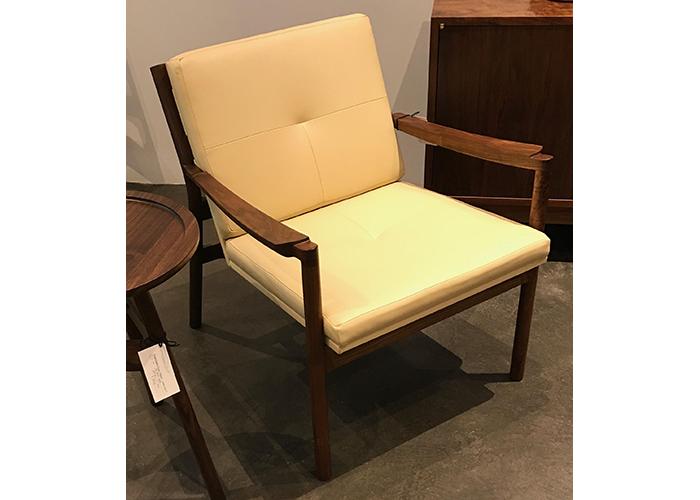}
	\end{subfigure}
	\begin{subfigure}[t]{0.15\textwidth}
		\includegraphics[width=\textwidth]  
		{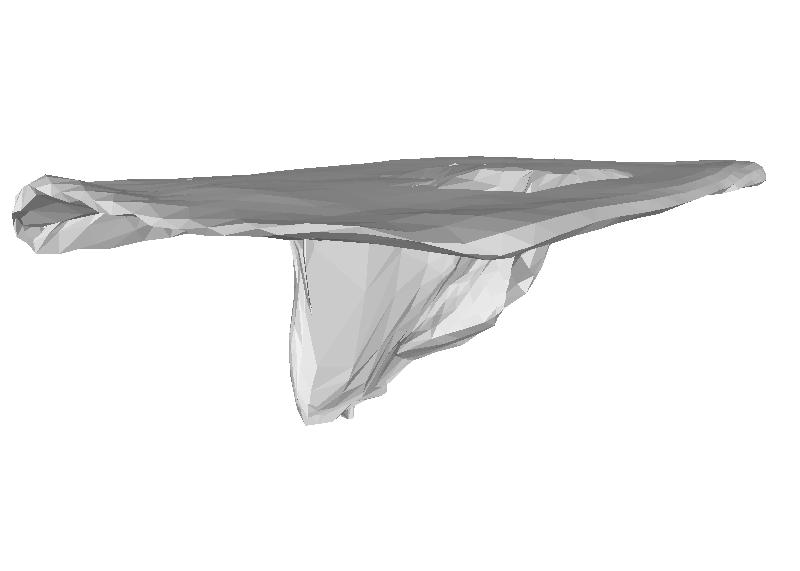}
		\includegraphics[width=\textwidth]  
		{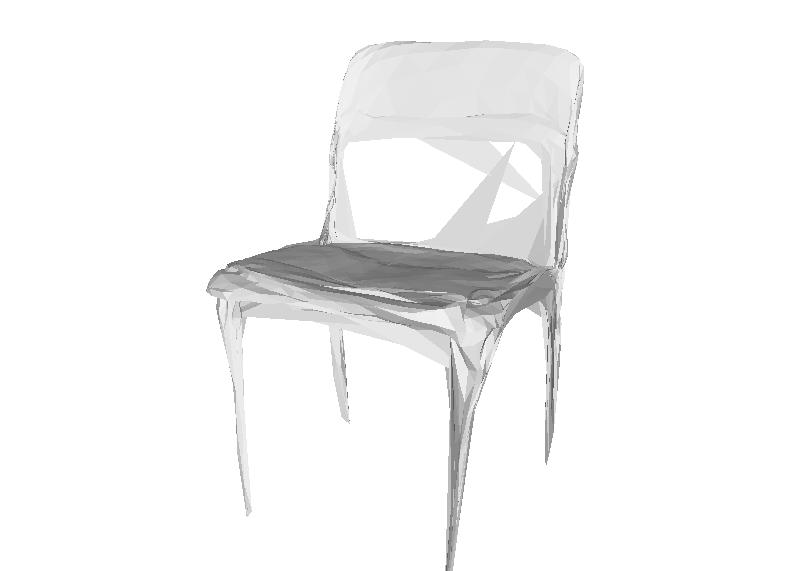}
		\includegraphics[width=\textwidth]  
		{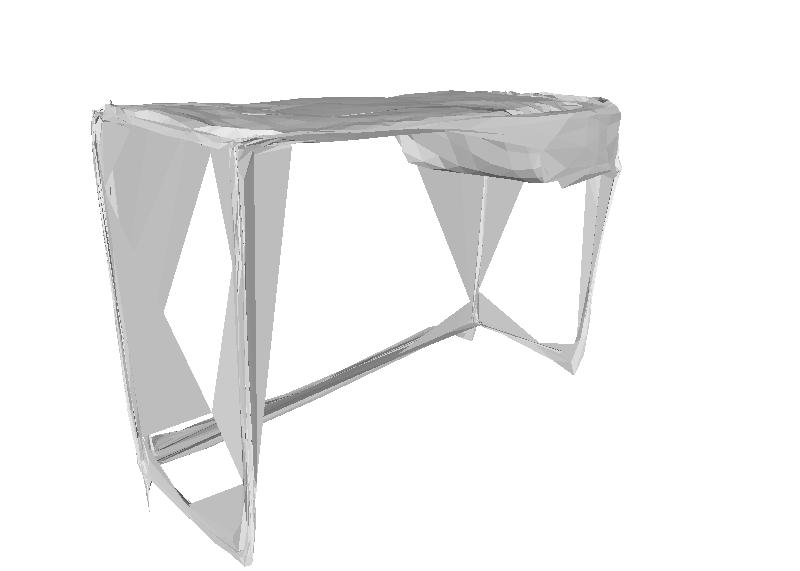}
		\includegraphics[width=\textwidth]  
		{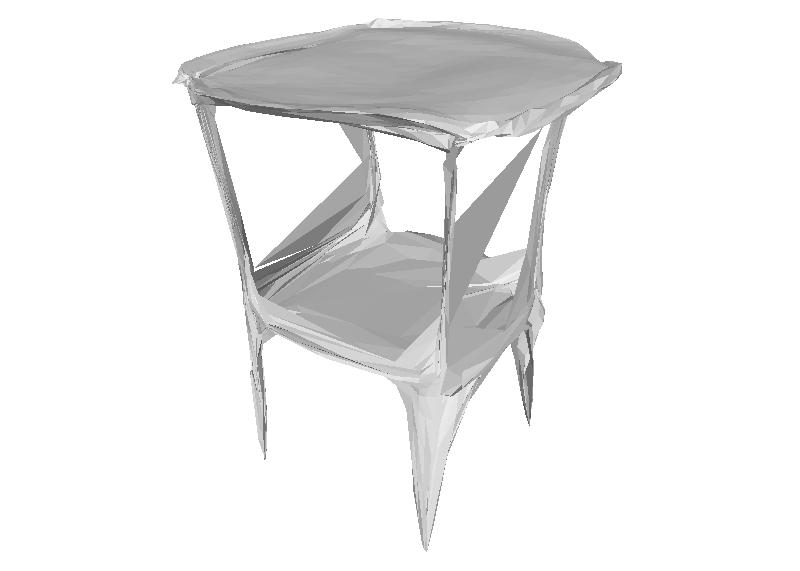}
		\includegraphics[width=\textwidth]  
		{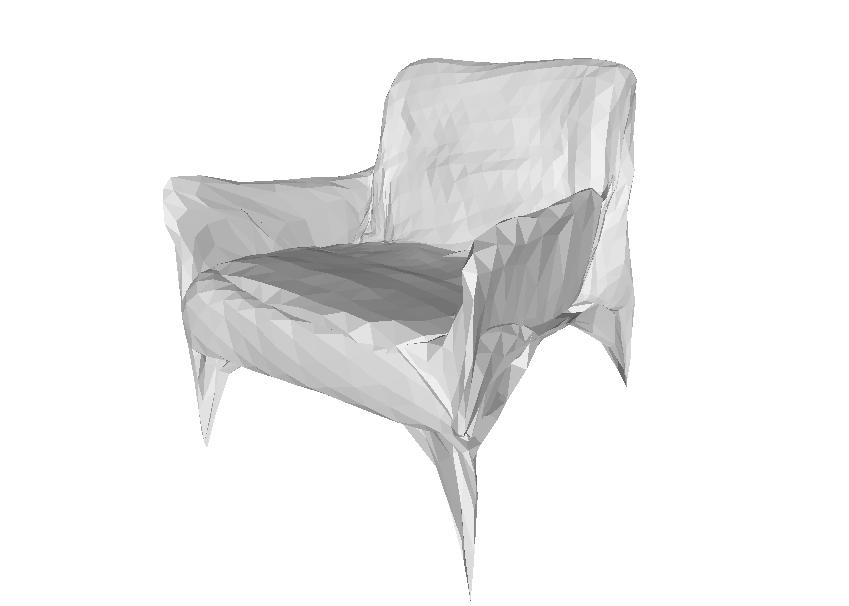}
	\end{subfigure}
	\begin{subfigure}[t]{0.15\textwidth}
		\includegraphics[width=\textwidth]  
		{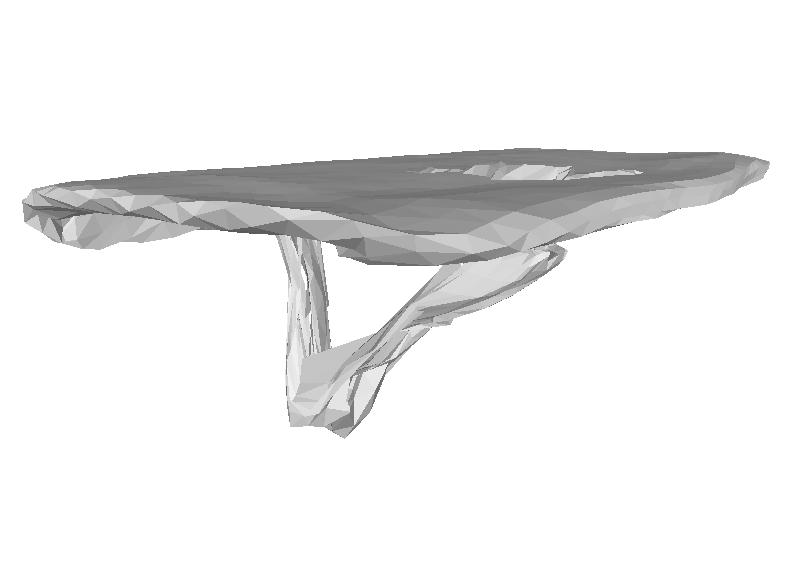}
		\includegraphics[width=\textwidth]  
		{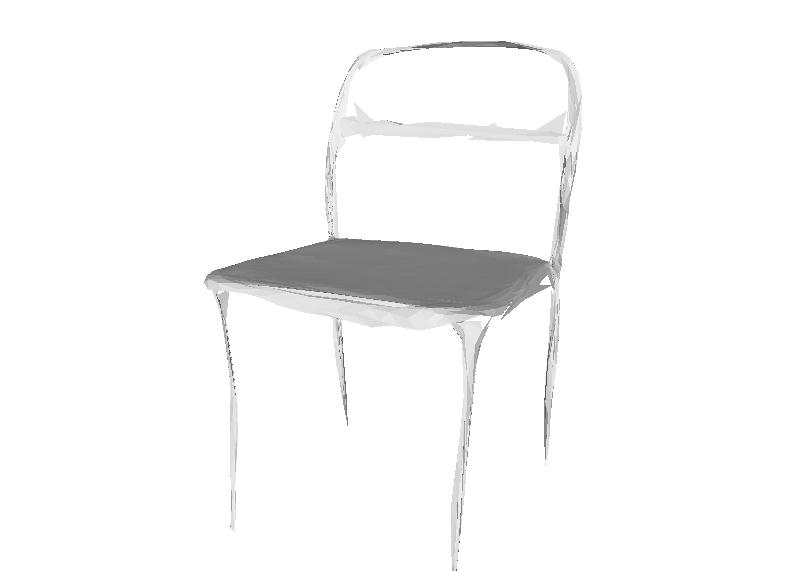}
		\includegraphics[width=\textwidth]  
		{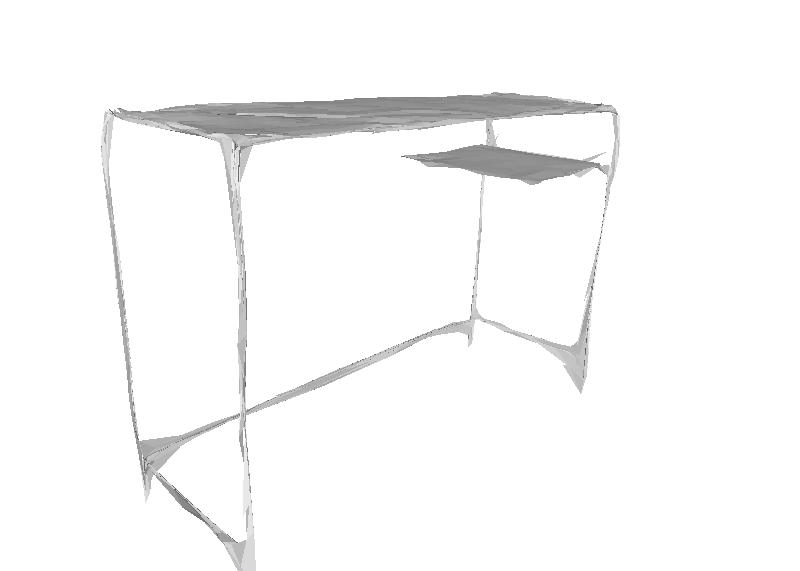}
		\includegraphics[width=\textwidth]  
		{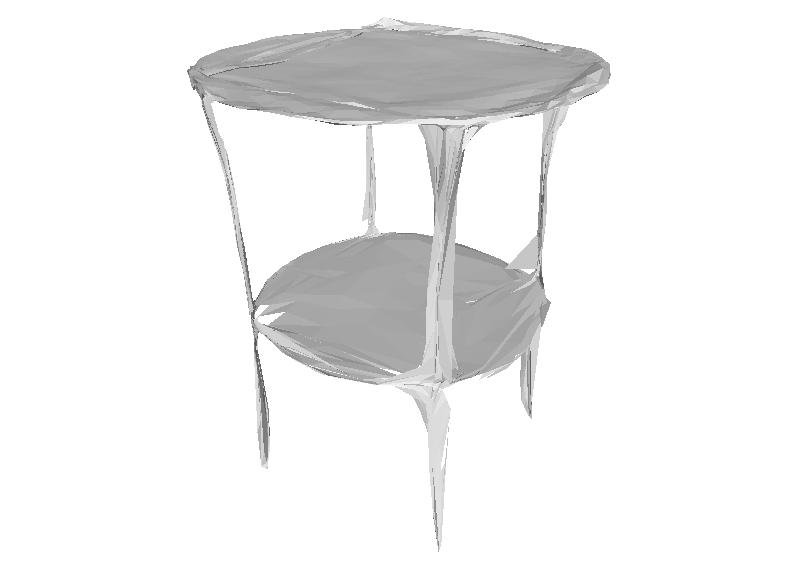}
		\includegraphics[width=\textwidth]  
		{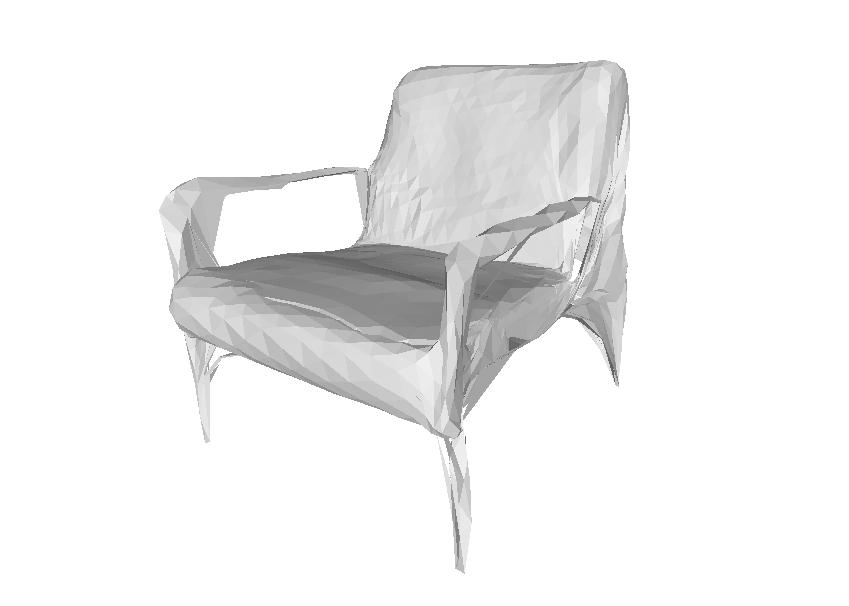}
	\end{subfigure}
	\rulesep
	\begin{subfigure}[t]{0.15\textwidth}
		\includegraphics[width=\textwidth]  
		{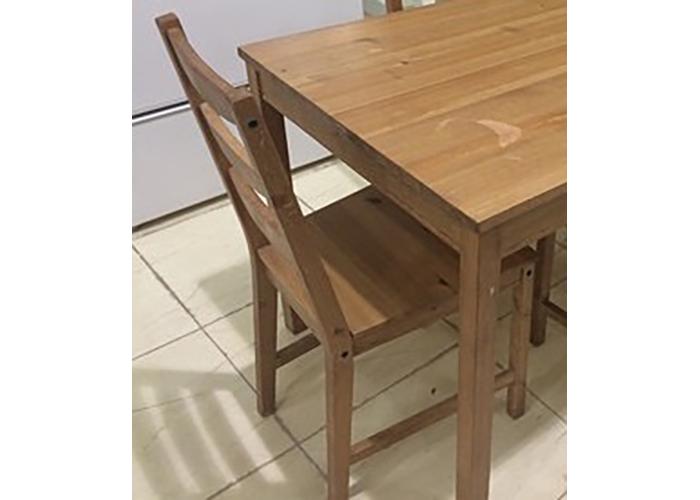}
		\includegraphics[width=\textwidth]  
		{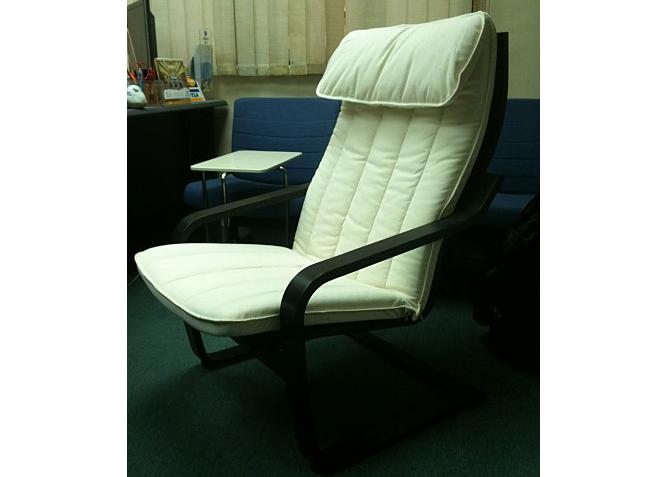}
		\includegraphics[width=\textwidth]  
		{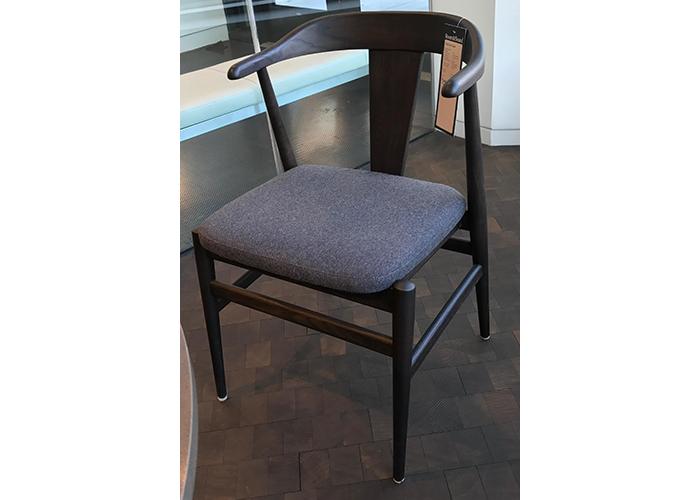}
		\includegraphics[width=\textwidth]  
		{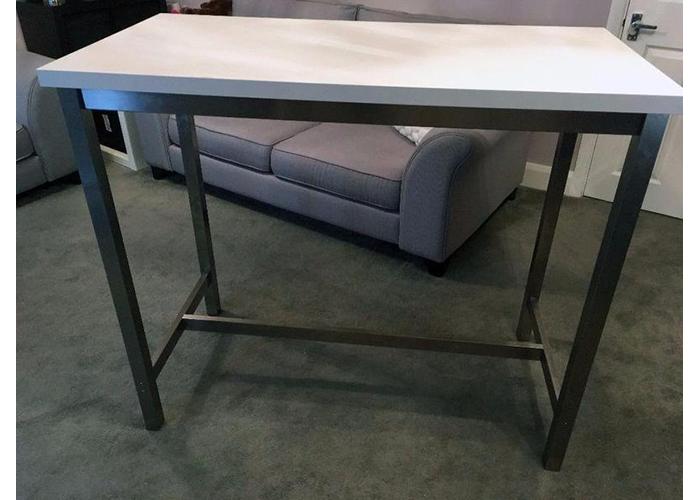}
		\includegraphics[width=\textwidth]  
		{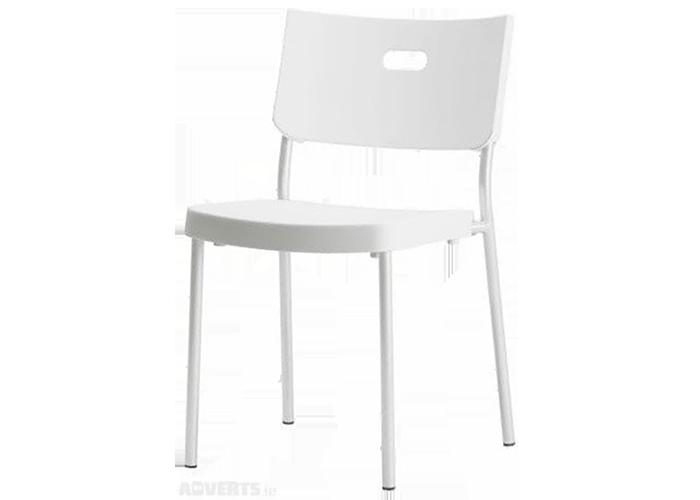}
	\end{subfigure}
	\begin{subfigure}[t]{0.15\textwidth}
		\includegraphics[width=\textwidth] 
		{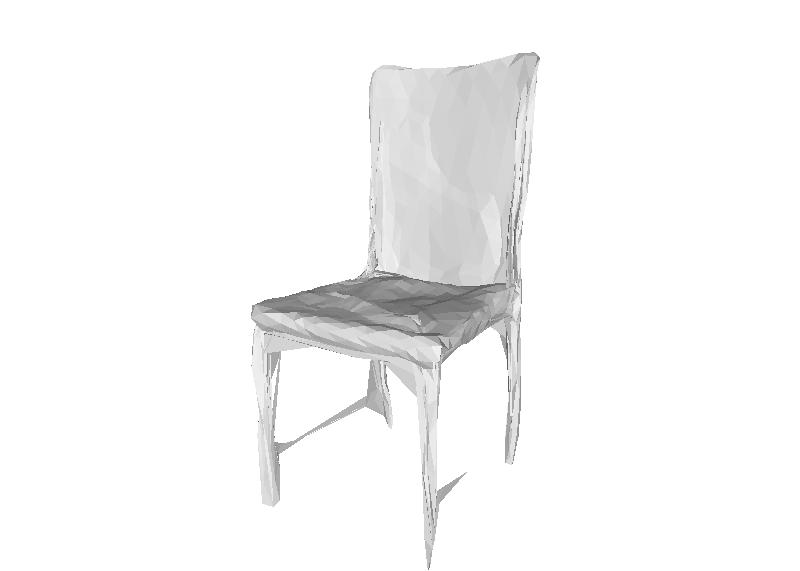}
		\includegraphics[width=\textwidth]  
		{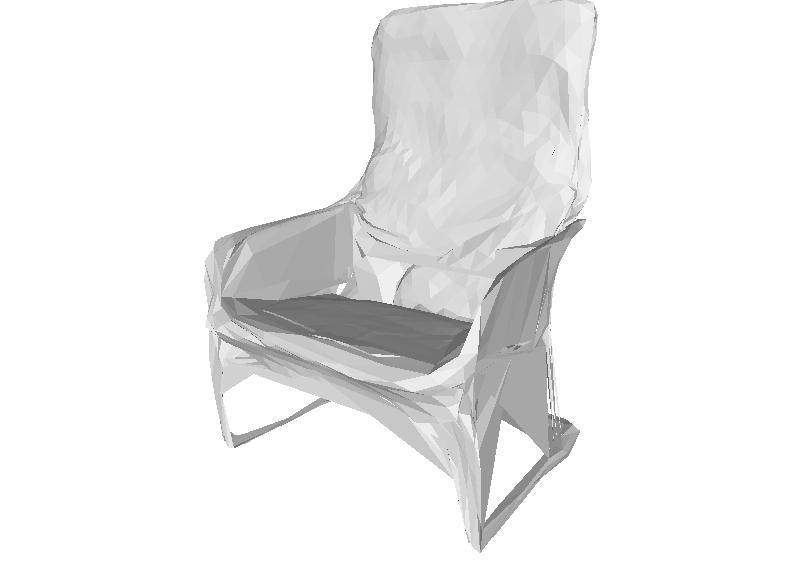}
		\includegraphics[width=\textwidth]  
		{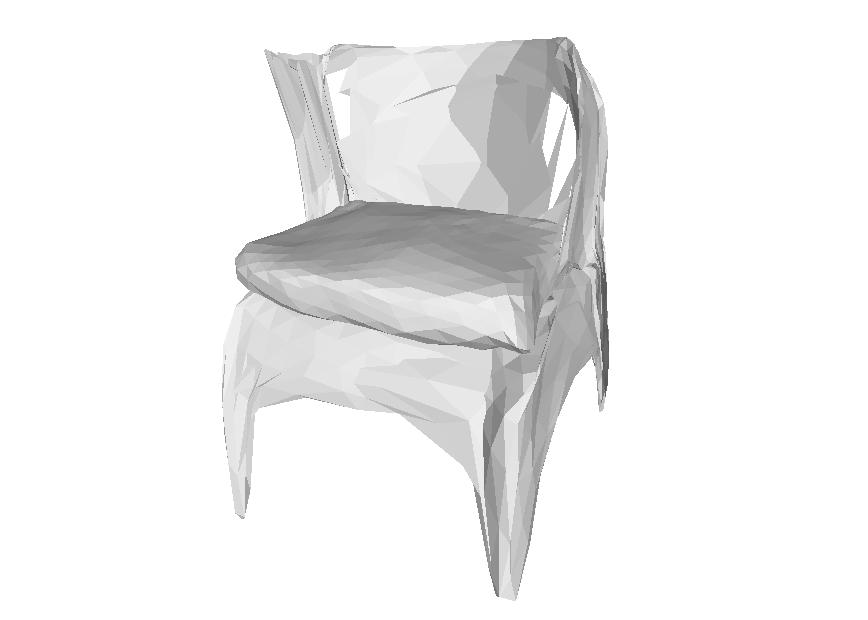}
		\includegraphics[width=\textwidth]  
		{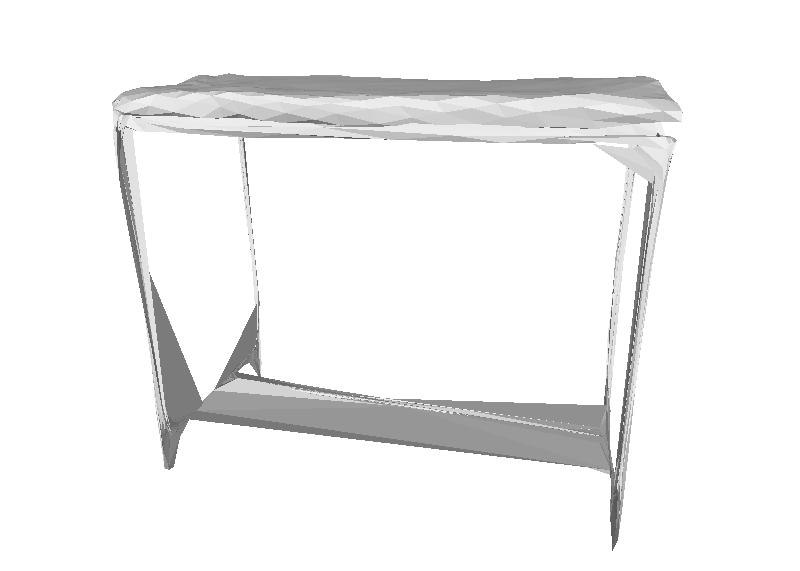}
		\includegraphics[width=\textwidth]  
		{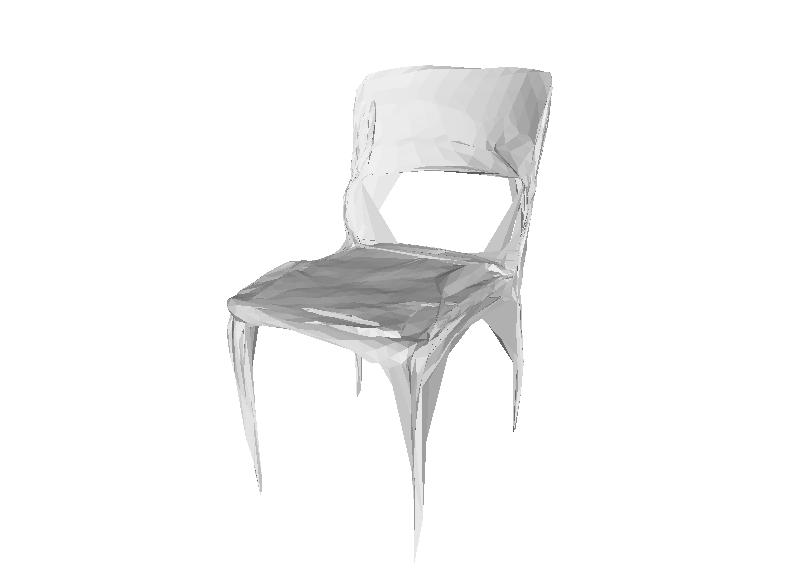}
	\end{subfigure}
	\begin{subfigure}[t]{0.15\textwidth}
		\includegraphics[width=\textwidth]  
		{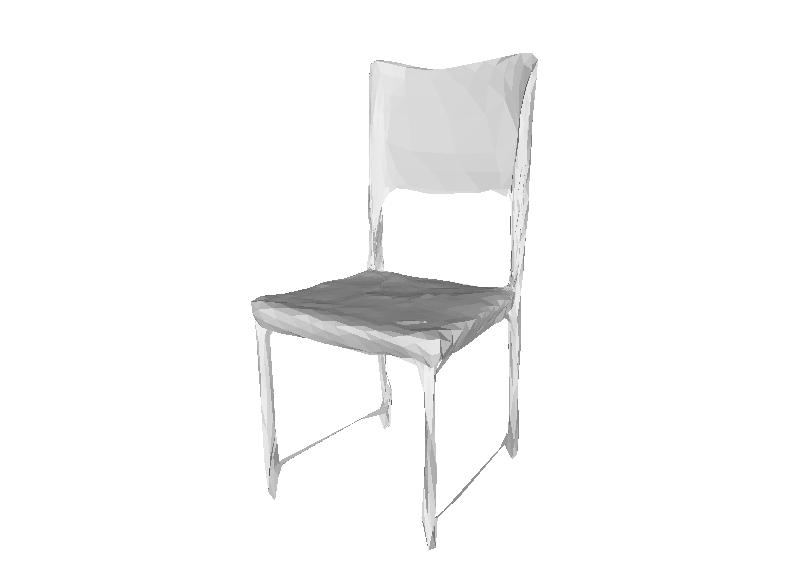}
		\includegraphics[width=\textwidth]  
		{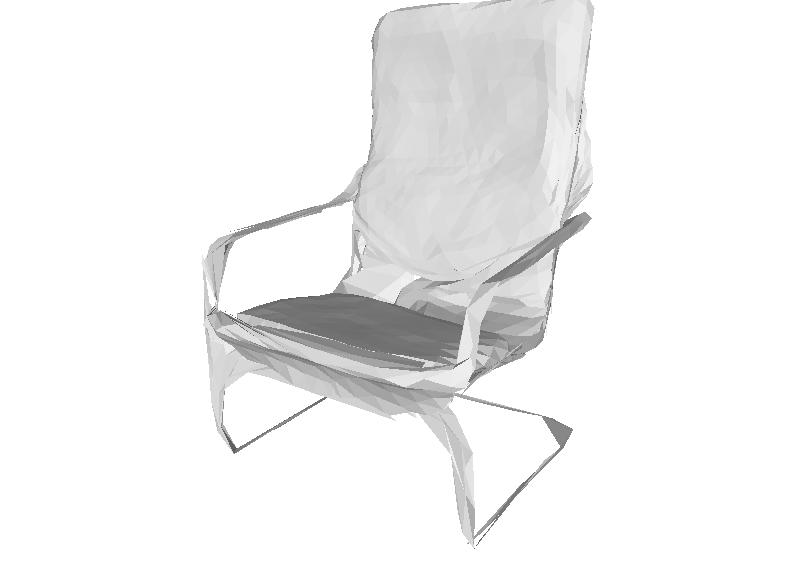}
		\includegraphics[width=\textwidth]  
		{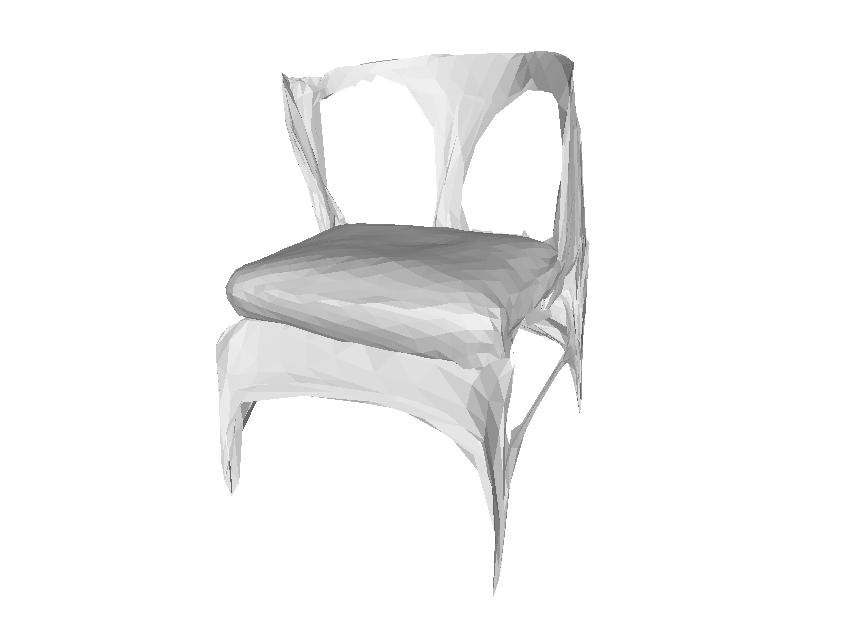}
		\includegraphics[width=\textwidth]  
		{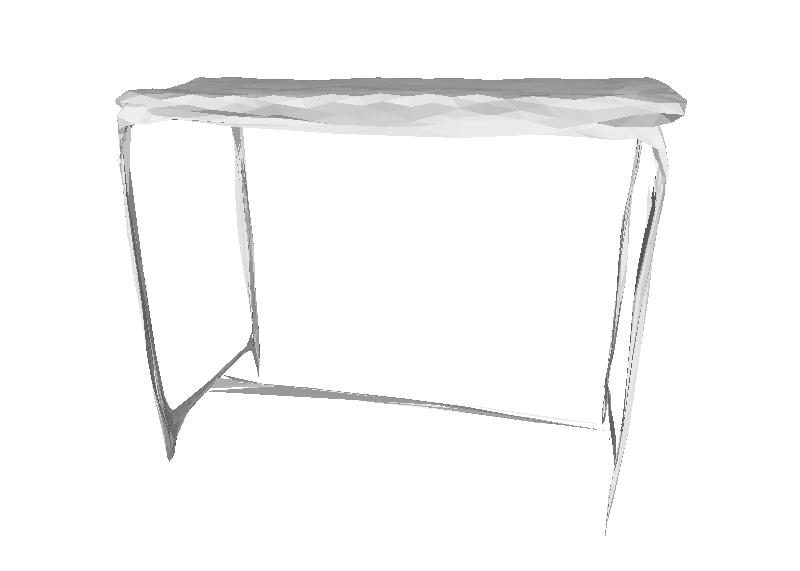}
		\includegraphics[width=\textwidth]  
		{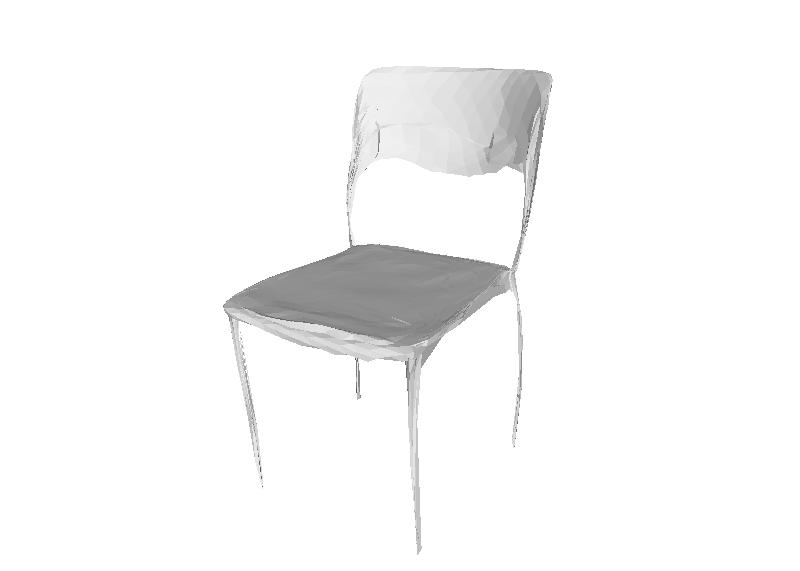}
	\end{subfigure}
	\caption{Qualitative comparisons between the proposed method and TMN \cite{Junyi} on object mesh reconstruction. From left to right: input images, results from TMN, and our results.}
	\label{compare_TMN}
\end{figure*}

More qualitative comparisons with Topology Modification Network (TMN) \cite{Junyi} are shown in Figure~\ref{compare_TMN}. The threshold $\tau$ in TMN is set at 0.1 to be consistent with their paper.

\section{More Samples of Scene Reconstruction on SUN RGB-D}
\begin{figure*}[!ht]
	\centering
	\begin{subfigure}[t]{0.15\textwidth}
		\includegraphics[width=\textwidth]
		{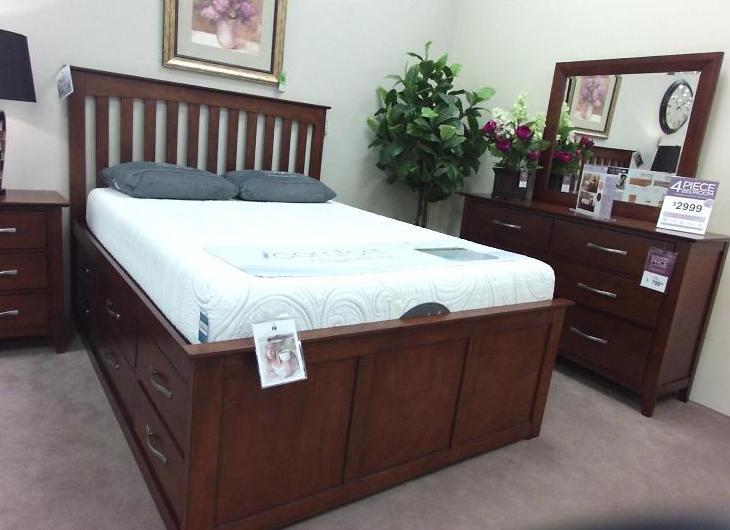}
		\includegraphics[width=\textwidth]
		{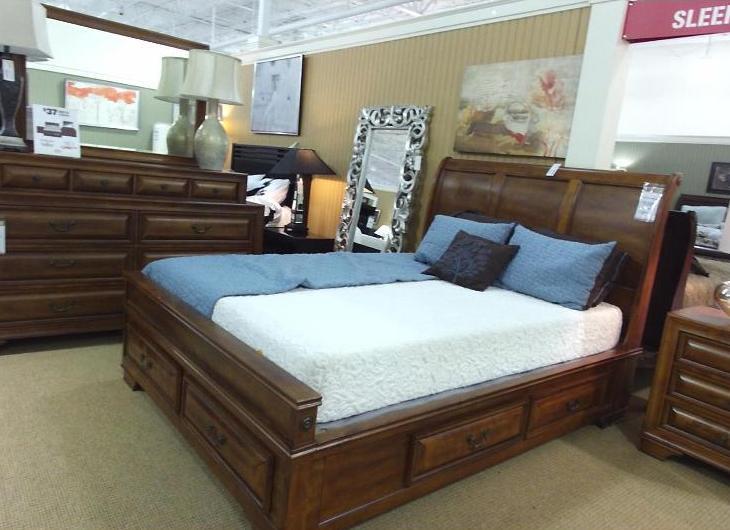}
		\includegraphics[width=\textwidth]
		{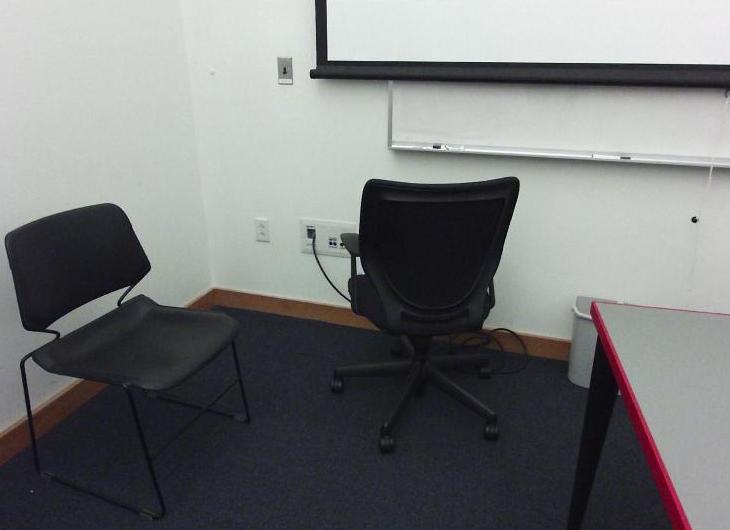}
		\includegraphics[width=\textwidth]  
		{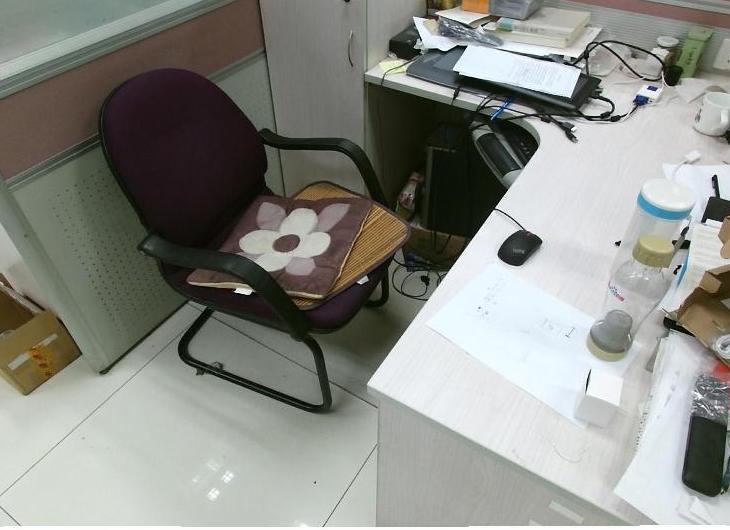}
		\includegraphics[width=\textwidth]
		{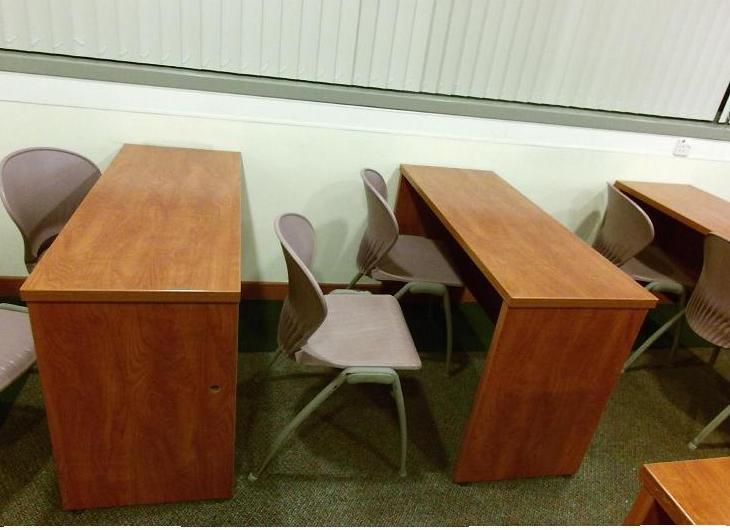}
		\includegraphics[width=\textwidth]
		{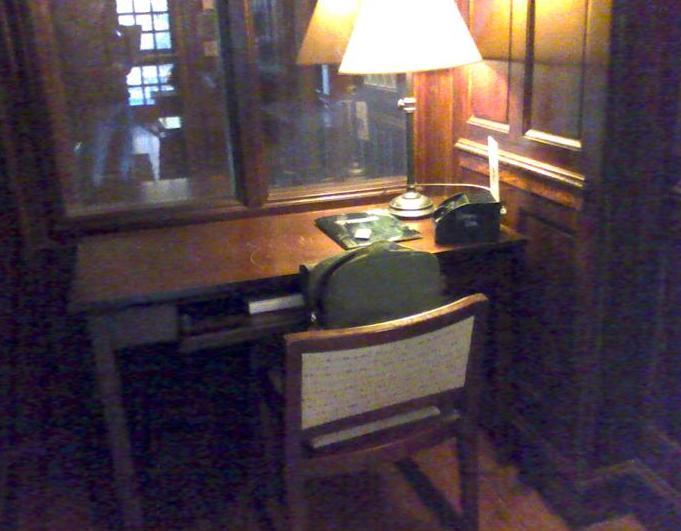}
		\includegraphics[width=\textwidth]
		{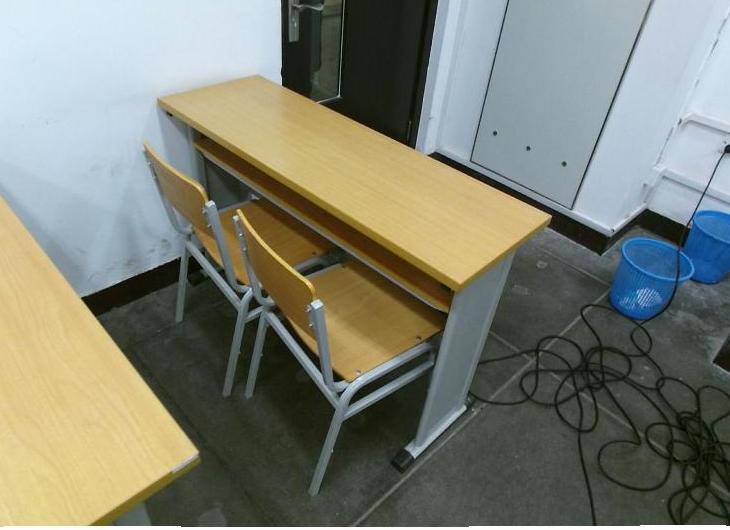}
		\includegraphics[width=\textwidth]
		{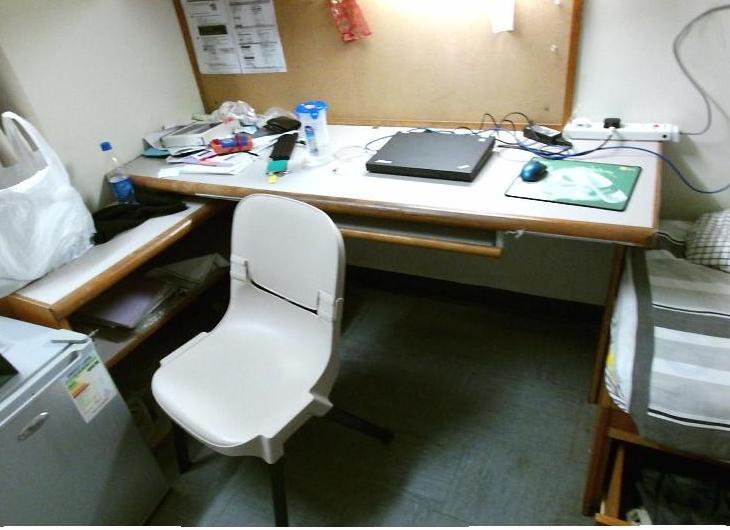}
		\includegraphics[width=\textwidth]
		{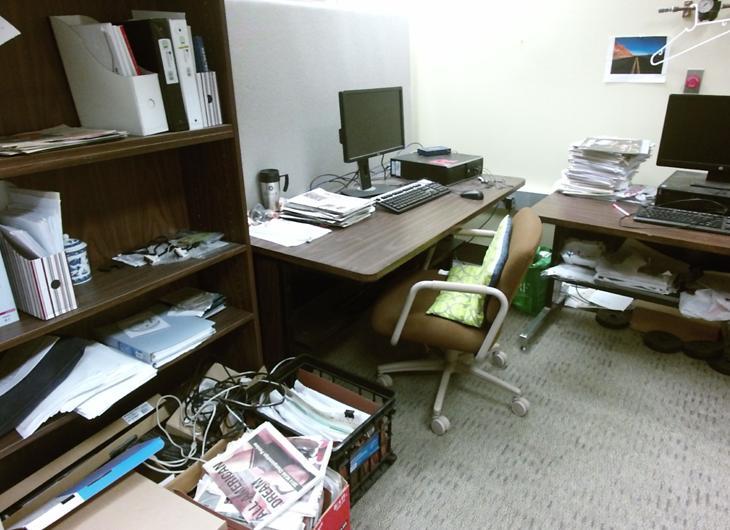}
		\includegraphics[width=\textwidth]
		{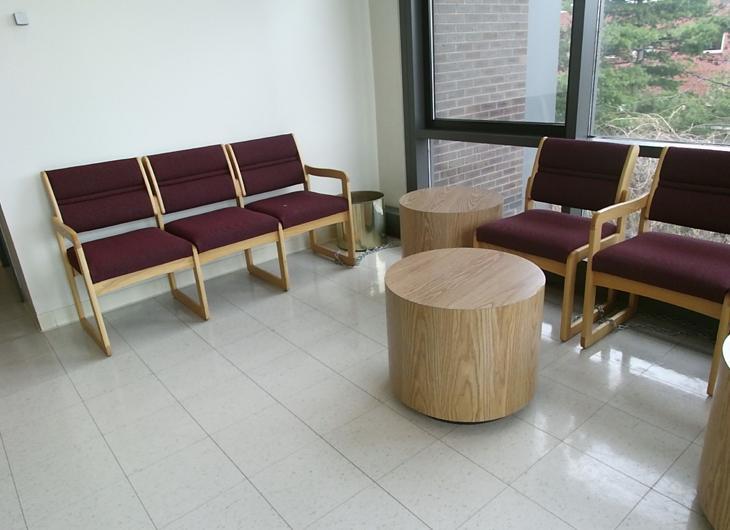}
		\includegraphics[width=\textwidth]
		{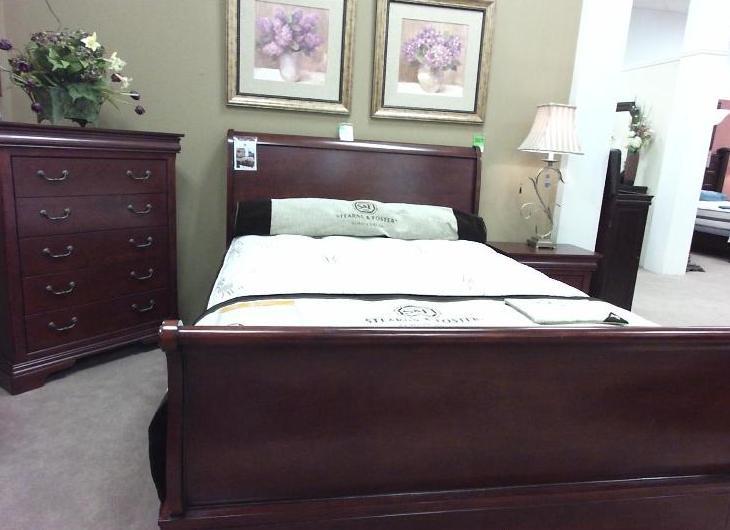}
	\end{subfigure}
	\begin{subfigure}[t]{0.15\textwidth}
		\includegraphics[width=\textwidth]
		{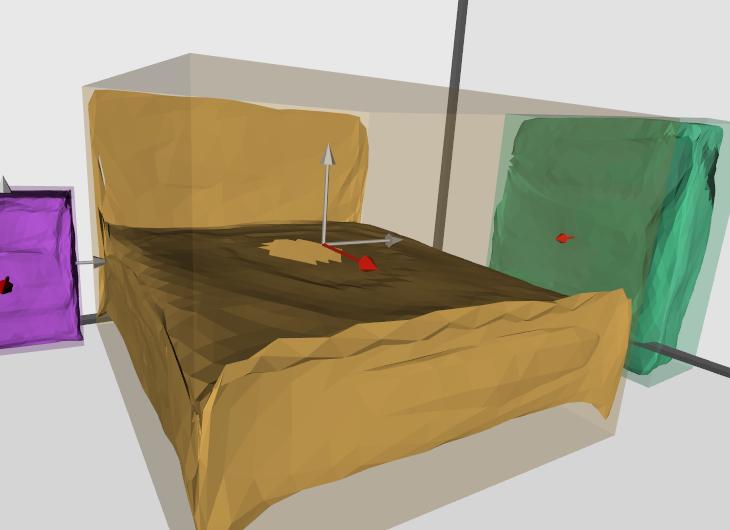}
		\includegraphics[width=\textwidth]
		{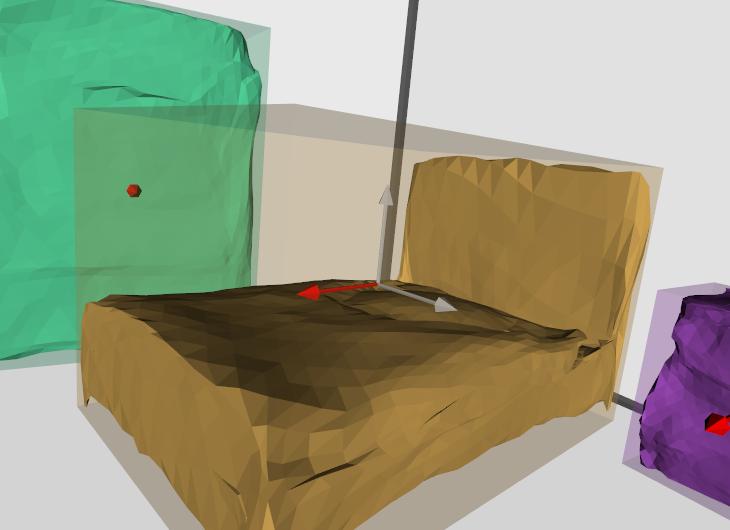}
		\includegraphics[width=\textwidth]
		{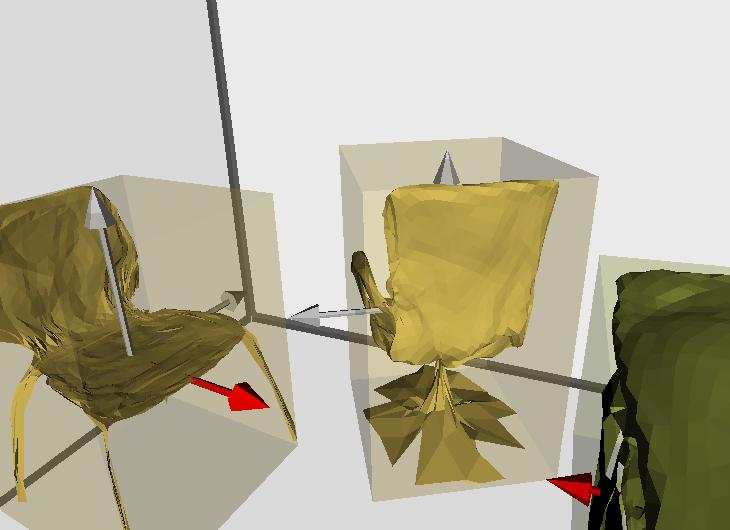}
		\includegraphics[width=\textwidth]  
		{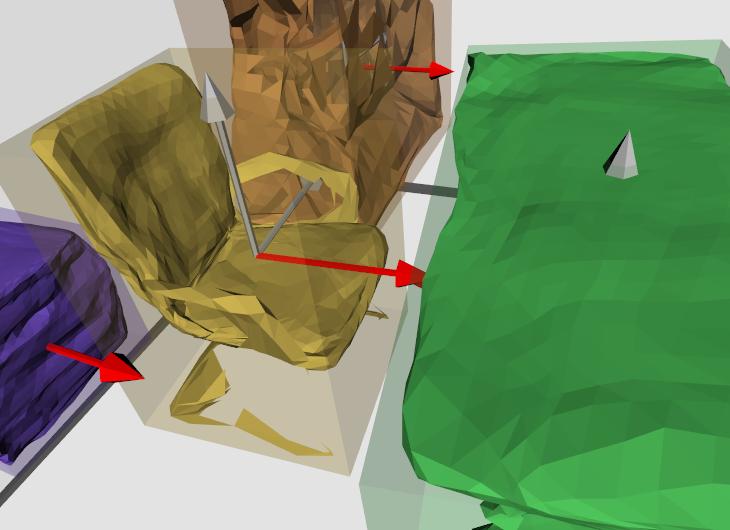}
		\includegraphics[width=\textwidth]
		{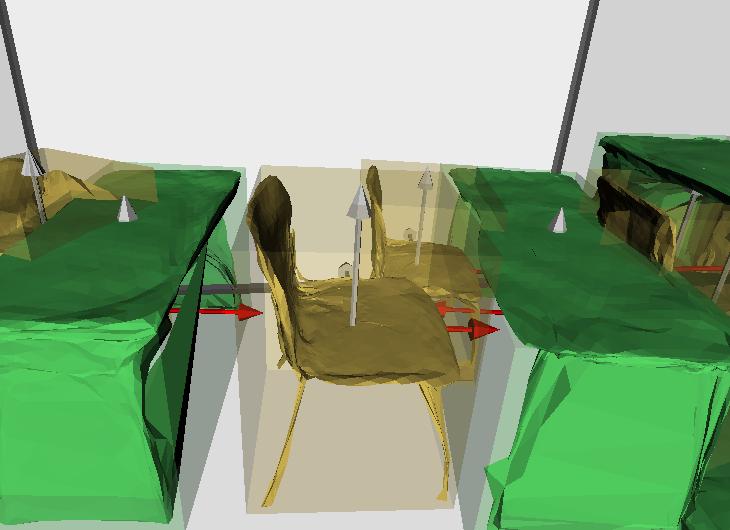}
		\includegraphics[width=\textwidth]
		{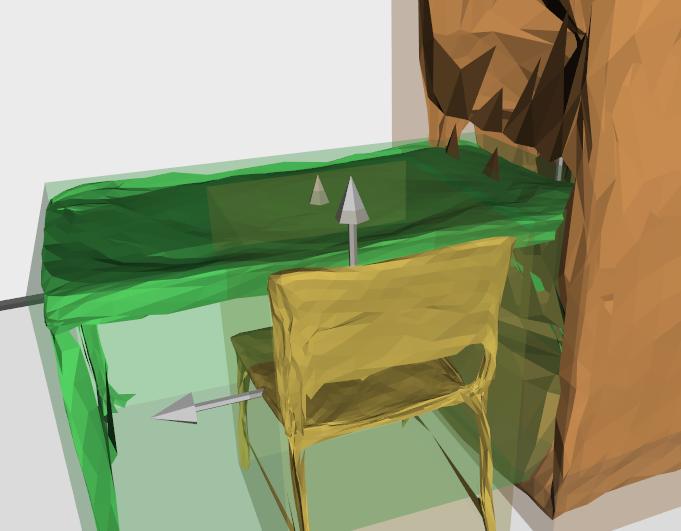}
		\includegraphics[width=\textwidth]
		{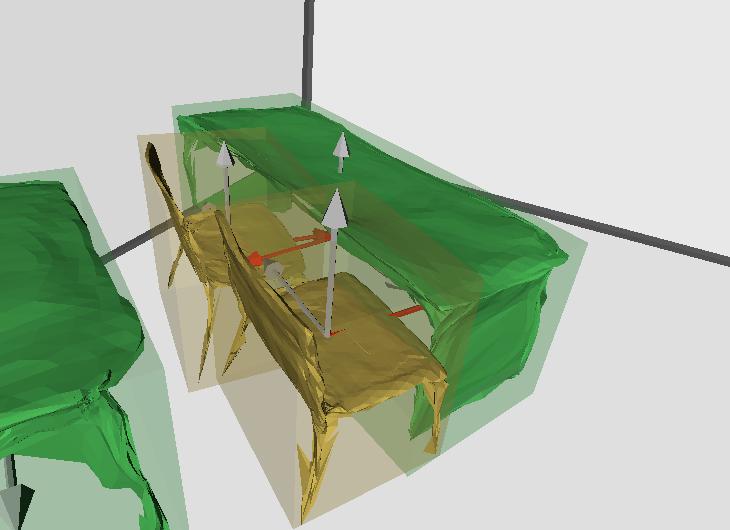}
		\includegraphics[width=\textwidth]
		{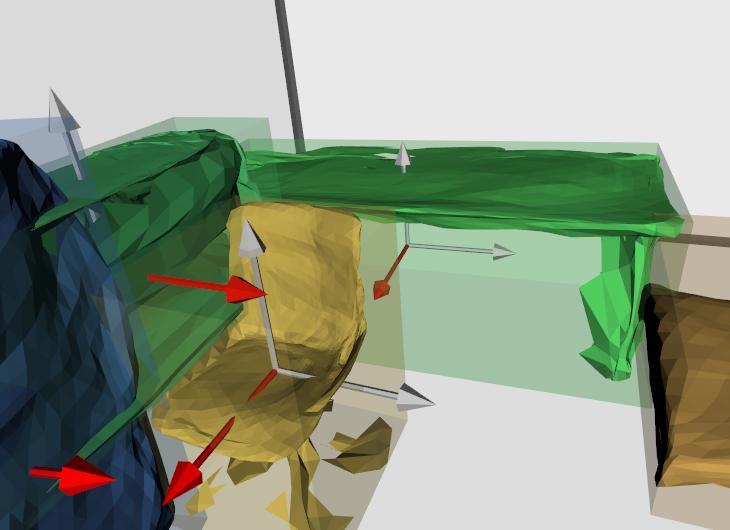}
		\includegraphics[width=\textwidth]
		{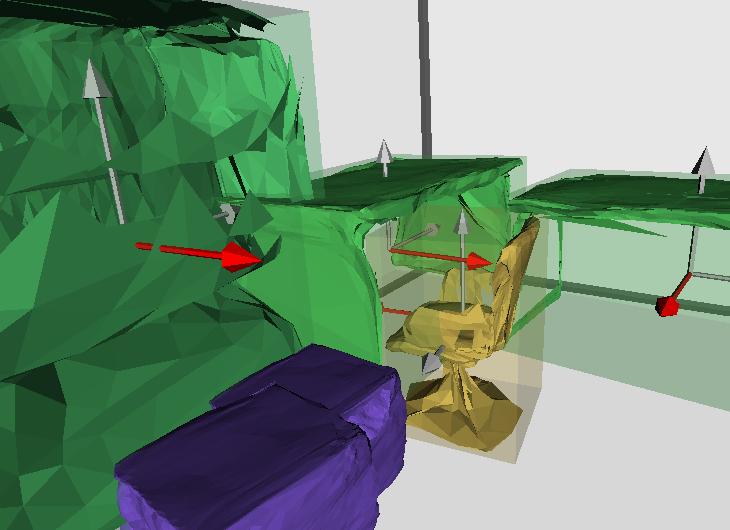}
		\includegraphics[width=\textwidth]
		{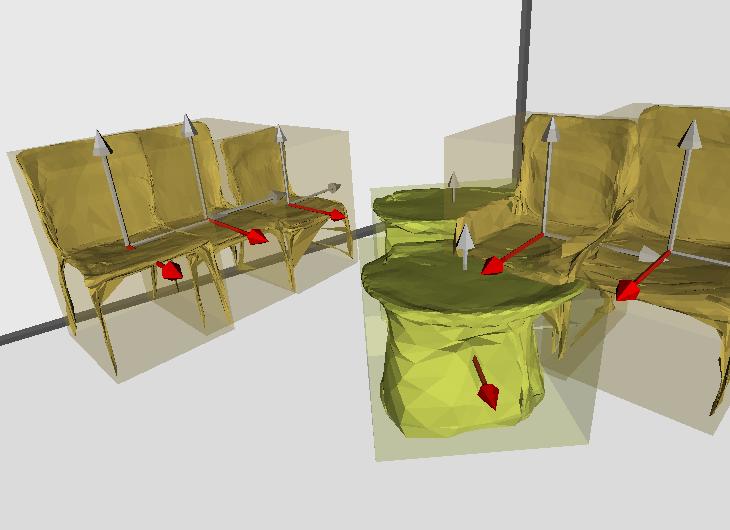}
		\includegraphics[width=\textwidth]
		{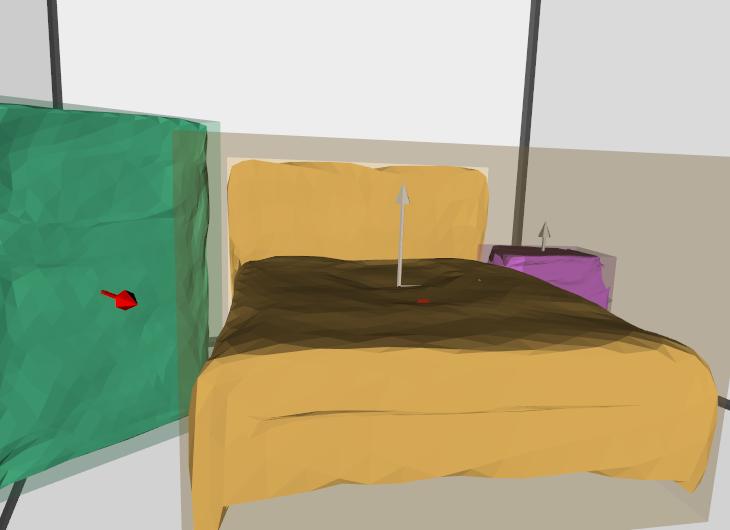}
	\end{subfigure}
	\begin{subfigure}[t]{0.15\textwidth}
		\includegraphics[width=\textwidth]
		{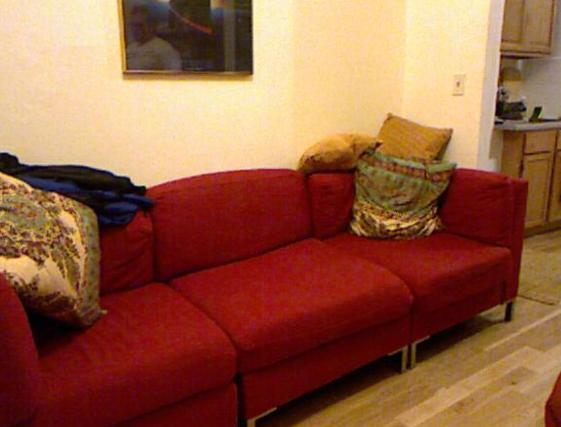}
		\includegraphics[width=\textwidth]
		{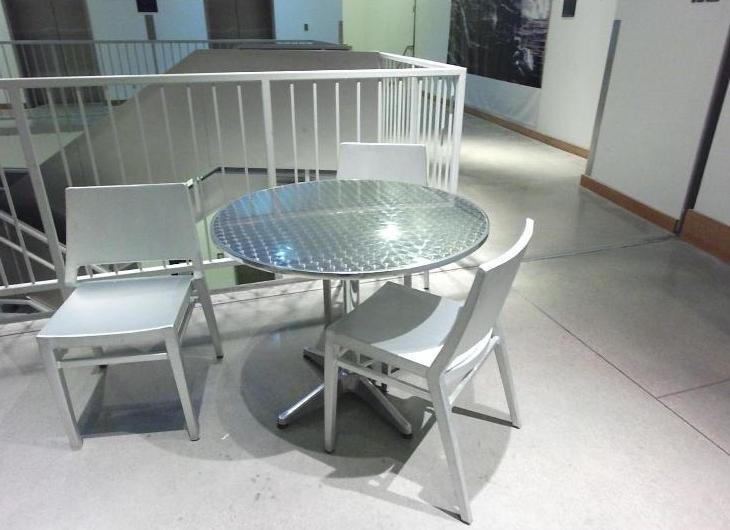}
		\includegraphics[width=\textwidth]
		{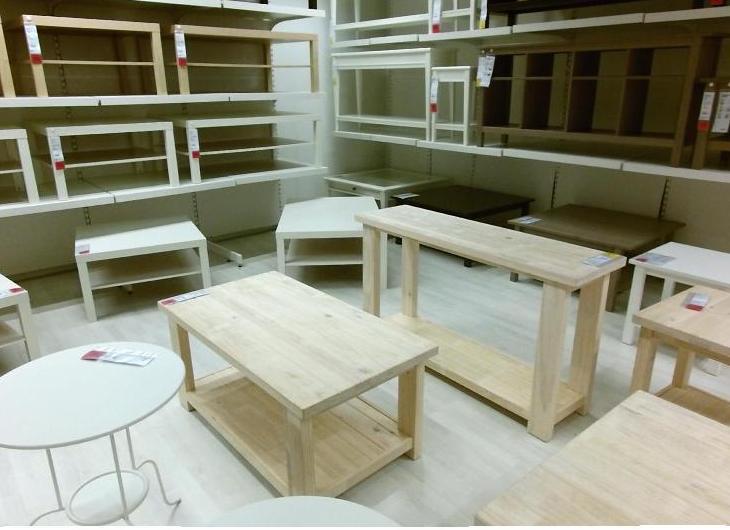}
		\includegraphics[width=\textwidth]  
		{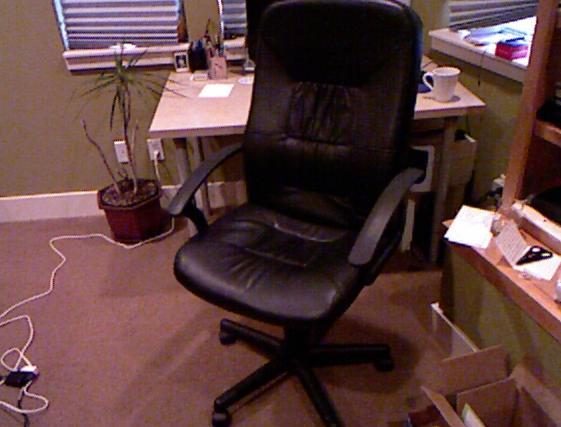}
		\includegraphics[width=\textwidth]
		{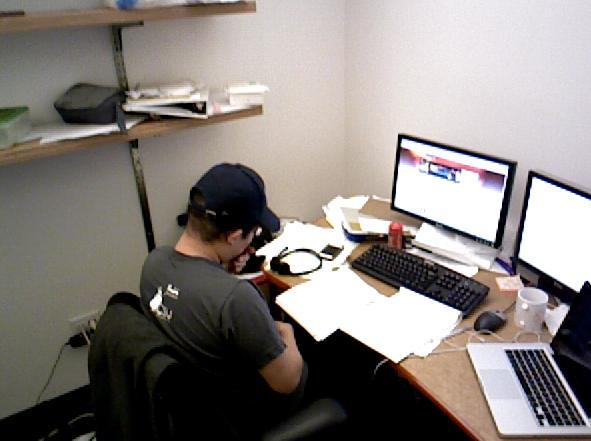}
		\includegraphics[width=\textwidth]
		{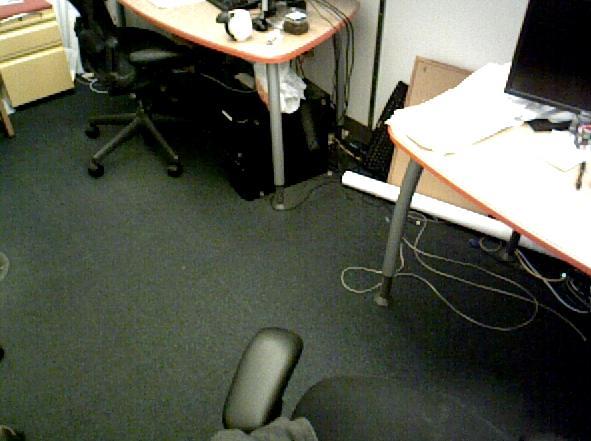}
		\includegraphics[width=\textwidth]
		{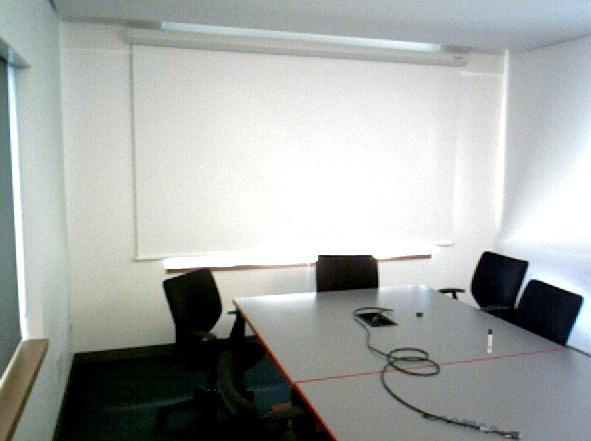}
		\includegraphics[width=\textwidth]
		{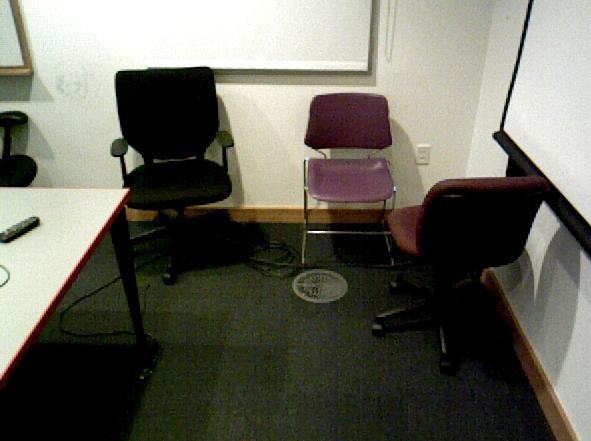}
		\includegraphics[width=\textwidth]
		{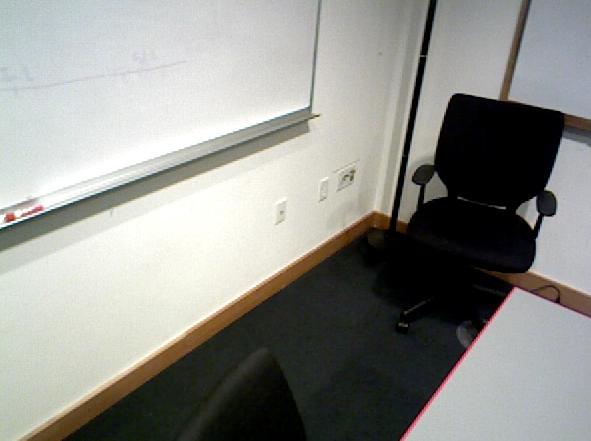}
		\includegraphics[width=\textwidth]
		{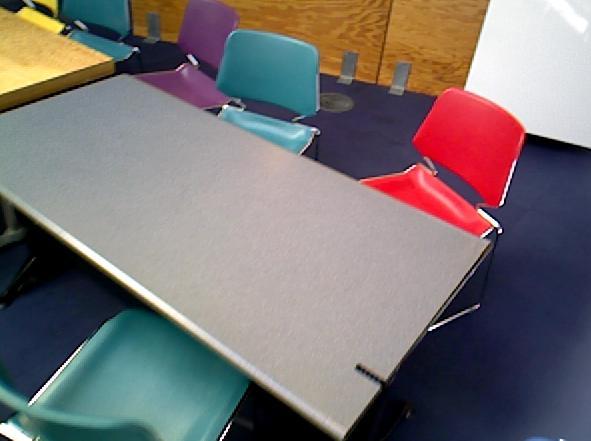}
		\includegraphics[width=\textwidth]
		{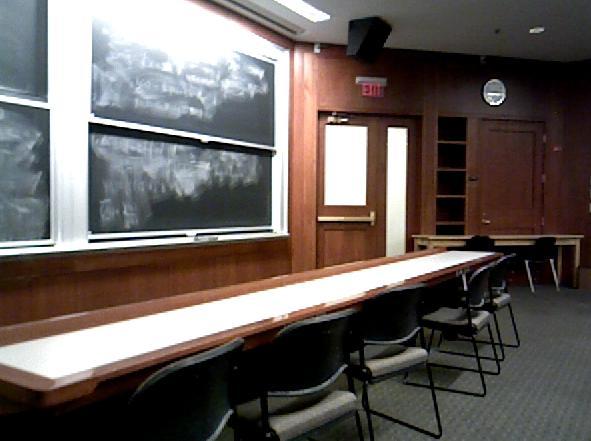}
	\end{subfigure}
	\begin{subfigure}[t]{0.15\textwidth}
		\includegraphics[width=\textwidth]
		{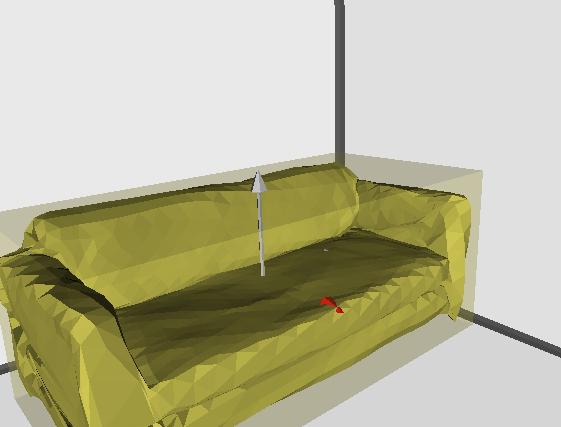}
		\includegraphics[width=\textwidth]
		{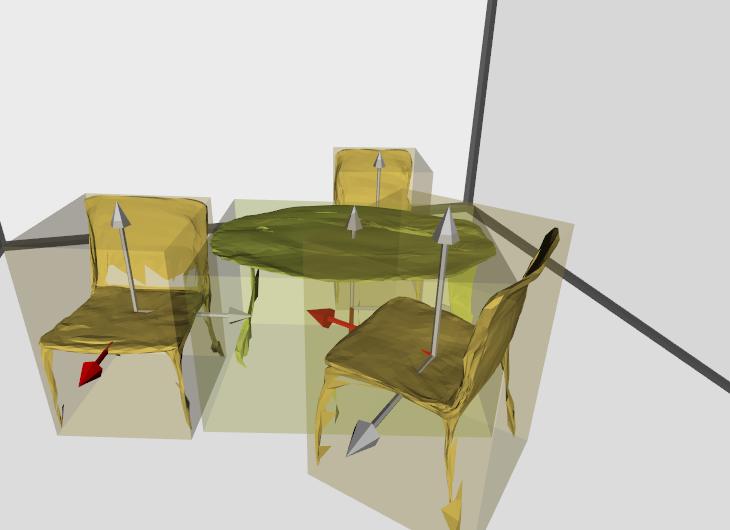}
		\includegraphics[width=\textwidth]
		{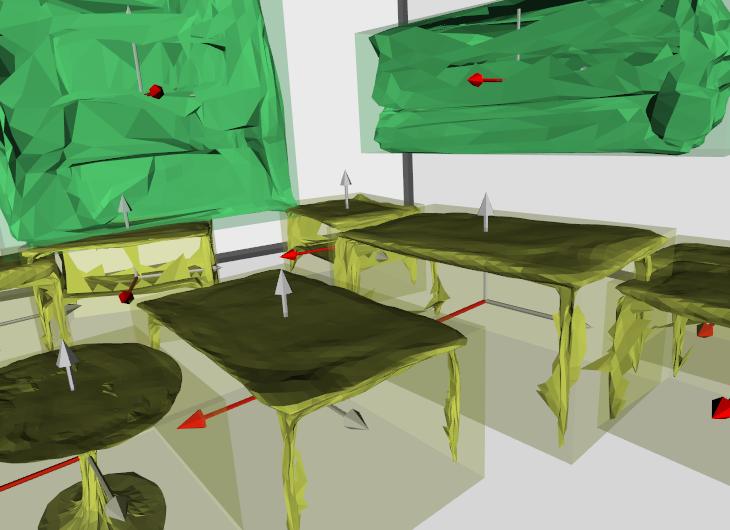}
		\includegraphics[width=\textwidth]  
		{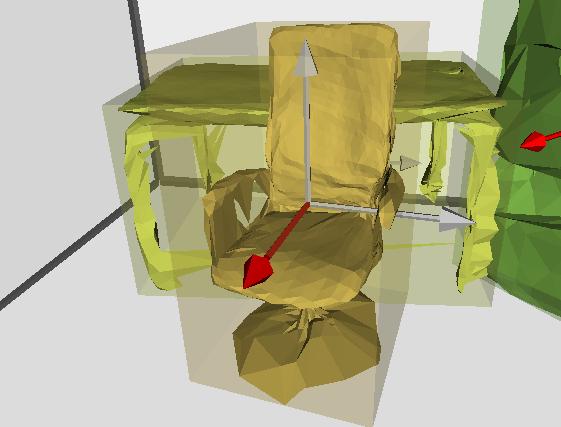}
		\includegraphics[width=\textwidth]
		{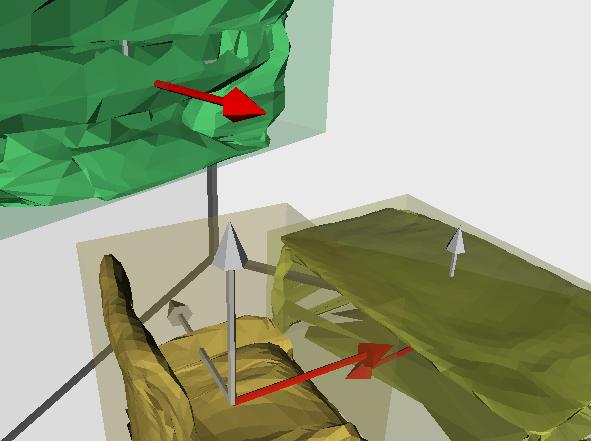}
		\includegraphics[width=\textwidth]
		{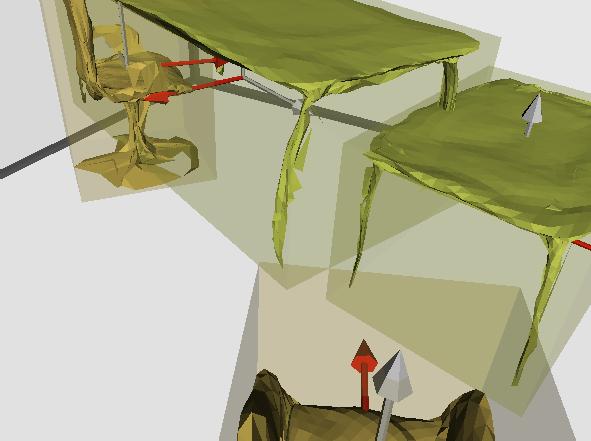}
		\includegraphics[width=\textwidth]
		{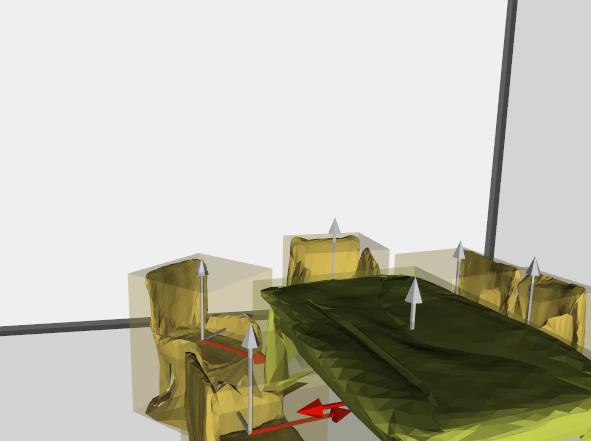}
		\includegraphics[width=\textwidth]
		{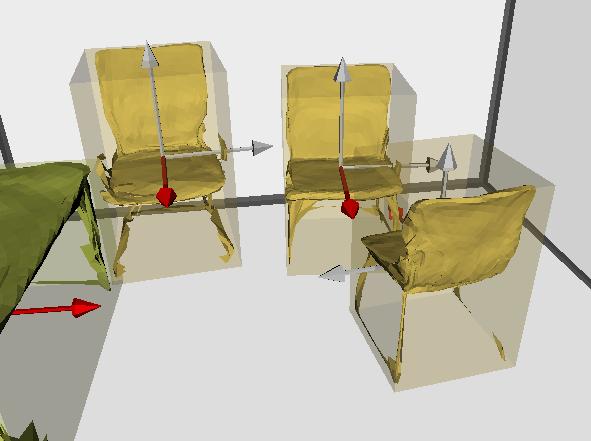}
		\includegraphics[width=\textwidth]
		{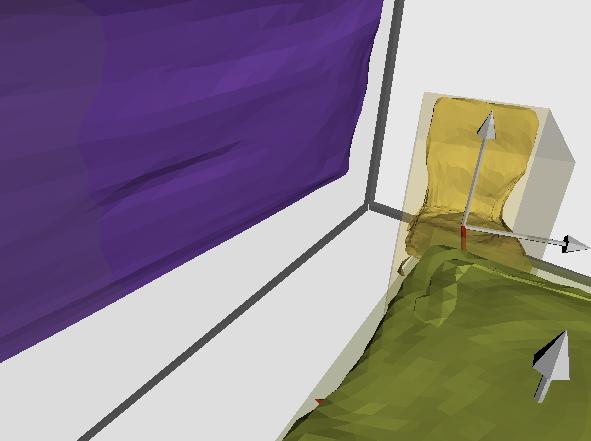}
		\includegraphics[width=\textwidth]
		{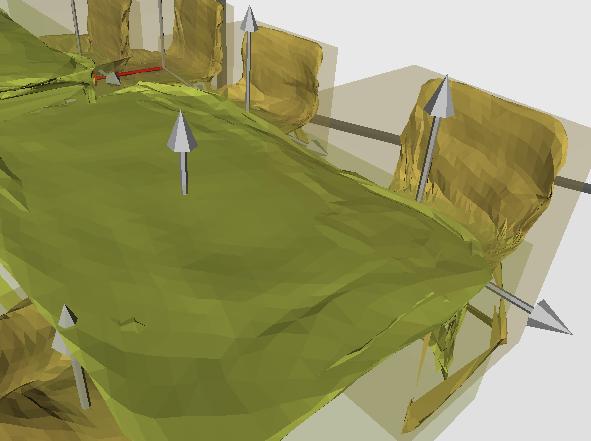}
		\includegraphics[width=\textwidth]
		{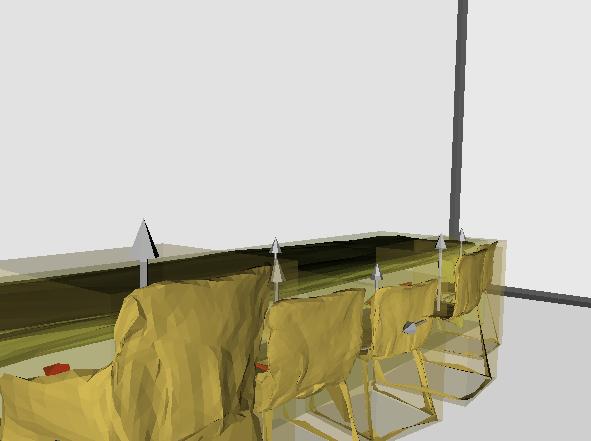}
	\end{subfigure}
	\begin{subfigure}[t]{0.15\textwidth}
		\includegraphics[width=\textwidth]
		{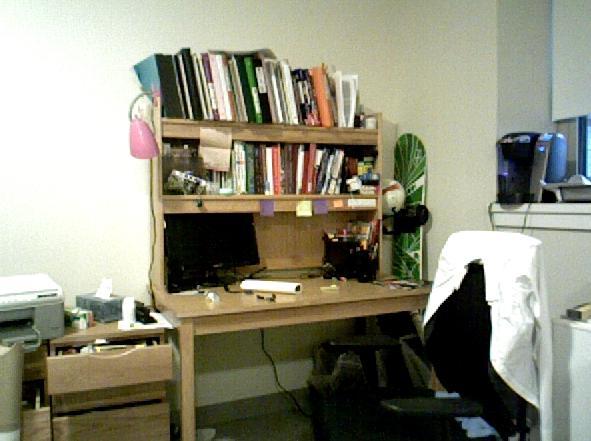}
		\includegraphics[width=\textwidth]
		{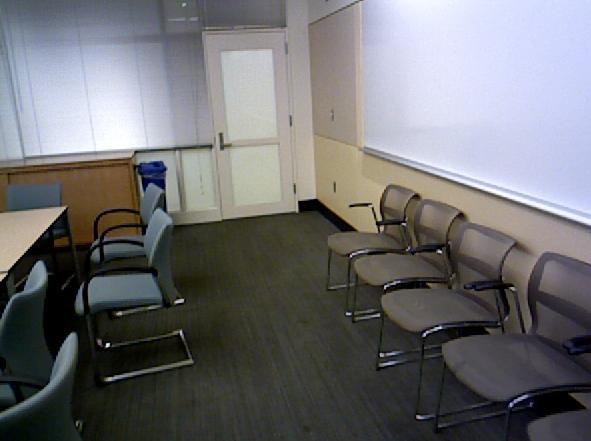}
		\includegraphics[width=\textwidth]
		{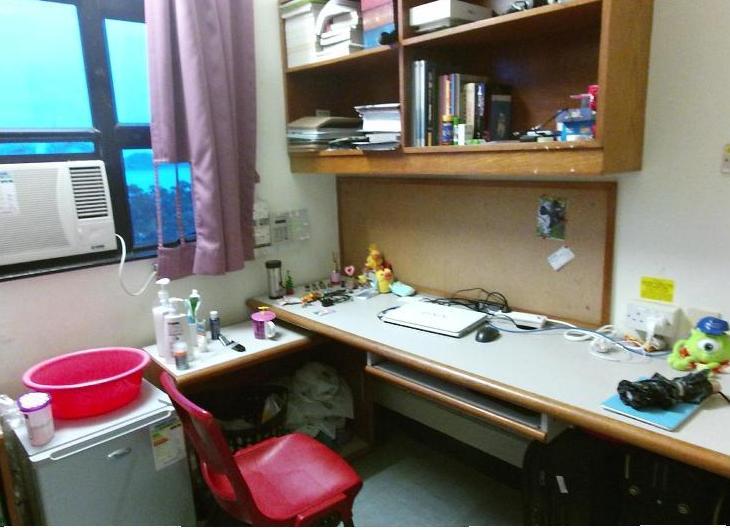}
		\includegraphics[width=\textwidth]  
		{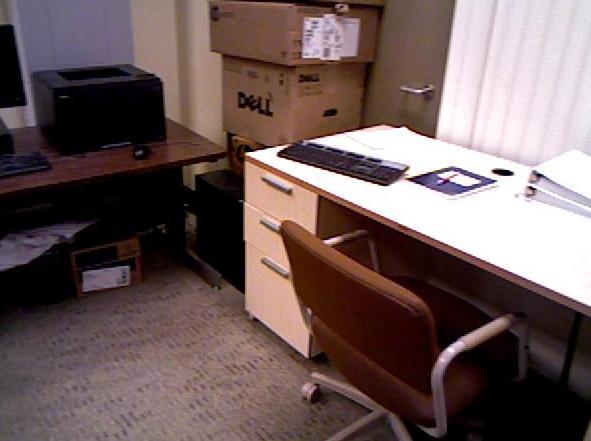}
		\includegraphics[width=\textwidth]
		{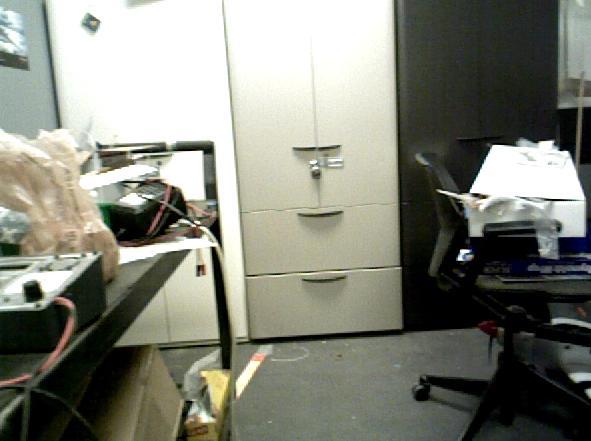}
		\includegraphics[width=\textwidth]
		{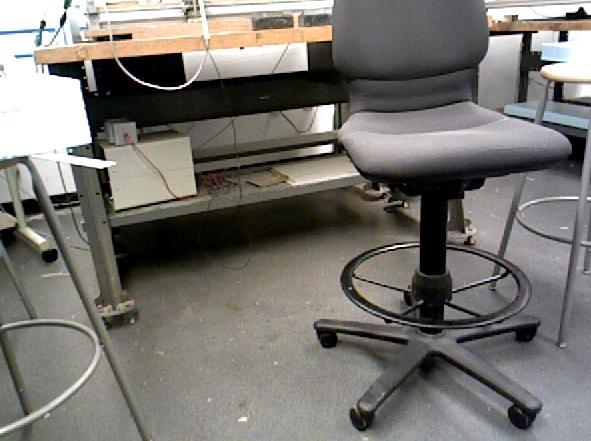}
		\includegraphics[width=\textwidth]
		{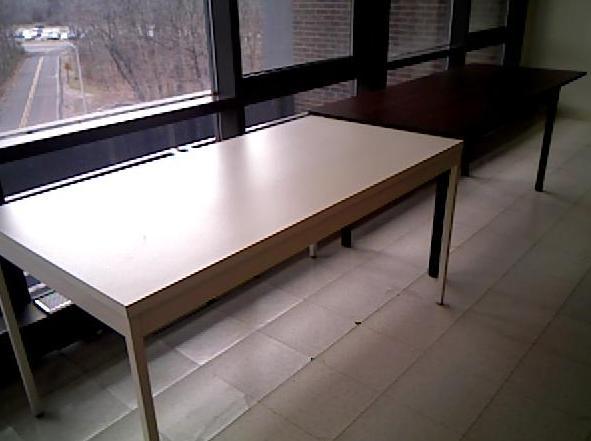}
		\includegraphics[width=\textwidth]
		{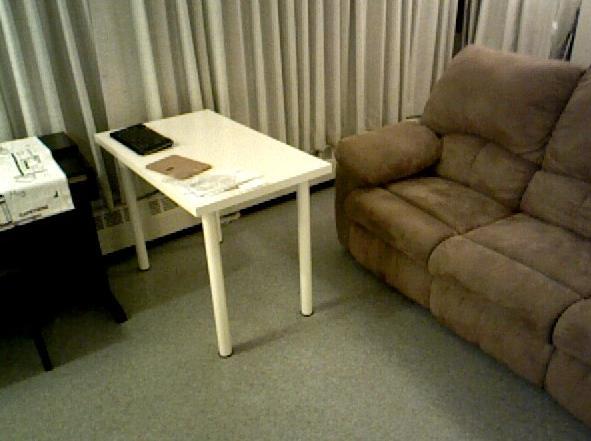}
		\includegraphics[width=\textwidth]
		{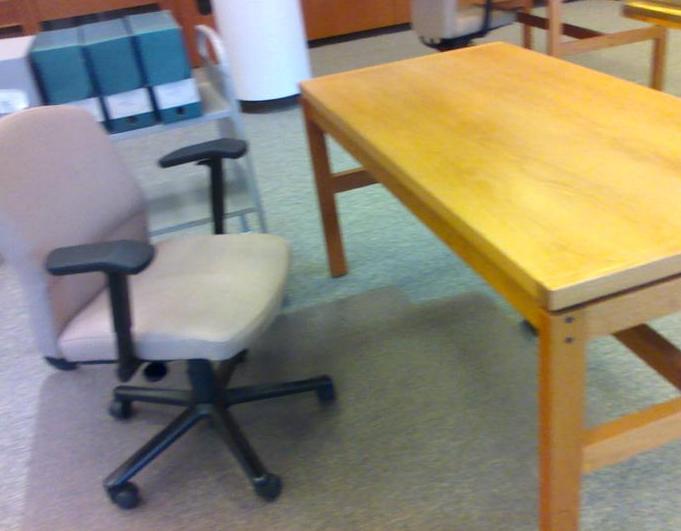}
		\includegraphics[width=\textwidth]
		{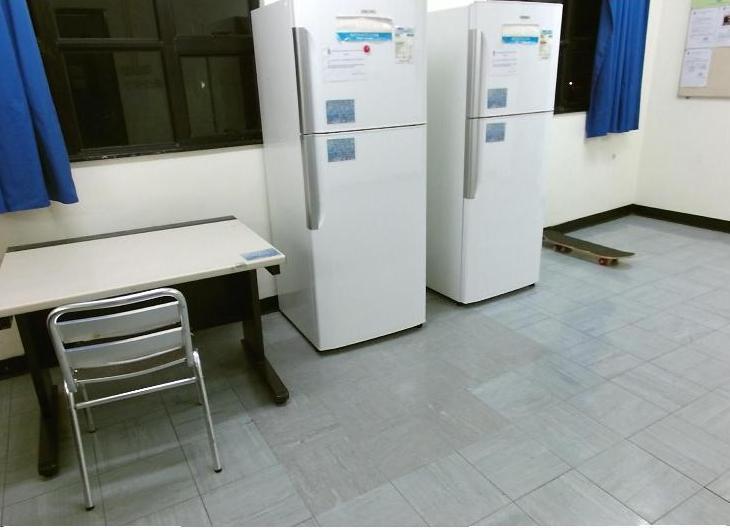}
		\includegraphics[width=\textwidth]
		{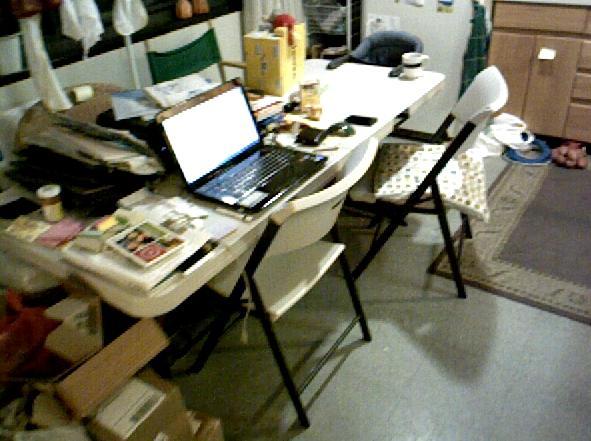}
	\end{subfigure}
	\begin{subfigure}[t]{0.15\textwidth}
		\includegraphics[width=\textwidth]
		{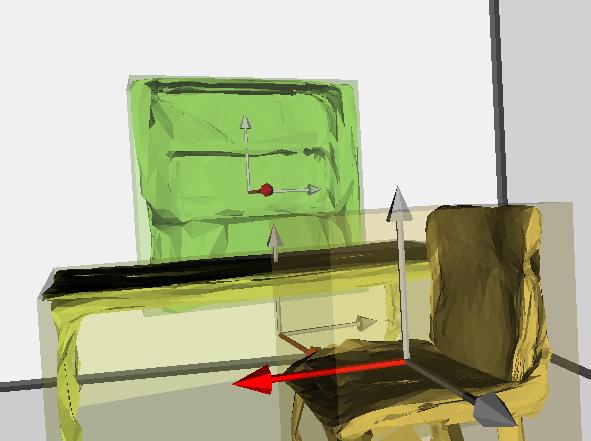}
		\includegraphics[width=\textwidth]
		{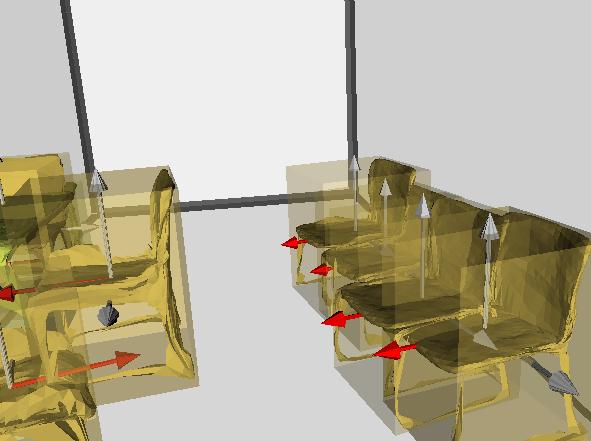}
		\includegraphics[width=\textwidth]
		{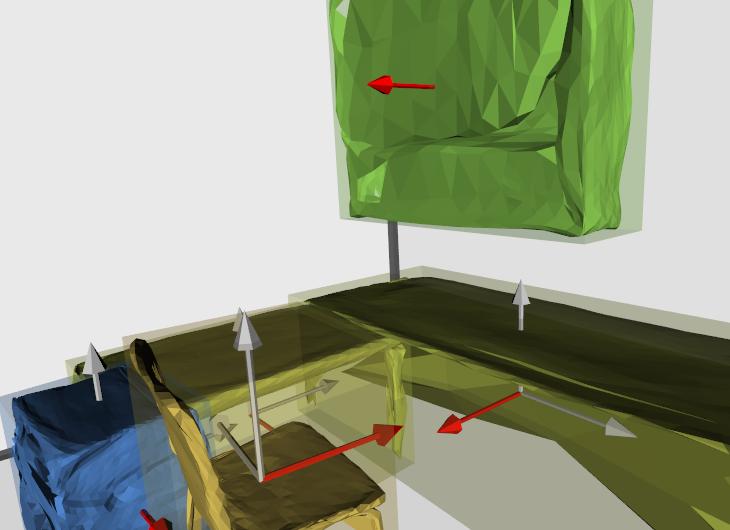}
		\includegraphics[width=\textwidth]  
		{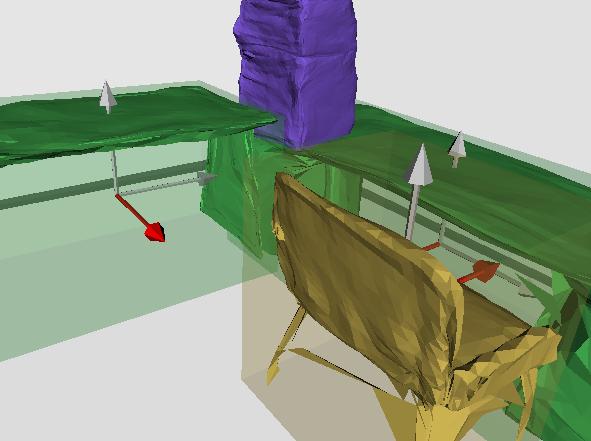}
		\includegraphics[width=\textwidth]
		{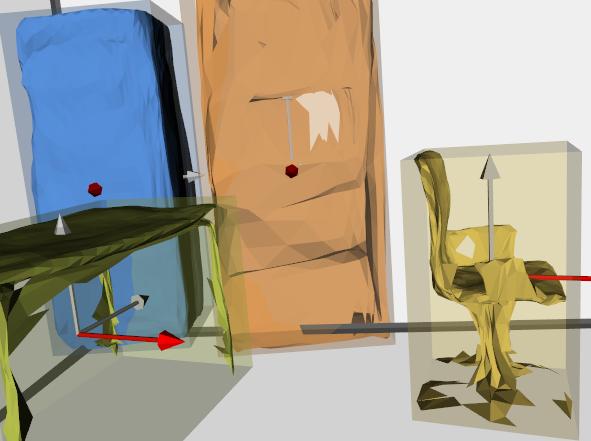}
		\includegraphics[width=\textwidth]
		{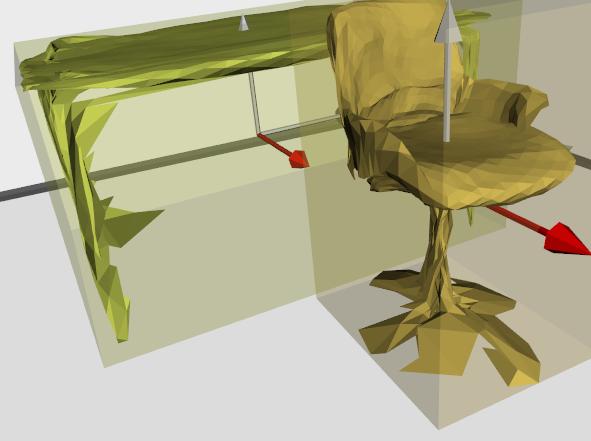}
		\includegraphics[width=\textwidth]
		{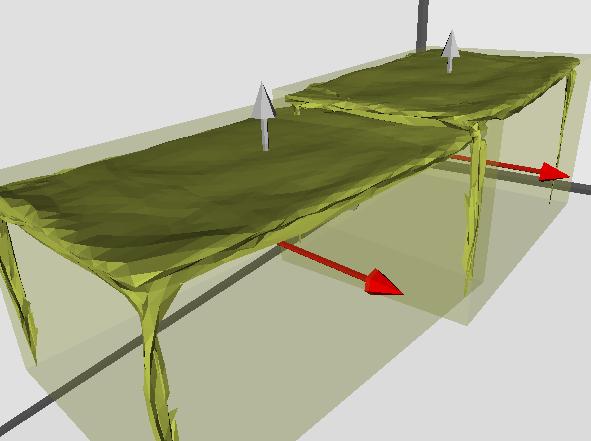}
		\includegraphics[width=\textwidth]
		{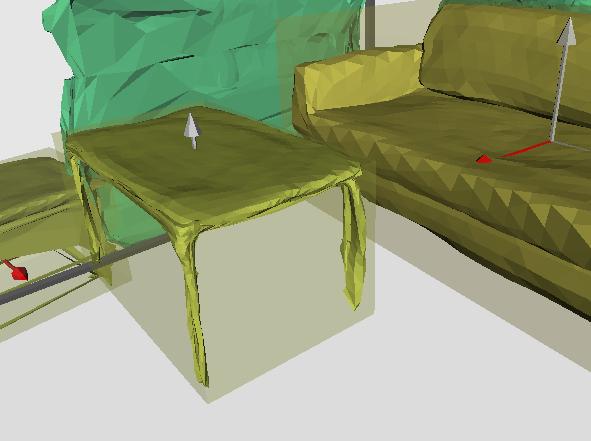}
		\includegraphics[width=\textwidth]
		{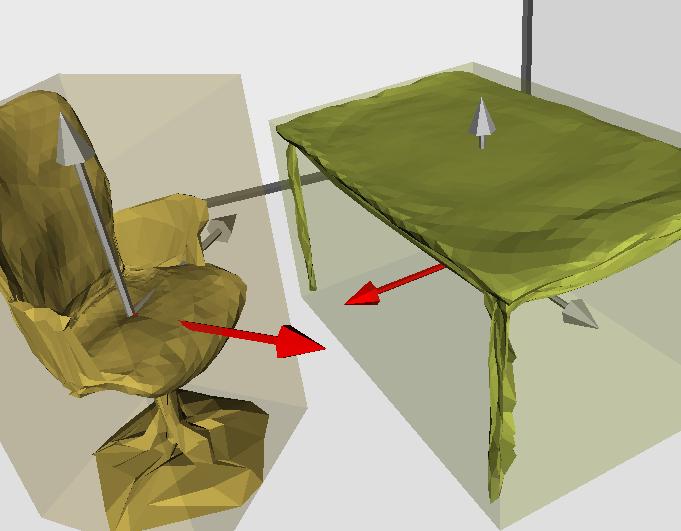}
		\includegraphics[width=\textwidth]
		{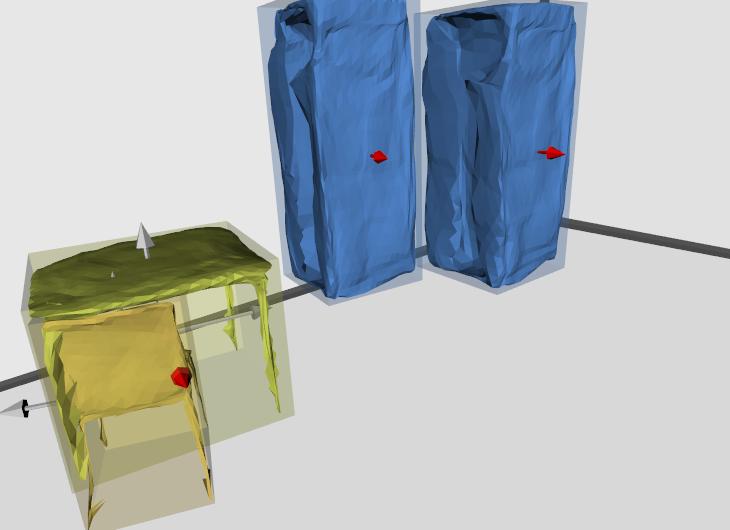}
		\includegraphics[width=\textwidth]
		{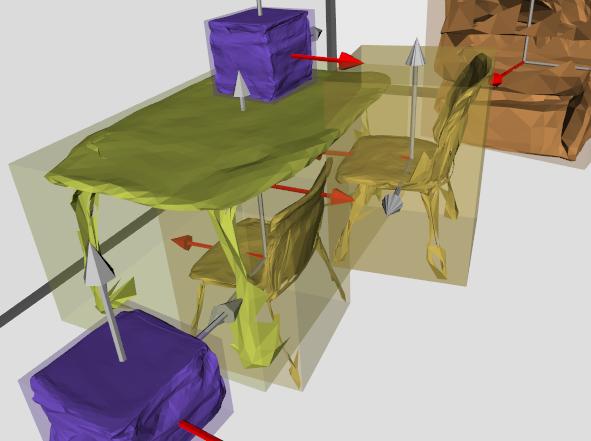}
	\end{subfigure}
	\caption{Reconstruction results of test samples on SUN RGB-D}
	\label{samples}
\end{figure*}

We list more reconstruction samples from the testing set of SUN RGB-D in Figure~\ref{samples}.

\end{document}